\documentclass[accepted]{uai2025} 
                        
\usepackage[american]{babel}

\usepackage[dvipsnames,svgnames]{xcolor}         
\usepackage{natbib} 
\usepackage{tikz} 

\usepackage{hyperref}       
\hypersetup{
    colorlinks=true,
    linkcolor=blue, 
    citecolor=green, 
    urlcolor=red 
}
 
\usepackage{url}            
\usepackage{booktabs}       
\usepackage{amsfonts}       
\usepackage{nicefrac}       
\usepackage{microtype}      

\usepackage{amsmath,amssymb}
\usepackage{mathtools}
\usepackage{amsthm}
\usepackage{bm}
\usepackage{enumitem}

\usepackage{algorithm}
\usepackage{algpseudocode}

\usepackage{multirow}
\usepackage{caption} 
\usepackage{adjustbox}
\usepackage{colortbl}
\usepackage{subcaption}
\usepackage{graphicx}

\usepackage{wrapfig,lipsum}

\newtheorem{theorem}{Theorem}
\newtheorem{lemma}{Lemma}
\newtheorem{proposition}{Proposition}

\definecolor{my_color}{named}{blue}

\definecolor{Gray}{gray}{0.85}
\newcolumntype{a}{>{\columncolor{Gray}}c}

\usepackage{authblk}

\newcommand{\correspondence}[1]{\thanks{Correspondence to: \texttt{\scriptsize{#1}}}}
\newcommand{\corresponding}{\thanks{Corresponding Authors}}
\newcommand{\aff}{\thanks{Work conducted while affiliated with KAIST}}

\title{Flat Posterior Does Matter For Bayesian Model Averaging}

\author[1]{Sungjun Lim}
\author[1]{Jeyoon Yeom}
\author[2]{Sooyon Kim}
\author[1]{Hoyoon Byun}
\author[3]{Jinho Kang}
\author[4]{Yohan Jung\corresponding\aff}
\author[3]{Jiyoung Jung$^\dagger$}
\author[1]{Kyungwoo Song\correspondence{kyungwoo.song@gmail.com}$^\dagger$}

\affil[ ]{%
\textsuperscript{1}Yonsei University \quad
\textsuperscript{2}Ohio State University \quad
\textsuperscript{3}University of Seoul \quad
\textsuperscript{4}RIKEN AIP%
}

\begin{document}
\maketitle

\begin{abstract}
Bayesian neural networks (BNNs) estimate the posterior distribution of model parameters and utilize posterior samples for Bayesian Model Averaging (BMA) in prediction. However, despite the crucial role of flatness in the loss landscape in improving the generalization of neural networks, its impact on BMA has been largely overlooked. In this work, we explore how posterior flatness influences BMA generalization and empirically demonstrate that \emph{(1) most approximate Bayesian inference methods fail to yield a flat posterior} and \emph{(2) BMA predictions, without considering posterior flatness, are less effective at improving generalization}. To address this, we propose Flat Posterior-aware Bayesian Model Averaging (FP-BMA), a novel training objective that explicitly encourages flat posteriors in a principled Bayesian manner. We also introduce a Flat Posterior-aware Bayesian Transfer Learning scheme that enhances generalization in downstream tasks. Empirically, we show that FP-BMA successfully captures flat posteriors, improving generalization performance\footnote{Code is available at~\url{https://github.com/MLAI-Yonsei/FP-BMA}}.  

\end{abstract}

\section{Introduction}\label{sec:introduction}
Bayesian neural networks (BNNs) provide a theoretically grounded framework for modeling uncertainty in deep learning by approximating the posterior distribution of model parameters~\citep{mackay1992practical, hinton1993keeping, neal2012bayesian}. The approximated posterior is used for making predictions through Bayesian Model Averaging (BMA)~\citep{wasserman2000bayesian, fragoso2018bayesian, wilson2020bayesian, zeng2024collapsed}. It allows BNNs to account for uncertainty in predictions, leading to more reliable outcomes compared to the deterministic neural networks (DNNs)~\citep{kapoor2022uncertainty, kristiadi2022being}. The accuracy and robustness of BNN predictions are heavily dependent on the quality of the approximated posterior~\citep{kristiadi2022posterior, wenzel2020good}.

The flatness of loss landscape has been strongly associated with better generalization ability, as they represent solutions that are less sensitive to small perturbations in model parameters~\citep{hochreiter1997flat, keskar2016large, neyshabur2017exploring}. The flatness has been extensively studied in the context of DNNs, but no comprehensive analysis has been conducted on its role in BNNs or its impact on BMA. SA-BNN~\citep{nguyen2023flat} incorporated a flat-seeking optimizer into BNNs but merely adapted a DNN-based optimizer without considering the probabilistic nature of BNNs, leading to only limited improvements. On the other hand, E-MCMC~\citep{li2023entropy} introduced a guidance model to achieve flat posteriors, but this approach is less suited for large-scale models.

In this work, we first demonstrate that BNNs often struggle to capture the flatness. In detail, we compare the flatness of various BNN frameworks against that of DNNs and demonstrate that \emph{(1) most approximate Bayesian inference methods fail to yield a flat posterior} and \emph{(2) BMA predictions, without considering posterior flatness, are less effective at improving generalization}. These findings highlight the need for an optimization strategy that accounts for the probabilistic nature of BNNs to estimate flat posteriors effectively.

Therefore, we propose Flat Posterior-aware Bayesian Model Averaging (FP-BMA), a novel optimization that explicitly targets the flat posterior. We first compute an adversarial posterior in the vicinity of the current posterior, which maximizes the BNN loss. After that, we update the posterior by employing the gradient of the adversarial posterior with respect to the loss. We show that the proposed FP-BMA is an extended version of previous flatness-aware optimizers, Sharpness-aware Minimization (SAM)~\citep{foret2020sharpness}, Fisher SAM (FSAM)~\citep{kim2022fisher}, and Natural Gradient (NG)~\citep{amari1998natural} with specific conditions. In addition, we introduce a Flat Posterior-aware Bayesian Transfer Learning scheme integrated with FP-BMA, enabling effective capture of flatness. This approach enhances robustness against model misspecification, when the prior is not well-suited for fine-tuning BNNs on downstream tasks. We show that FP-BMA improves the generalization performance of BNNs, particularly in few-shot classification and distribution shift, by ensuring a flat posterior.

Our major contributions are summarized as follows:
\begin{itemize}
    \item We demonstrate that BNNs often struggle to capture the flatness, and BMA can be ineffective without flatness.
    \item We propose FP-BMA, a flat posterior-seeking optimizer that generalizes loss geometric optimizers such as SAM, FSAM, and NG.
    \item We introduce Flat Posterior-aware Bayesian Transfer Learning, which leverages a pre-trained model as a prior and effectively enhances robustness against model misspecification through a flat posterior.
\end{itemize}

\section{Preliminary}\label{sec:preliminary}
\subsection{Bayesian Neural Networks}\label{subsec:bayesian_neural_network}

\paragraph{Training}
Let $w \subseteq \mathbb{R}^p$ be the model parameter of BNN and $\mathcal{D} = \{(x, y)\}$ be the datasets with inputs $x$ and outputs $y$. In principle, training BNNs aims to estimate the posterior distribution $p(w|\mathcal{D})$ based on Bayes' Rule:
\begin{equation}
\label{eq:bayes_rule}
p(w|\mathcal{D}) = \frac{p(\mathcal{D}|w)p(w)}{\int_w p(\mathcal{D}|w)p(w) dw},
\end{equation}
where $p(\mathcal{D}|w)$ and $p(w)$ denote the likelihood of data $\mathcal{D}$ and the prior distribution over $w$, respectively. 

However, the posterior of BNNs $p(w|\mathcal{D})$ is intractable in general. Hence, many approximate inference methods, including Markov Chain Monte Carlo (MCMC)~\citep{welling2011bayesian, chen2014stochastic} and Variational Inference (VI)~\citep{graves2011practical, blundell2015weight}, and other variants~\citep{ritter2018scalable, gal2016dropout, maddox2019simple}, have been employed to obtain approximate posterior $q_\theta(w|\mathcal{D})$, with distribution's parameter $\theta \subseteq \mathbb{R}^q$, pursuing $q_\theta(w|\mathcal{D}) \approx p(w|\mathcal{D})$.

\paragraph{Prediction}
For the approximate posterior $q_\theta (w | \mathcal{D})$, BNNs make predictions on unobserved data $(x^*, y^*)$ via \textit{Bayesian Model Averaging (BMA)}, which integrates predictions over the posterior distribution of the model parameters:

\begin{align}
p(y^*|x^*, \mathcal{D}) 
&\approx \int_w p(y^*|f_w(x^*))q_\theta(w|\mathcal{D}) dw \label{eq:mc_predictive} \\
&\approx \frac{1}{M} \sum_{m=1}^M p(y^*|f_{w_m}(x^*)), \ \  w_m \sim q_\theta(w|\mathcal{D}), \nonumber
\end{align}
where $f_w(\cdot)$ is predictions with parameter $w$ and $M$ denotes the number of sampled model; the first approximation uses $q_\theta(w|\mathcal{D})$ and second approximation in Eq.~\ref{eq:mc_predictive} employs Monte Carlo integration. This approach is known to improve generalization by averaging over a diverse set of models sampled from the approximate posterior, which is the core idea of BMA~\citep{wilson2020bayesian}.

\begin{figure*}[t]
    \centering
    \begin{subfigure}[t]{0.32\textwidth}
        \centering
        \includegraphics[width=1.0\linewidth]{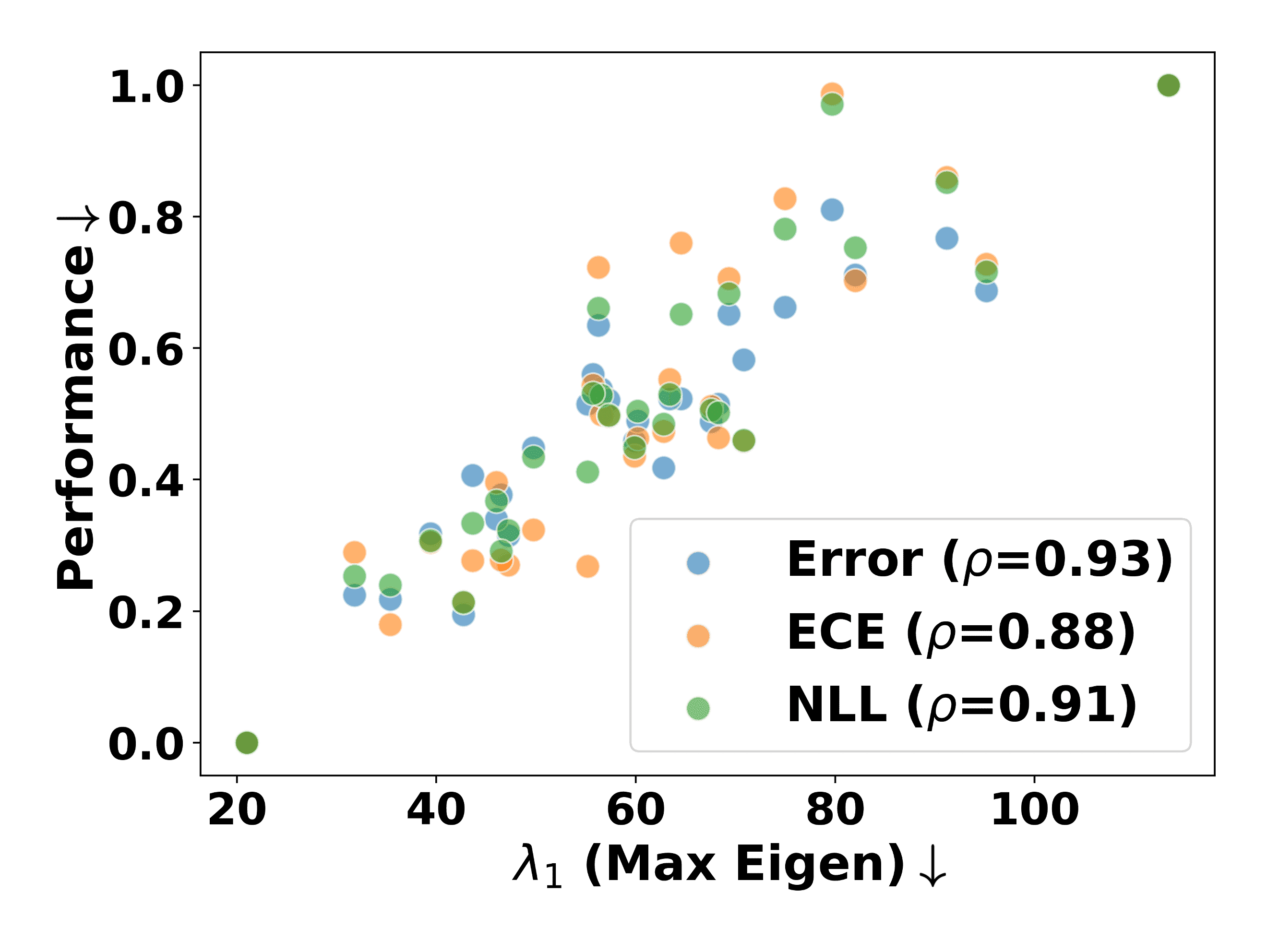}
        \captionsetup{justification=centering}
        \caption{Correlation between flatness and generalization within sampled models}
        \label{fig:correlation_plot}
    \end{subfigure}
    \hfill
    \begin{subfigure}[t]{0.32\textwidth}
        \centering
        \includegraphics[width=1.0\linewidth]{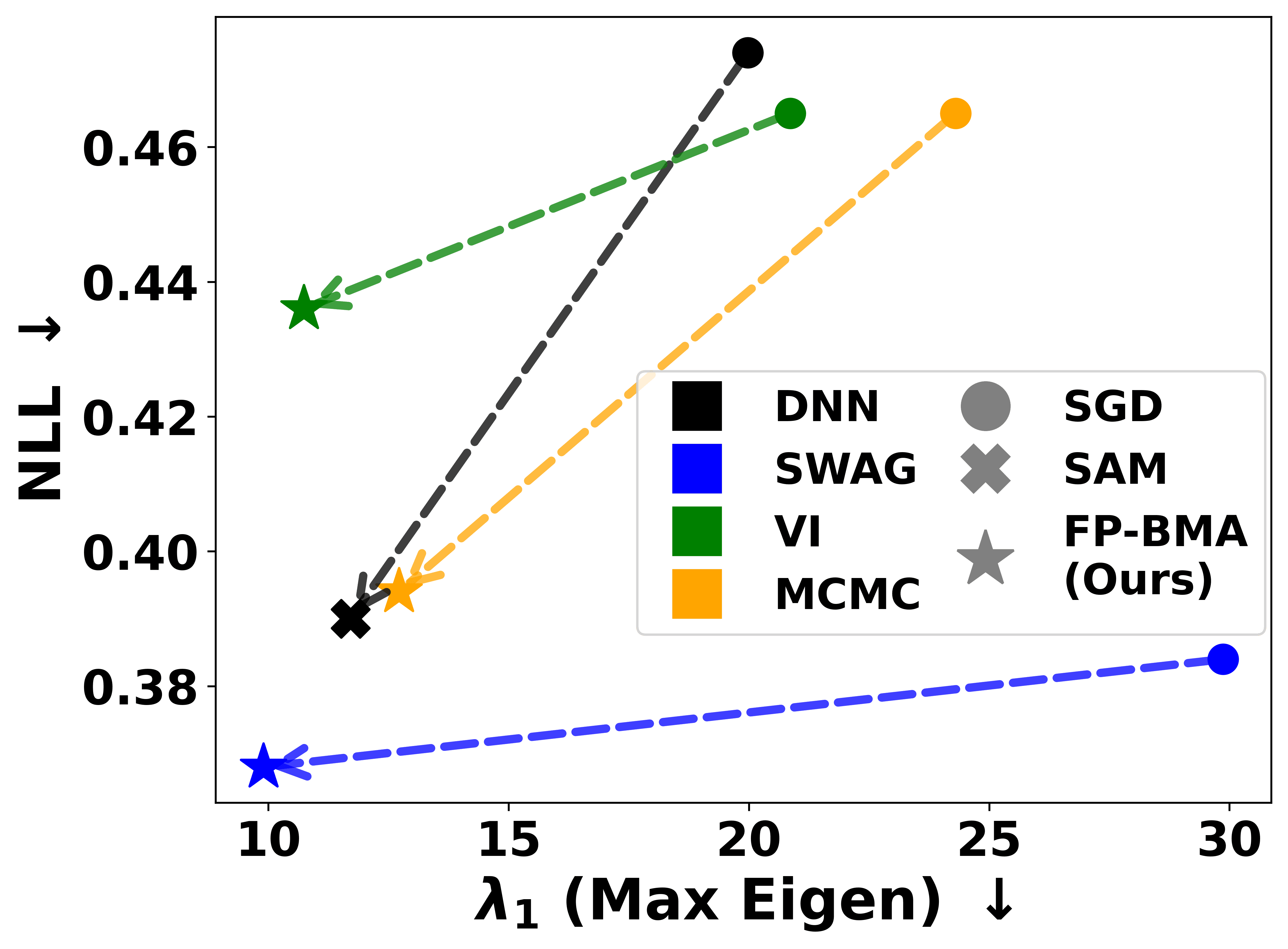}
        \captionsetup{justification=centering}
        \caption{Flatness and generalization according to the training methods}
        \label{fig:sgd_to_fpbma}
    \end{subfigure}
    \hfill
    \begin{subfigure}[t]{0.32\textwidth}
        \centering
        \includegraphics[width=1.0\linewidth]{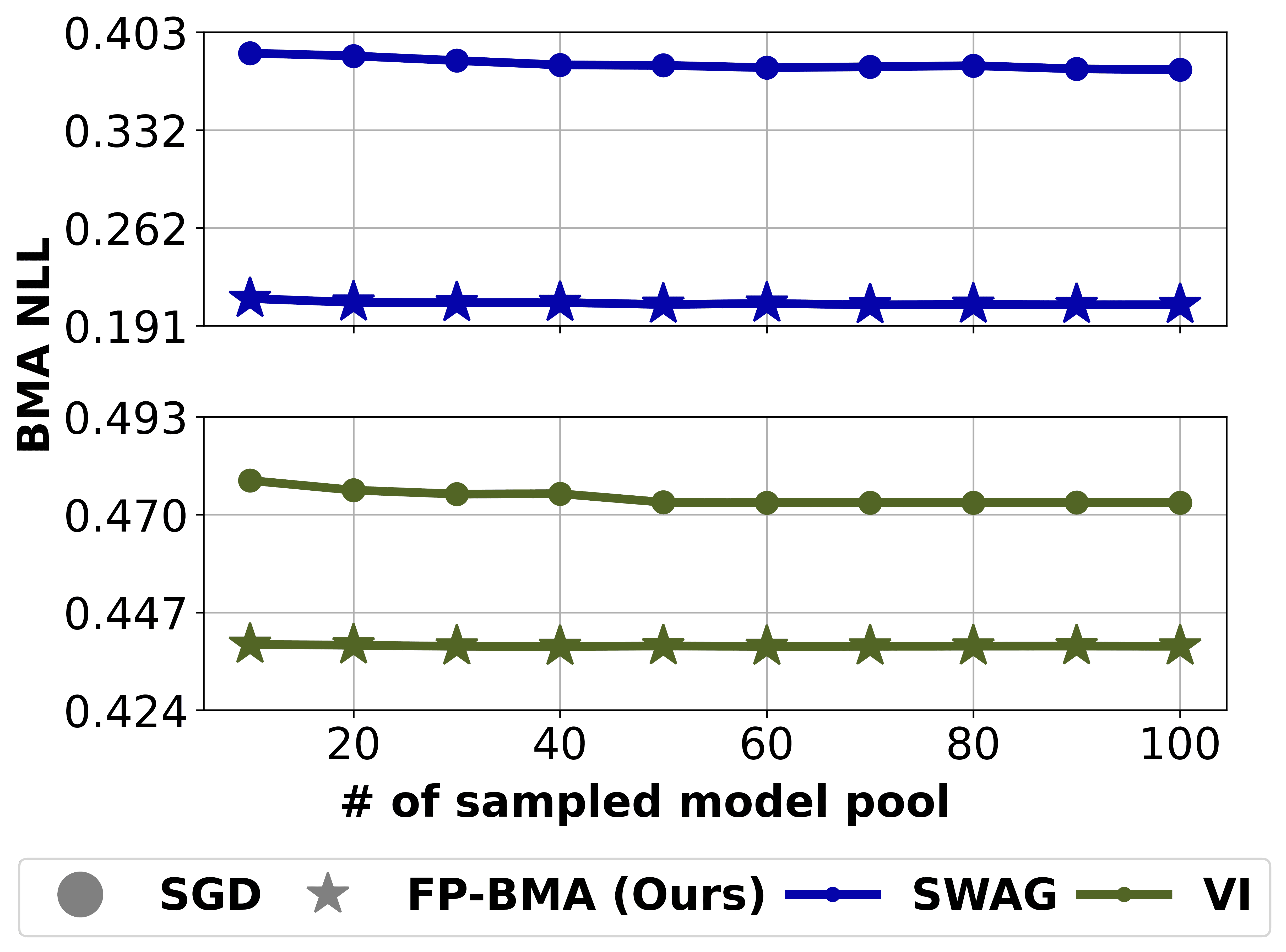}
        \captionsetup{justification=centering}
        \caption{NLL with the number of BMA models}
        \label{fig:bma_num_plot_nll}
    \end{subfigure}
    \caption{(a) Flatness, measured via the maximal Hessian eigenvalue ($\lambda_1$), is highly correlated with generalization ability (classification error, ECE, and NLL), suggesting that the flatness of models sampled from the posterior is correlated with generalization ability. (b) The existing inferences of BNNs (SWAG, VI, and MCMC) with SGD struggle to capture the flatness compared to DNNs. In contrast, the proposed Bayesian flat posterior-aware optimizer FP-BMA allows BNNs to seek flat minima, improving performance. (c) Flat posteriors are necessary, as increasing the number of BMA samples does not lead to better performance if the posterior is not flat. The proposed FP-BMA enhances posterior quality by capturing flatness and requires fewer samples for improved BMA.}
    \label{fig:combined_fig}
\end{figure*}

\subsection{Flatness and Optimization}\label{subsec:flatness_and_optimization}
As the flatness of loss surface has been known to be connected to the generalizability~\citep{hochreiter1994simplifying, hochreiter1997flat,neyshabur2017exploring}, new training methods have been presented to find the flat local optimum. Sharpness-Aware Minimization (SAM)~\citep{foret2020sharpness} is a widely adopted technique that seeks flat minima by making the model robust to small perturbations in parameters. SAM performs adversarial training by minimizing the worst-case loss in an $L_2$ neighborhood of the weights:
\begin{equation}
\label{eq:sam_loss}
\ell^\gamma_{\text{SAM}}(w) = \min_w \max_{\|\Delta w\|_p \leq \gamma} \ell(f_{w+\Delta w}(x), y),
\nonumber
\end{equation}
where $\ell(\cdot)$ is the empirical loss function (e.g., cross-entropy for classification tasks) and  $p$ is practically set to $p=2$, yielding $\Delta w = \gamma \nabla_w \ell(w) / \|\nabla_w \ell(w) \|_2$.

However, SAM's isotropic $L_2$ ball may not accurately reflect the an isotropic geometry of the loss landscape. To address this, Fisher SAM (FSAM)~\citep{kim2022fisher} improves SAM by replacing the Euclidean ball with a non-Euclidean one defined by the Fisher information matrix (FIM):
\begin{equation}
\label{eq:fsam_loss}
\ell^\gamma_{\text{FSAM}}(w) = \min_w \max_{\|F_y(w) \Delta w \|_p \leq \gamma^2} \ell(f_{w+\Delta w}(x), y),
\nonumber
\end{equation}
where $F_y(w)$ denotes the FIM and is approximated as $F_y(w) = 1/|B| \nabla_w \log p(y|x, w)^2$ with $|B|$ batch size. SAM and FSAM are both derived under deterministic $w$, and $F_y (w)$ is defined over the predictive distribution $p(y|f_w(x))$, not in the parameter space.

\section{Flatness Does Matter For Bayesian Model Averaging}\label{sec:flatness_does_matter_for_bayesian_model_averaging}
In this section, we explore the flatness of BNNs' posterior obtained from the widely-used approximate Bayesian inferences and demonstrate that flatness should be considered for BNNs based on both empirical and theoretical grounds.

\paragraph{Experimental Setup}
We empirically inspect the flatness of BNNs and observe that the generalization ability of BMA prediction improves as weight samples are drawn from a flatter posterior. To this end, we train ResNet18~\citep{he2016deep} without Batch Normalization~\citep{ioffe2015batch} on CIFAR10~\citep{krizhevsky2009learning} using following Bayesian inference methods-VI, SWAG, and MCMC-to yield the approximate posterior $q_\theta(w|\mathcal{D})$. We then compare the generalization ability, classification error, negative log-likelihood (NLL), and expected calibration error (ECE) with the flatness of the approximate posterior.

\paragraph{Flatness criterion for BNNs}
To evaluate the flatness of the posterior, we use the average of Hessian's eigenvalues, unlike in DNNs, where flatness is assessed using individual eigenvalues. This difference stems from the fact that the loss of BNNs is formulated as the marginal likelihood over the posterior, incorporating multiple parameter samples $\{w_m\}_{m=1}^{M}$ drawn from $w_m \sim q_\theta(w|\mathcal{D})$. We use the averaged $i$-th maximal eigenvalue of Hessian:
\begin{equation}
\begin{split}
\label{eq:bma_hessian}
&\lambda_i \approx \frac{1}{M}\sum_{m=1}^M \lambda_i (H_{f_m}), \quad H_{f_m} = \nabla^2 \ell\big(f_{w_m}(x), y\big),
\end{split}
\end{equation}

where $\ell\big(f_{w_m}(x), y\big) \coloneqq -\log{p(y|f_{w_m}(x))}$ denotes the likelihood using  $m$-th parameter sample $w_m \sim q_\theta(w|\mathcal{D})$. The $H_{f_m}$ denotes the Hessian of the loss $\ell \big(f_{w_m}(x), y\big)$ and $\lambda_i(H_{f_m})$ denotes the $i$-th maximal eigenvalue of Hessian. Notably, the smaller largest eigenvalues of the Hessian indicate that the model parameters lie in a flatter region of the loss surface. Therefore, the maximal eigenvalue $\lambda_{1}$ or the eigenvalue ratio $\lambda_{1}/\lambda_5$ is often used to assess the flatness of model parameters~\citep{keskar2016large, foret2020sharpness, jastrzebski2020break}.

\subsection{Need for Flatness in BMA}\label{subsec:need_for_flatness_in_bma}
\paragraph{Takeaway 1: The flatness of models sampled from the posterior is correlated with generalization ability.}
Figure~\ref{fig:correlation_plot} compares normalized generalization ability—measured by Error, ECE, and NLL—against flatness of BMA models sampled from posterior trained with SWAG. The results reveal a strong positive correlation between flatness and generalization ability, suggesting that \emph{models sampled from the posterior is correlated with generalization ability, same as DNNs.}
We confirm that this property holds across different learning rate schedulers, as shown in Figure~\ref{fig:additional_corr_plot} (Appendix~\ref{subsubsec:correlation_between_flatness_and_generalization}).

\paragraph{Takeaway 2: It is essential to approximate a flat posterior for BMA.}
We also establish a generalization error bound for BMA that explicitly involves the flatness of the posterior. First, we show that the flatness of BMA is determined by that of individual BMA samples, highlighting
the necessity of a flat posterior for effective BMA performance.

\begin{lemma}
\label{lemma:bma_eign_bound}
Let twice differentiable loss $\ell(\cdot)$, predictions of model $f_m(\cdot)$ parameterized by $w_m$, and predictions of BMA $f_{\text{BMA}}(\cdot).$
With $M$ model sample $\{w_m\}_{m=1}^M$, the maximal eigenvalue of averaged Hessian of loss $\lambda_{\text{max}}(H_{f_{\text{BMA}}})$ is bounded as follow:
\begin{align}
\label{eq:bma_eign_bound}
&\max \left( \Bigg{\{} \frac{1}{M} \bigg{(} \lambda_{\max}(H_{f_m}) + \sum_{\substack{n=1 \\ n \neq m}}^{M} \lambda_{\min}(H_{f_n}) \bigg{)} \Bigg{\}}_{m=1}^M \right) \\
& \qquad \le \lambda_{\max}(H_{f_{\text{BMA}}}) \le \frac{\sum_{m=1}^M \lambda_{\max}(H_{f_m})}{M}.
\end{align}
\end{lemma}

Lemma~\ref{lemma:bma_eign_bound} implies that as $\lambda_{\max}(H_{f_m})$ decreases in Eq.~\ref{eq:bma_eign_bound}, where it appears in both the lower and upper bounds, the corresponding $\lambda_{\max}(H_{f_{\text{BMA}}})$ also decreases. This decrease in
$\lambda_{\max}(H_{f_{\text{BMA}}})$ represents that that the BMA prediction operates within flatter minima. Given Lemma~\ref{lemma:bma_eign_bound}, the following theorem shows that the generalization error of the BMA predictor is directly controlled by the flatness of the posterior, as measured by the maximal Hessian eigenvalue~\citep{luo2024explicit}.

\begin{theorem}[Informal]\label{theorem:bma_generalization}
Let $f_{\text{BMA}}$ be the BMA predictor obtained by averaging over posterior samples. Then, with high probability,
\begin{align*}
    \ell_{\mathcal{D}}(f_{\text{BMA}})
    \leq \ell_{\mathcal{S}}(f_{\text{BMA}})
    + \frac{p \sigma^2}{2} \lambda_{\max}(H_{f_{\text{BMA}}})
    + O\left(\sigma^3 p^3 \right)
\end{align*}
where $\ell_{\mathcal{D}}$ and $\ell_{\mathcal{S}}$ denote the population and empirical loss, respectively, $p$ is the number of model parameters, $n$ is the sample size, $\sigma$ is a smoothing parameter, and $H_{f_{\text{BMA}}}$ is the Hessian of the loss evaluated at $f_{\text{BMA}}$.
\end{theorem}

This result formally supports our main message: \emph{BMA with a flat posterior---that is, a posterior whose samples $f_m$ exhibit smaller $\lambda_{\max}(H_{f_m})$---leads to a tighter generalization error bound.} Thus, posterior flatness is not only empirically correlated with generalization, but also a theoretically well-justified objective for BMA. A detailed proof of Theorem~\ref{theorem:bma_generalization} is provided in Appendix~\ref{subsec:proof_of_theorem_1}.

\subsection{Insufficient Flatness of BMA} \label{subsec:insufficient_flatness_of_bma}
\paragraph{Takeaway 3: Most approximate Bayesian inference methods struggle to produce a flat posterior.}
We investigate whether existing approximate Bayesian inference methods can produce the flat posterior of BNNs. Figure~\ref{fig:sgd_to_fpbma} illustrates how NLL and posterior flatness vary depending on whether flatness in the loss surface is taken into account during optimization. We observe that \emph{the approximate posteriors of BNNs do not show better flatness compared to that of DNNs, obtained from the SAM optimizer}.

On the other hand, the proposed FP-BMA, which will be described in Section~\ref{subsec:flat_posterior_aware_optimizer}, allows BNNs to seek flat minima and thus leads to better performance. We also confirm consistent results on various learning rate schedulers and generalization performance metrics, as described in Appendix~\ref{subsec:insufficient_flatness_of_bma_app}.

\paragraph{Takeaway 4: Increasing the
number of BMA samples does not lead to better performance without a flat posterior.}
Figure~\ref{fig:bma_num_plot_nll} compares the NLL of BMA predictions for two posteriors—one considering flatness and the other not. The results show that \emph{simply increasing the number of weight samples in BMA does not outperform BMA with a flat posterior}, highlighting the importance of posterior flatness for better generalization. On the other hand, the proposed FP-BMA, which will be described in Section~\ref{subsec:flat_posterior_aware_optimizer}, enhances posterior quality by capturing flatness and requires fewer samples for improved BMA. Additional results in Appendix~\ref{subsec:performance_changes_based_on_the_number_of_models_in_bma} confirm this trend across different learning rate schedulers and metrics.

\section{Bayesian Model Averaging with Flat Posterior}\label{sec:bayesian_model_averaging_with_flat_posterior}
Our theoretical analysis and empirical findings suggest that a flat posterior in BNNs is crucial for generalization but is not achieved by existing approximate Bayesian inference methods. To address this, we propose an optimizer that encourages a flat posterior (Section \ref{subsec:flat_posterior_aware_optimizer}) introduce Bayesian transfer learning combined with diverse BNN frameworks (Section \ref{subsec:bayesian_transfer_learning}).

\subsection{Flat Posterior-aware Optimizer}
\label{subsec:flat_posterior_aware_optimizer}
To deal with the probabilistic nature of BNNs, we suggest a
new objective function based on VI:
\begin{equation}
\label{eq:main_loss}
\begin{aligned}
    \ell_{\text{FP-BMA}}^\gamma(\theta) = &\min_\theta \max_{d|\theta+\Delta\theta, \theta| \leq \gamma^2} \ell(\theta+\Delta\theta) \\
    &+ \beta \textrm{D}_{\textrm{KL}} [q_\theta (w|\mathcal{D}) || p (w)] 
\end{aligned}
\end{equation}

\begin{equation}
\label{eq:divergence}
    \textrm{s.t.} \ \  d|\theta+\Delta\theta, \theta| = \textrm{D}_{\textrm{KL}} \big[ q_{\theta+\Delta\theta}(w |\mathcal{D}) \ || \ q_{\theta}(w | \mathcal{D}) \big],
\end{equation}
where $\theta$ and $\Delta\theta$ denote the variational parameters and their perturbation, respectively. $\ell(\theta + \Delta\theta)$ denotes empirical loss under perturbated posterior $q_{\theta + \Delta\theta}(w|\mathcal{D})$, and $\beta$ is a hyperparameter that controls the influence of the prior.

Given new objective  $\ell_{\text{FP-BMA}}^\gamma(\theta)$ in Eq.~\ref{eq:main_loss},
the variational parameter $\theta$ is updated using the approximate gradient
\begin{equation}
\label{eq:update_sabma}
\nabla_\theta \ell^\gamma_{\text{FP-BMA}} (\theta) \approx \nabla_\theta \ell(\theta + \Delta\theta_{\text{FP-BMA}}),
\end{equation}
where the parameter perturbation $\Delta\theta_{\text{FP-BMA}}$ is first computed as:
\begin{equation}
\label{eq:Delta_theta_star}
        \Delta\theta_{\text{FP-BMA}} = \gamma\frac{F_\theta(\theta)^{-1} \nabla_\theta \ell(\theta)}{\sqrt{\nabla_\theta \ell(\theta)^T F_\theta(\theta)^{-1} \nabla_\theta \ell(\theta)}},
\end{equation}
using FIM $F_\theta({\theta}){=}\mathbb{E}_{w,\mathcal{D}} [\nabla_{\theta}\log q_\theta(w|\mathcal{D}) \nabla_{\theta} \log q_\theta(w|\mathcal{D})^{T}]$. After that, the gradient $\nabla_{\theta} \ell(\theta)$ is evaluated at $\theta + \Delta\theta_{\text{FP-BMA}}$. We notate our objective as FP-BMA and provide a detailed formula derivation in Appendix~\ref{subsec:derivation_odf_bayesian_flat_seeking_optimizer}. 

Our proposed FP-BMA optimizer offers several key advantages, which are detailed in the following paragraphs. The main advantages of FP-BMA are summarized as follows:
\begin{itemize}
    \item \textbf{Implicit Flatness Control}
    \item \textbf{KL-based Bayesian Perturbation Ball}
    \item \textbf{Generalized Version of Geometric Optimizers}
\end{itemize}

\paragraph{Implicit Flatness Control}
Eq.~\ref{eq:main_loss} implicitly controls sharpness by penalizing solutions that are sensitive to parameter perturbations. This can be formalized via a second-order Taylor expansion of the inner maximization:
\begin{align}
\label{eq:second_order_inner_max}
&\max_{d|\theta+\Delta\theta, \theta| \leq \gamma^2} \ell(\theta+\Delta\theta) \\
& \quad \approx \max_{d|\theta+\Delta\theta, \theta| \leq 1} \left( \ell(\theta) + \gamma^2 \Delta\theta^\top \nabla_\theta \ell(\theta) + \frac{\gamma^4 \lambda_1(H_{f_\theta})}{2} \right) \nonumber
\end{align}
where $\lambda_1(H_{f_\theta})$ is the maximal eigenvalue of the Hessian. Thus, minimizing Eq.~\ref{eq:main_loss} inherently seeks solutions with lower sharpness, ensuring the variational posterior is concentrated in flatter regions of the loss landscape.

\paragraph{KL-based Bayesian Perturbation Ball}
Unlike deterministic optimizers such as SAM and FSAM, our method constrains the perturbation in distributional space via the KL-divergence, as shown in Eq.~\ref{eq:divergence}. This approach leverages local curvature information without expensive inner-loop optimization, making the method scalable and practical for large models. In addition, for Gaussian variational posteriors, the KL-based constraint naturally captures both mean and variance changes—providing a richer and more Bayesian-consistent notion of flatness.

\begin{figure}[t]
\centering
\includegraphics[width=\linewidth]{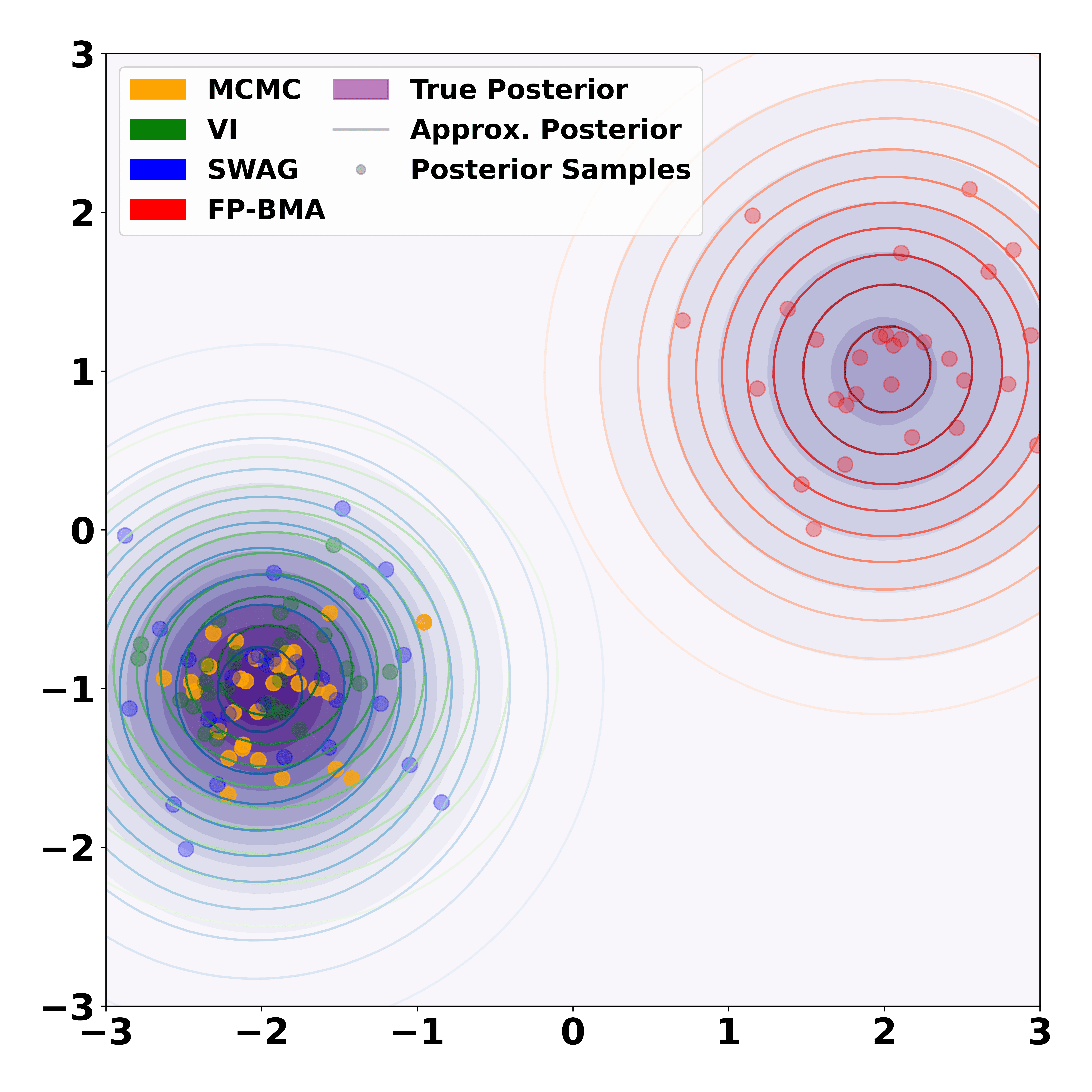} 
\caption{Posterior approximation with synthetic example. When both flat and sharp modes coexist, we compared how optimizers approximate the posterior. Unlike other methods, the proposed FP-BMA converged to the flat mode.}
\label{fig:posterior_approx}
\end{figure}

\begin{table*}[t]
\caption{Performances (ACC, ECE, and NLL) of learning from scratch with ResNet18 and modified ViT-B/16$^{\dagger}$. FP-BMA (VI), FP-BMA (MCMC), and FP-BMA (SWAG) indicate the specific BNN framework combined with FP-BMA. \textbf{Bold} highlights the best performance within each BNN framework, while \textcolor{red}{red} indicates the overall best performance across all frameworks. FP-BMA achieves superior performance across all BNN frameworks on both CIFAR10 and CIFAR100.}
\label{tab:scratch_r18_vitb16}
\adjustbox{width=1.0\linewidth}{
\begin{tabular}{ccccccccccccc}
\toprule
Backbone & \multicolumn{6}{c}{ResNet18} & \multicolumn{6}{c}{ViT-B/16$^{\dagger}$} \\ \cmidrule(lr){2-7} \cmidrule(lr){8-13}   
Dataset & \multicolumn{3}{c}{CIFAR10}        & \multicolumn{3}{c}{CIFAR100} & \multicolumn{3}{c}{CIFAR10}        & \multicolumn{3}{c}{CIFAR100}      \\ \cmidrule(lr){2-4} \cmidrule(lr){5-7} \cmidrule(lr){8-10} \cmidrule(lr){11-13}
Method        & ACC $\uparrow$     & ECE $\downarrow$    & NLL $\downarrow$    & ACC $\uparrow$     & ECE $\downarrow$    & NLL $\downarrow$  & ACC $\uparrow$     & ECE $\downarrow$    & NLL $\downarrow$    & ACC $\uparrow$     & ECE $\downarrow$    & NLL $\downarrow$   \\ \midrule\midrule
SGD & $83.28_{\pm 0.49}$ & $0.058_{\pm 0.005}$ & $0.540_{\pm 0.006}$ & $50.33_{\pm 0.62}$ & $0.123_{\pm 0.016}$ & $1.976_{\pm 0.055}$ & $81.20_{\pm 1.31}$ & $0.050._{\pm 0.002}$ & $0.569_{\pm 0.027}$ & $48.66_{\pm 0.21}$ & $0.062_{\pm 0.013}$ & $1.956_{\pm 0.021}$ \\
SAM          & $\textbf{87.59}_{\pm 3.10}$ & $\textbf{0.031}_{\pm 0.017}$ & $\textbf{0.389}_{\pm 0.065}$ & $51.48_{\pm 0.05}$ & $0.096_{\pm 0.026}$ & $1.873_{\pm 0.042}$ & $81.25_{\pm 0.10}$ & $\textbf{0.020}_{\pm 0.003}$ & $\textbf{0.550}_{\pm 0.002}$ & $54.91_{\pm 4.20}$ & $0.053_{\pm 0.020}$ & $1.709_{\pm 0.148}$ \\
FSAM          & $83.38_{\pm 0.86}$ & $0.052_{\pm 0.003}$ & $0.540_{\pm 0.010}$ & $50.87_{\pm 1.29}$ & $0.114_{\pm 0.008}$ & $1.963_{\pm 0.058}$ & $\textbf{81.57}_{\pm 1.49}$ & $0.046_{\pm 0.006}$ & $0.563_{\pm 0.036}$ & $48.75_{\pm 0.42}$ & $0.055_{\pm 0.010}$ & $1.956_{\pm 0.003}$ \\
bSAM          & $84.28_{\pm 0.32}$ & $0.051_{\pm 0.010}$ & $0.502_{\pm 0.012}$ & $\textbf{52.55}_{\pm 0.30}$ & $\textbf{0.087}_{\pm 0.011}$ & $\textbf{1.802}_{\pm 0.027}$ & $80.33_{\pm 0.88}$ & $0.037_{\pm 0.007}$ & $0.588_{\pm 0.012}$ & $\textbf{57.75}_{\pm 0.29}$ & $\textbf{0.040}_{\pm 0.014}$ & $\textbf{1.573}_{\pm 0.015}$ \\ \midrule\midrule
VI         & $82.61_{\pm 0.51}$ & $0.067_{\pm 0.003}$ & $0.632_{\pm 0.008}$ & $51.45_{\pm 0.32}$ & $0.037_{\pm 0.007}$ & $1.874_{\pm 0.007}$ & $75.81_{\pm 0.88}$ & $0.027_{\pm 0.021}$ & $0.715_{\pm 0.038}$ & $48.97_{\pm 0.20}$ & $0.037_{\pm 0.012}$ & $1.965_{\pm 0.002}$ \\
\textbf{FP-BMA (VI)}       & $\textbf{85.34}_{\pm 0.18}$ & $\textbf{0.028}_{\pm 0.006}$ & $\textbf{0.431}_{\pm 0.001}$ & $\textbf{54.49}_{\pm 0.82}$ & $\textcolor{red}{\textbf{0.016}_{\pm 0.003}}$ & $\textbf{1.699}_{\pm 0.021}$ & $\textbf{76.23}_{\pm 0.44}$ & $\textbf{0.018}_{\pm 0.006}$ & $\textbf{0.692}_{\pm 0.010}$ & $\textbf{51.62}_{\pm 1.12}$ & $0.038_{\pm 0.013}$ & $\textbf{1.884}_{\pm 0.026}$ \\ \midrule\midrule
MCMC         & $84.82_{\pm 0.13}$ & $0.049_{\pm 0.001}$ & $0.523_{\pm 0.008}$ & $58.38_{\pm 0.16}$ & $0.090_{\pm 0.002}$ & $1.742_{\pm 0.014}$ & $81.80_{\pm 0.46}$ & $0.014_{\pm 0.003}$ & $0.542_{\pm 0.023}$ & $51.79_{\pm 0.29}$ & $0.081_{\pm 0.001}$ & $2.068_{\pm 0.016}$ \\
E-MCMC         & $85.45_{\pm 0.27}$ & $0.037_{\pm 0.002}$ & $0.479_{\pm 0.006}$ & $60.38_{\pm 0.21}$ & $0.074_{\pm 0.003}$ & $1.574_{\pm 0.002}$ & $81.97_{\pm 0.49}$ & $0.034_{\pm 0.004}$ & $0.545_{\pm 0.014}$ & $50.48_{\pm 0.13}$ & $0.068_{\pm 0.005}$ & $2.010_{\pm 0.007}$ \\ 
\textbf{FP-BMA (MCMC)}       & $\textbf{86.98}_{\pm 0.19}$ & $\textbf{0.030}_{\pm 0.004}$ & $\textbf{0.393}_{\pm 0.001}$ & $\textbf{61.94}_{\pm 0.37}$ & $\textbf{0.029}_{\pm 0.003}$ & $\textbf{1.467}_{\pm 0.006}$ & $\textbf{82.49}_{\pm 1.95}$ & $\textcolor{red}{\textbf{0.012}_{\pm 0.003}}$ & $\textbf{0.528}_{\pm 0.067}$ & $\textcolor{red}{\textbf{61.10}_{\pm 1.44}}$ & $\textbf{0.046}_{\pm 0.005}$ & $\textcolor{red}{\textbf{1.461}_{\pm 0.067}}$ \\ \midrule\midrule
SWAG & $88.95_{\pm 0.09}$ & $0.044_{\pm 0.015}$ & $0.349_{\pm 0.013}$ & $59.48_{\pm 0.19}$ & $\textbf{0.030}_{\pm 0.002}$ & $1.594_{\pm 0.011}$ & $83.70_{\pm 0.30}$ & $0.044_{\pm 0.011}$ & $0.493_{\pm 0.020}$ & $54.76_{\pm 2.20}$ & $0.151_{\pm 0.025}$ & $2.008_{\pm 0.136}$ \\
F-SWAG & $89.35_{\pm 0.19}$ & $0.028_{\pm 0.013}$ & $0.323_{\pm 0.010}$ & $60.44_{\pm 0.20}$ & $0.074_{\pm 0.023}$ & $1.566_{\pm 0.006}$ & $83.57_{\pm 0.41}$ & $0.046_{\pm 0.015}$ & $0.498_{\pm 0.029}$ & $56.80_{\pm 1.44}$ & $0.061_{\pm 0.017}$ & $1.733_{\pm 0.073}$ \\
\textbf{FP-BMA (SWAG)}       & $\textcolor{red}{\textbf{89.84}_{\pm 0.30}}$ & $\textcolor{red}{\textbf{0.019}_{\pm 0.002}}$ & $\textcolor{red}{\textbf{0.306}_{\pm 0.006}}$ & $\textcolor{red}{\textbf{63.63}_{\pm 0.60}}$ & $0.052_{\pm 0.007}$ & $
\textcolor{red}{\textbf{1.342}_{\pm 0.003}}$ & $\textcolor{red}{\textbf{84.44}_{\pm 0.58}}$ & $\textbf{0.028}_{\pm 0.008}$ & $\textcolor{red}{\textbf{0.464}_{\pm 0.011}}$ & $\textbf{57.64}_{\pm 1.42}$ & $\textcolor{red}{\textbf{0.032}_{\pm 0.005}}$ & $\textbf{1.590}_{\pm 0.050}$ \\\bottomrule
\end{tabular}
}
\end{table*}


\paragraph{Generalized version of geometric optimizers}
FP-BMA is a generalized version of SAM and FSAM and approximates NG under deterministic parameters, as shown in Theorem~\ref{theorem:generalized_sabma}. Proof of Theorem~\ref{theorem:generalized_sabma} is provided in Appendix~\ref{subsec:proof_of_theorem2}.

\begin{theorem}
\label{theorem:generalized_sabma}
(Informal) Suppose the model parameter $w$ is deterministic and the loss function $\ell(\cdot)$ is twice continuously differentiable. Let $\gamma^\prime = \gamma/\sqrt{\nabla_\theta \ell(\theta)^T F_\theta (\theta)^{-1} \nabla_\theta \ell(\theta)}$, then
\begin{enumerate}[label=\roman*)]
    \item FP-BMA degenerates to SAM if FIM is an identity matrix.
    \item FP-BMA degenerates to FSAM by using the diagonal terms of FIM.
    \item FP-BMA approximates the update rule of NG with learning rate $\eta_{\text{FP-BMA}} = \frac{\eta_{\text{NG}}}{(1 + \gamma^\prime)} F_\theta (\theta)^{-1}$, where $\eta_{\text{NG}}$ denotes the learning rate of NG.
\end{enumerate}
\end{theorem}

This unifying perspective implies that FP-BMA can seamlessly adapt to both deterministic and Bayesian scenarios, providing a principled way to leverage geometric properties of the loss landscape in probabilistic models. As a result, FP-BMA inherits the empirical benefits of sharpness-aware and natural gradient optimizers—such as improved generalization and robustness—while extending their applicability to Bayesian neural networks in a theoretically grounded manner.

\subsection{Flat Posterior-aware Bayesian Transfer Learning}\label{subsec:bayesian_transfer_learning}
Additionally, we extend the proposed objective to seek the flat posterior for Bayesian transfer learning. For the given approximate posterior $q^{\text{pr}}_{\theta}(w|\mathcal{D}^{\text{pr}})$ on source or downstream task $\mathcal{D}^{\text{pr}}$, we set our objective:
\begin{equation}
\label{eq:bayesian_transfer_learning}
\begin{aligned}
    \ell^\gamma_{\text{FP-BMA}}(\theta) &= \min_\theta \max_{d|\theta+\Delta\theta, \theta| \leq \gamma^2} \ell(\theta + \Delta\theta)\\
    &+ \beta \textrm{D}_{\textrm{KL}}[q_\theta (w|\mathcal{D}^{\text{ft}}) || q^{\text{pr}}_{\theta}(w|\mathcal{D}^{\text{pr}})]
\end{aligned}
\end{equation}
\begin{equation}
\label{eq:divergence_transfer}
    \textrm{s.t.} \ \  d|\theta+\Delta\theta, \theta| = \textrm{D}_{\textrm{KL}} \big[ q_{\theta+\Delta\theta}(w |\mathcal{D^{\text{ft}}}) \ || \ q_{\theta}(w | \mathcal{D}^{\text{ft}}) \big], \nonumber
\end{equation}
where $\mathcal{D}^{\text{ft}}$ is the downstream dataset. Intuitively, this objective replaces the prior distribution of Eq.~\ref{eq:divergence} by the approximate posterior $q^{\text{pr}}_{\theta}(w|\mathcal{D}^{\text{pr}})$ on source dataset. Notably, \emph{the proposed objective $\ell^\gamma_{\text{FP-BMA}}(\theta)$ can be effective in general transfer learning where the model misspecification~\citep{muller2013risk, wilson2020bayesian} exists; the prior is not suitable for the BNNs to be fine-tuned on downstream tasks, and flat parameters have been shown to improve the model's robustness~\citep{kim2022sufficient,zhang2023flatness}.}

For computational efficiency, we adopt a sub-network BNN strategy, focusing training on normalization and last-layer parameters, as explored in prior works~\citep{izmailov2020subspace, daxberger2021bayesian, sharma2023bayesian}. During fine-tuning, we reinitialize the last layer with a Gaussian distribution, $\mathcal{N}(0, \alpha I)$, where $\alpha$ is a hyperparameter to control variance. This approach ensures scalable and stable training by leveraging pre-trained DNNs. The complete FP-BMA procedure is given in Algorithm \ref{alg:FP-BMA} (Appendix \ref{subsec:algorith_of_fpbma}).

\section{Experiments}\label{sec:experiments}
\subsection{Synthetic Example}\label{subsec:synthetic_example}
We demonstrate whether the proposed FP-BMA can estimate a flat posterior when a sharp and flat minima coexists. To this end, we consider a synthetic dataset generated from true posterior having flat mode and sharp mode, as depicted in Figure~\ref{fig:posterior_approx}. This controlled setting allows us to directly observe the optimizer's preference for flat versus sharp regions in the loss landscape, isolating the effect of posterior flatness from other confounding factors. We then estimate the posterior using the proposed loss FP-BMA using SWAG. For comparison, we consider the following baseline methods: SGD, MCMC, SWAG, and VI, to estimate posterior. Figure~\ref{fig:posterior_approx} shows that MCMC, SWAG, and VI yield the posterior at sharp mode. In contrast, the proposed FP-BMA captures the flat posterior, demonstrating its effectiveness in identifying solutions with better generalization potential. We provide additional results in Appendix~\ref{subsec:synthetic_example_app}.

\begin{table*}[ht]
\caption{Downstream task performances (ACC, ECE, and NLL) with ResNet18 and ViT-B/16 pre-trained on IN 1K. \textbf{Bold} highlights the best performance within each BNN framework, while \textcolor{red}{red} indicates the overall best performance across all frameworks. FP-BMA shows superior performance both on the CIFAR10 and CIFAR100 10-shot, with the sole exception being the ECE on the CIFAR100 10-shot in ResNet18.}
\label{tab:r18_vitb16}
\adjustbox{width=1.0\linewidth}{
\begin{tabular}{ccccccccccccc}
\toprule
Backbone & \multicolumn{6}{c}{ResNet18} & \multicolumn{6}{c}{ViT-B/16} \\ \cmidrule(lr){2-7} \cmidrule(lr){8-13}   
Dataset & \multicolumn{3}{c}{CIFAR10 10-shot}        & \multicolumn{3}{c}{CIFAR100 10-shot} & \multicolumn{3}{c}{CIFAR10 10-shot}        & \multicolumn{3}{c}{CIFAR100 10-shot}      \\ \cmidrule(lr){2-4} \cmidrule(lr){5-7} \cmidrule(lr){8-10} \cmidrule(lr){11-13}
Method        & ACC $\uparrow$     & ECE $\downarrow$    & NLL $\downarrow$    & ACC $\uparrow$     & ECE $\downarrow$    & NLL $\downarrow$  & ACC $\uparrow$     & ECE $\downarrow$    & NLL $\downarrow$    & ACC $\uparrow$     & ECE $\downarrow$    & NLL $\downarrow$   \\ \midrule\midrule
SGD & $55.52_{\pm 0.32}$ & $\textbf{0.062}_{\pm 0.006}$ & $1.302_{\pm 0.020}$ & $44.29_{\pm 0.83}$ & $\textbf{0.025}_{\pm 0.005}$ & $2.133_{\pm 0.043}$ & $84.37_{\pm 1.47}$ & $0.056_{\pm 0.061}$ & $0.503_{\pm 0.038}$ & $68.78_{\pm 0.21}$ & $0.143_{\pm 0.007}$ & $1.193_{\pm 0.019}$ \\
SAM          & $56.54_{\pm 2.57}$ & $0.129_{\pm 0.013}$ & $1.354_{\pm 0.089}$ & $\textbf{44.51}_{\pm 0.07}$ & $0.065_{\pm 0.007}$ & $\textbf{2.089}_{\pm 0.013}$ & $84.35_{\pm 0.81}$ & $\textbf{0.035}_{\pm 0.012}$ & $\textbf{0.486}_{\pm 0.023}$ & $\textbf{68.93}_{\pm 0.37}$ & $0.153_{\pm 0.005}$ & $1.200_{\pm 0.021}$ \\
FSAM          & $54.04_{\pm 4.11}$ & $0.139_{\pm 0.010}$ & $1.432_{\pm 0.068}$ & $44.07_{\pm 1.21}$ & $0.056_{\pm 0.005}$ & $2.159_{\pm 0.064}$ & $\textbf{84.51}_{\pm 0.50}$ & $0.073_{\pm 0.085}$ & $0.517_{\pm 0.061}$ & $68.74_{\pm 0.39}$ & $\textbf{0.110}_{\pm 0.007}$ & $\textbf{1.166}_{\pm 0.024}$ \\
bSAM         & $\textbf{56.56}_{\pm 1.18}$ & $0.083_{\pm 0.006}$ & $\textbf{1.280}_{\pm 0.027}$ & $43.93_{\pm 0.48}$ & $0.060_{\pm 0.003}$ & $2.167_{\pm 0.026}$ & $82.85_{\pm 2.10}$ & $0.113_{\pm 0.008}$ & $0.583_{\pm 0.062}$ & $68.42_{\pm 0.40}$ & $0.148_{\pm 0.019}$ & $1.219_{\pm 0.031}$ \\ \midrule\midrule
MOPED          & $57.29_{\pm 1.20}$ & $0.093_{\pm 0.006}$ & $1.297_{\pm 0.045}$ & $44.30_{\pm 0.42}$ & $\textbf{0.047}_{\pm 0.006}$ & $2.127_{\pm 0.005}$ & $84.50_{\pm 1.36}$ & $\textbf{0.023}_{\pm 0.009}$ & $0.474_{\pm 0.038}$ & $68.80_{\pm 0.77}$ & $0.111_{\pm 0.001}$ & $1.165_{\pm 0.029}$\\
\textbf{FP-BMA (VI)}       & $\textcolor{red}{\textbf{64.98}_{\pm 1.37}}$ & $\textcolor{red}{\textbf{0.016}_{\pm 0.007}}$ & $\textcolor{red}{\textbf{0.997}_{\pm 0.046}}$ & $\textcolor{red}{\textbf{49.09}_{\pm 1.38}}$ & $0.071_{\pm 0.004}$ & $\textcolor{red}{\textbf{1.893}_{\pm 0.036}}$ & $\textcolor{red}{\textbf{87.56}_{\pm 1.10}}$ & $0.044_{\pm 0.012}$ & $\textcolor{red}{\textbf{0.397}_{\pm 0.026}}$ & $\textcolor{red}{\textbf{71.37}_{\pm 0.36}}$ & $\textbf{0.060}_{\pm 0.007}$ & $\textcolor{red}{\textbf{1.023}_{\pm 0.012}}$ \\\midrule\midrule
MCMC         & $56.31_{\pm 1.27}$ & $0.083_{\pm 0.003}$ & $1.305_{\pm 0.063}$ & $44.28_{\pm 0.95}$ & $0.021_{\pm 0.002}$ & $2.155_{\pm 0.038}$ & $83.93_{\pm 1.33}$ & $0.069_{\pm 0.010}$ & $0.523_{\pm 0.039}$ & $66.48_{\pm 1.18}$ & $0.077_{\pm 0.011}$ & $1.224_{\pm 0.044}$ \\
PTL         & $57.26_{\pm 1.44}$ & $0.116_{\pm 0.003}$ & $1.345_{\pm 0.004}$ & $43.00_{\pm 1.05}$ & $0.120_{\pm 0.006}$ & $2.383_{\pm 0.062}$  & $\textbf{85.76}_{\pm 1.37}$ & $0.080_{\pm0.014}$ & $0.482_{\pm 0.027 }$ & $65.52_{\pm 2.45 }$ & $\textbf{0.056}_{\pm 0.006 }$ & $1.260_{\pm 0.095}$ \\
E-MCMC         & $56.69_{\pm 2.14}$ & $0.142_{\pm 0.004}$ & $1.266_{\pm 0.054}$ & $41.57_{\pm 0.04}$ & $0.046_{\pm 0.012}$ & $2.370_{\pm 0.175}$ & $83.91_{\pm 1.16}$ & $0.333_{\pm 0.010}$ & $0.877_{\pm 0.044}$ & $63.40_{\pm 0.01}$ & $0.280_{\pm 0.008}$ & $1.655_{\pm 0.024}$ \\
\textbf{FP-BMA (MCMC)}  & $\textbf{57.49}_{\pm 0.64}$ & $\textbf{0.039}_{\pm 0.00.}$ & $\textbf{1.248}_{\pm 0.048}$ & $\textbf{45.72}_{\pm 0.56}$ & $\textcolor{red}{\textbf{0.016}_{\pm 0.003}}$ & $\textbf{2.062}_{\pm 0.050}$ & $84.82_{\pm 1.84}$ & $\textbf{0.051}_{\pm 0.018}$ & $\textbf{0.449}_{\pm 0.048}$ & $\textbf{68.73}_{\pm 1.09}$ & $0.061_{\pm 0.004}$ & $\textbf{1.117}_{\pm 0.042}$ \\ \midrule\midrule
SWAG  & $56.31_{\pm 0.60}$ & $0.094_{\pm 0.013}$ & $1.315_{\pm 0.056}$ & $44.14_{\pm 1.28}$ & $\textbf{0.034}_{\pm 0.010}$ & $2.161_{\pm 0.058}$ & $83.51_{\pm 2.22}$ & $0.022_{\pm 0.015}$ & $0.510_{\pm 0.072}$ & $68.72_{\pm 0.45}$ & $0.065_{\pm 0.005}$ & $1.136_{\pm 0.014}$  \\
F-SWAG & $57.65_{\pm 1.20}$ & $0.075_{\pm 0.003}$ & $1.249_{\pm 0.038}$ & $46.09_{\pm 0.44}$ & $0.062_{\pm 0.006}$ & $2.089_{\pm 0.002}$ & $83.87_{\pm 1.28}$ & $0.013_{\pm 0.005}$ & $0.492_{\pm 0.040}$ & $68.84_{\pm 0.77}$ & $0.076_{\pm 0.012}$ & $1.137_{\pm 0.020}$ \\
\textbf{FP-BMA (SWAG)}       & $\textbf{61.79}_{\pm 4.34}$ & $\textbf{0.026}_{\pm 0.004}$ & $\textbf{1.214}_{\pm 0.119}$ & $\textbf{47.45}_{\pm 0.60}$ & $0.055_{\pm 0.018}$ & $\textbf{2.044}_{\pm 0.022}$ & $\textbf{86.81}_{\pm 0.78}$ & $\textcolor{red}{\textbf{0.010}_{\pm 0.003}}$ & $\textbf{0.399}_{\pm 0.034}$ & $\textbf{70.10}_{\pm 0.18}$ & $\textcolor{red}{\textbf{0.045}_{\pm 0.015}}$ & $\textbf{1.063}_{\pm 0.023}$\\ \bottomrule
\end{tabular}
}
\end{table*}


\begin{table*}[ht]
\caption{Downstream task accuracy with ResNet50 and ViT-B/16 pre-trained on IN 1K. \textbf{Bold} and \underline{underline} denote best and second best performance each. FP-BMA demonstrates superior performance across all 16-shot datasets.}
\label{tab:fine-grained}
\adjustbox{width=1.0\linewidth}{
\begin{tabular}{cccccacccca}
\toprule
Backbone &
  \multicolumn{5}{c}{ResNet50} &
  \multicolumn{5}{c}{ViT-B/16} \\ \cmidrule(lr){2-6} \cmidrule(lr){7-11}
Method &
  EuroSAT &
  Flowers102 &
  Pets &
  UCF101 &
  Avg &
  EuroSAT &
  Flowers102 &
  Pets &
  UCF101 &
  Avg \\ \midrule
SGD &
  $86.75_{\pm 1.47}$ &
  $93.16_{\pm 0.27}$ &
  $89.95_{\pm 0.51}$ &
  $66.34_{\pm 0.59}$ &
  $84.05_{\pm 0.33}$ &
  $81.25_{\pm 1.03}$ &
  $91.24_{\pm 0.83}$ &
  $88.68_{\pm 0.92}$ &
  $68.64_{\pm 0.51}$ &
  $82.45_{\pm 0.56}$ \\
SAM &
  $87.85_{\pm 0.49}$ &
  $94.80_{\pm 0.17}$ &
  $90.23_{\pm 0.78}$ &
  $70.40_{\pm 0.76}$ &
  $85.82_{\pm 0.25}$ &
  $82.53_{\pm 0.65}$ &
  $\underline{93.08}_{\pm 0.87}$ &
  $\underline{90.66}_{\pm 0.74}$ &
  $\underline{70.66}_{\pm 1.03}$ &
  $\underline{84.23}_{\pm 0.60}$ \\
SWAG &
  $88.97_{\pm 1.56}$ &
  $93.27_{\pm 0.15}$ &
  $89.95_{\pm 0.46}$ &
  $66.41_{\pm 0.30}$ &
  $84.65_{\pm 0.37}$ &
  $81.62_{\pm 0.66}$ &
  $91.21_{\pm 0.91}$ &
  $88.67_{\pm 0.42}$ &
  $67.65_{\pm 0.45}$ &
  $82.29_{\pm 0.31}$ \\
F-SWAG &
  $90.03_{\pm 1.08}$ &
  $\underline{94.84}_{\pm 0.26}$ &
  $\underline{90.12}_{\pm 0.57}$ &
  $\underline{70.00}_{\pm 0.87}$ &
  $\underline{86.25}_{\pm 0.19}$ &
  $82.72_{\pm 0.49}$ &
  $92.93_{\pm 0.93}$ &
  $90.60_{\pm 0.55}$ &
  $68.67_{\pm 0.39}$ &
  $83.73_{\pm 0.35}$ \\
MOPED &
  $85.21_{\pm 3.14}$ &
  $92.15_{\pm 0.73}$ &
  $89.25_{\pm 0.61}$ &
  $65.85_{\pm 0.99}$ &
  $83.11_{\pm 0.86}$ &
  $\underline{83.97}_{\pm 0.49}$ &
  $91.71_{\pm 0.87}$ &
  $89.90_{\pm 0.54}$ &
  $69.66_{\pm 0.53}$ &
  $83.81_{\pm 0.51}$ \\
PTL &
  $\underline{90.01}_{\pm 0.39}$ &
  $92.55_{\pm 0.53}$ &
  $89.43_{\pm 0.41}$ &
  $65.00_{\pm 1.24}$ &
  $84.25_{\pm 0.30}$ &
  $83.76_{\pm 0.61}$ &
  $88.43_{\pm 1.27}$ &
  $88.54_{\pm 0.53}$ &
  $60.38_{\pm 1.84}$ &
  $80.28_{\pm 0.03}$ \\
FP-BMA &
  $\boldsymbol{90.16}_{\pm 1.04}$ &
  $\boldsymbol{95.85}_{\pm 1.26}$ &
  $\boldsymbol{90.23}_{\pm 0.58}$ &
  $\boldsymbol{71.57}_{\pm 0.27}$ &
  $\boldsymbol{86.95}_{\pm 0.65}$ &
  $\boldsymbol{84.60}_{\pm 0.25}$ &
  $\boldsymbol{94.15}_{\pm 0.80}$ &
  $\boldsymbol{91.30}_{\pm 0.25}$ &
  $\boldsymbol{72.63}_{\pm 1.12}$ &
  $\boldsymbol{85.67}_{\pm 0.14}$ \\ \bottomrule
\end{tabular}
}
\vspace{-1.5em}
\end{table*}


\subsection{Learning from scratch}\label{subsec:learning_from_scratch}
We verify the effectiveness of FP-BMA in improving the performance of BNNs trained from scratch. Specifically, we use Bayesian ResNet18 and a modified ViT-B/16$^{\dagger}$~\citep{dosovitskiy2020image, liu2021efficient, zhu2023understanding} on CIFAR10 and CIFAR100. We adopt the modified ViT-B/16$^{\dagger}$ to address the underfitting issue of ViTs on small datasets. Due to computational constraints in large-scale models, we apply variational distributions to the parameters of normalization and last layers. We then train these variational parameters using approximate Bayesian inference (VI, MCMC, and SWAG) with the gradient $\nabla_\theta \ell^\gamma_{\text{FP-BMA}}(\theta)$ in Eq.~\ref{eq:update_sabma}, while updating the remaining parameters using the gradient $\nabla_\theta \ell(\theta)$. This setup allows us to assess the benefits of FP-BMA in both convolutional and transformer-based architectures under realistic training constraints.

For comparison, we consider SGD, SAM~\citep{foret2020sharpness}, and FSAM~\citep{kim2022fisher} seeking flat minima in DNNs. For the training of BNNs, we consider SWAG, VI, F-SWAG~\citep{nguyen2023flat}, bSAM~\citep{mollenhoff2022sam}, and E-MCMC~\citep{li2023entropy}, which utilizes SGLD. For fair comparison, we use the same BNN architecture employed for FP-BMA. All baseline methods are carefully tuned with respect to their key hyperparameters to ensure a fair and meaningful comparison of generalization performance, and the detailed hyperparameter configurations for each baseline are provided in Appendix~\ref{subsubsec:hyperparameters_for_experiments_1}.

Table~\ref{tab:scratch_r18_vitb16} showcases the generalization performance, including accuracy (ACC), ECE, and NLL. The FP-BMA consistently improves performances when integrated with VI, MCMC, and SWAG. Also, The FP-BMA  leads to superior performances compared to other baselines of SGD, SAM, FSAM, and bSAM. Additional experimental details are provided in Appendix~\ref{subsec:learning_from_scratch_app}.

\subsection{Bayesian Transfer Learning}\label{subsec:few-shot_image_classification}

\paragraph{Finetuning on CIFARs}
We validate the effectiveness of the FP-BMA on a transfer learning task. We first adopt RN18 and ViT-B/16 pre-trained on ImageNet (IN) 1K~\citep{russakovsky2015imagenet} as a backbone. The pre-trained models are fine-tuned on CIFAR10 and CIFAR100 10-shot, using 10 data instances per class.

For comparison, we consider the following Bayesian transfer learning methods: MOPED~\citep{krishnan2020specifying} and Pre-Train Your Loss (PTL)~\citep{shwartz2022pre}. We describe additional configurations in Appendix~\ref{subsec:few_shot_image_classification_with_bayesian_transfer_learning_app}.

Table~\ref{tab:r18_vitb16} shows 
FP-BMA with diverse BNN frameworks consistently outperforms existing baselines in terms of both accuracy and uncertainty quantification. Unlike scratch learning, FP-BMA (VI) outperforms FP-BMA (SWAG) in few-shot image classification tasks. This can be attributed to the nature of few-shot tasks, where VI, which only learns a diagonal covariance, is less prone to underfitting due to the limited amount of data.

\begin{table*}[t]
\caption{Downstream task accuracy of CLIP with visual encoder, ResNet50 and ViT-B/16. \textbf{Bold} and \underline{underline} denote best and second best performance each. FP-BMA shows superior performance in average over five datasets.}
\label{tab:IN_variants}
\adjustbox{width=1.0\linewidth}{
\begin{tabular}{ccccccaccccca}
\toprule
Backbone & \multicolumn{6}{c}{ResNet50} & \multicolumn{6}{c}{ViT-B/16} \\ \cmidrule(lr){2-7} \cmidrule(lr){8-13} 
Method & IN & IN-V2 & IN-R & IN-A & IN-S & Avg  & IN & IN-V2 & IN-R & IN-A & IN-S & Avg \\ \midrule
Zero-Shot     & $59.83_{\pm 0.00}$ & $52.89_{\pm 0.00}$    & $60.73_{\pm 0.00}$   & $\underline{23.25}_{\pm 0.00}$   & $35.45_{\pm 0.00}$   & $46.43_{\pm 0.00}$  & $68.33_{\pm 0.00}$ & $61.91_{\pm 0.00}$    & $77.71_{\pm 0.00}$   & $49.93_{\pm 0.00}$   & $48.22_{\pm 0.00}$   & $61.22_{\pm 0.00}$ \\
SGD    & $61.70_{\pm 0.01}$ & $54.31_{\pm 0.01}$    & $60.87_{\pm 0.01}$   & $22.74_{\pm 0.01}$   & $\underline{35.68}_{\pm 0.00}$   & $47.06_{\pm 0.01}$  & $69.97_{\pm 0.00}$ & $62.97_{\pm 0.01}$    & $78.05_{\pm 0.00}$   & $50.31_{\pm 0.02}$   & $48.76_{\pm 0.00}$   & $62.01_{\pm 0.00}$ \\
SAM    & $61.73_{\pm 0.01}$ & $\underline{54.35}_{\pm 0.01}$    & $60.86_{\pm 0.01}$   & $22.76_{\pm 0.01}$   & $35.67_{\pm 0.00}$   & $47.07_{\pm 0.01}$  & $70.01_{\pm 0.01}$ & $63.03_{\pm 0.02}$    & $78.03_{\pm 0.01}$   & $50.37_{\pm 0.00}$   & $48.75_{\pm 0.00}$   & $62.04_{\pm 0.00}$ \\
SWAG     & $\underline{61.77}_{\pm 0.22}$ & $54.10_{\pm 0.19}$    & $\textbf{61.25}_{\pm 0.21}$   & $\textbf{23.25}_{\pm 0.08}$   & $35.55_{\pm 0.27}$   & $\underline{47.18}_{\pm 0.19}$  & $\underline{70.11}_{\pm 0.02}$ & $\underline{63.44}_{\pm 0.06}$    & $\textbf{78.33}_{\pm 0.03}$   & $\textbf{50.55}_{\pm 0.02}$   & $\underline{48.95}_{\pm 0.01}$   & $\underline{62.28}_{\pm 0.02}$ \\
\textbf{FP-BMA} & $\textbf{63.33}_{\pm 0.92}$ & $\textbf{55.06}_{\pm 0.79}$    & $\underline{61.14}_{\pm 0.37}$   & $22.78_{\pm 0.68}$   & $\textbf{35.82}_{\pm 0.11}$   & $\textbf{47.63}_{\pm 0.17}$ & $\textbf{72.41}_{\pm 0.33}$ & $\textbf{64.85}_{\pm 0.11}$    & $\underline{78.14}_{\pm 0.31}$   & $\underline{50.52}_{\pm 0.25}$   & $\textbf{49.25}_{\pm 0.03}$   & $\textbf{63.03}_{\pm 0.04}$ \\ \bottomrule
\end{tabular}
}
\end{table*}


\begin{figure*}[ht]
\centering
\includegraphics[width=\textwidth]{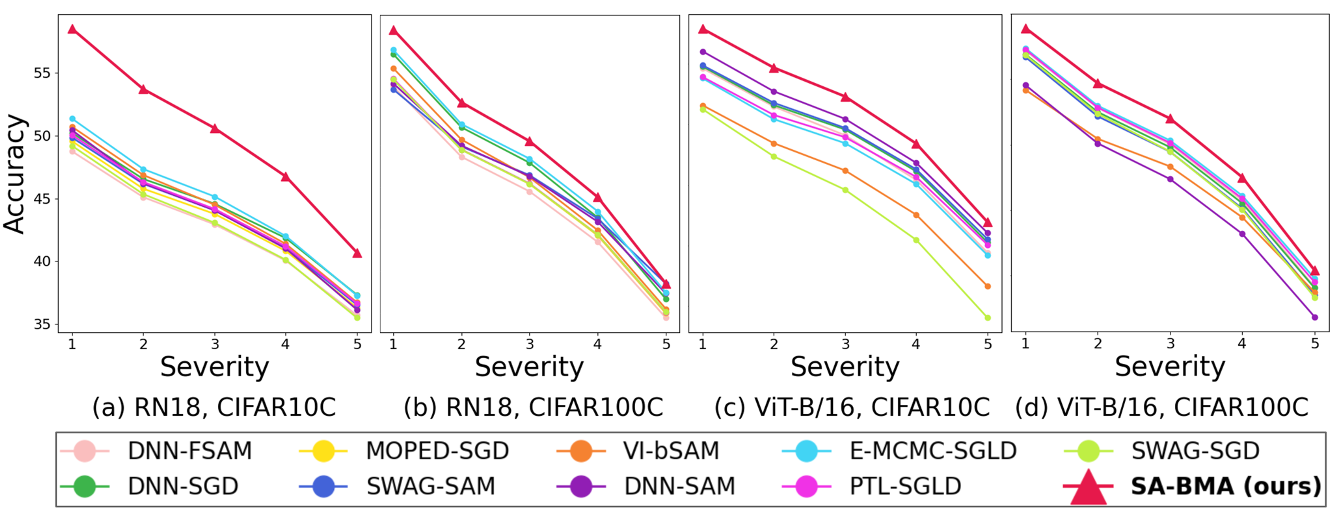}
\caption{Accuracy under distributional shift. We evaluate the accuracy of RN18 and ViT-B/16 models trained on CIFAR10 and CIFAR100 10-shot across all severity levels of CIFAR10C and CIFAR100C. FP-BMA consistently outperforms all baseline methods across all levels of corruption.}  
\label{fig:distribution_shift}
\end{figure*}

\paragraph{Fine-tuning on fine-grained image classification tasks}
Furthermore, we confirm the effectiveness of FP-BMA on general fine-grained image classification tasks, including EuroSAT~\citep{helber2019eurosat}, Flowers102~\citep{nilsback2008automated}, Pets~\citep{parkhi2012cats}, and UCF101~\citep{soomro2012ucf101}. All experiments were conducted using a 16-shot setting across all datasets. From this point forward, we perform all experiments using FP-BMA with SWAG only.

Table~\ref{tab:fine-grained} shows that the FP-BMA achieves the best accuracy. Table~\ref{tab:fine-grained_nll} (Appendix~\ref{subsec:fine-grained_image_classification_app}) shows that the FP-BMA achieves the best NLL, as well. This implies that FP-BMA seeking the flat posterior during fine-tuning procedure is effective in improving the performance of Bayesian transfer learning.

\paragraph{Fine-tuning with CLIP}
We also show the effectiveness of FP-BMA on the pre-trained vision language models.  We fine-tune only the last layer of the CLIP visual encoder on the IN 1K 16-shot dataset. Then, we evaluate the trained model on IN and its variants—IN-V2~\citep{recht2019imagenet}, IN-R~\citep{hendrycks2021many}, IN-A~\citep{hendrycks2021natural}, and IN-S~\citep{wang2019learning}—following the protocols outlined in~\citet{radford2021learning, zhu2023enhancing}.

Table~\ref{tab:IN_variants} shows that FP-BMA outperforms baselines on IN set. Also, FP-BMA shows superior or comparable accuracy on out-of-distribution datasets, representing the effectiveness of robustness.

\subsection{Robustness on Distribution Shift}\label{subsec:robustness_on_distribution_shift}
We evaluate the trained models on CIFAR10 and CIFAR100 10-shots using the corrupted datasets CIFAR10C and CIFAR100C~\citep{hendrycks2019benchmarking} to demonstrate the robustness of FP-BMA. These benchmarks simulate a wide variety of real-world corruptions, including noise, blur, weather, and digital effects, thereby providing a comprehensive testbed for evaluating model reliability under distribution shift.

Figure~\ref{fig:distribution_shift} presents the accuracy on the corrupted datasets CIFAR10C and CIFAR100C~\citep{hendrycks2019benchmarking}, demonstrating that FP-BMA outperforms baselines on corrupted datasets across all corruption levels. FP-BMA consistently outperforms all baselines in NLL, as shown in Figure~\ref{fig:severity_nll}. Detailed results are provided in Appendix~\ref{subsec:performance_under_distribution_shift_app}.

The results on IN variants in Table~\ref{tab:IN_variants} and the corrupted datasets in Figure~\ref{fig:distribution_shift} show that FP-BMA enhances the robustness of trained BNNs under distribution shifts, suggesting that the Flat Posterior-aware Bayesian Transfer Learning scheme with FP-BMA effectively improves robustness.

\subsection{Flatness Analysis}\label{subsec:flatness_analysis}
\begin{figure*}[ht]
\captionsetup{skip=0pt}
\centering
\includegraphics[width=\textwidth]{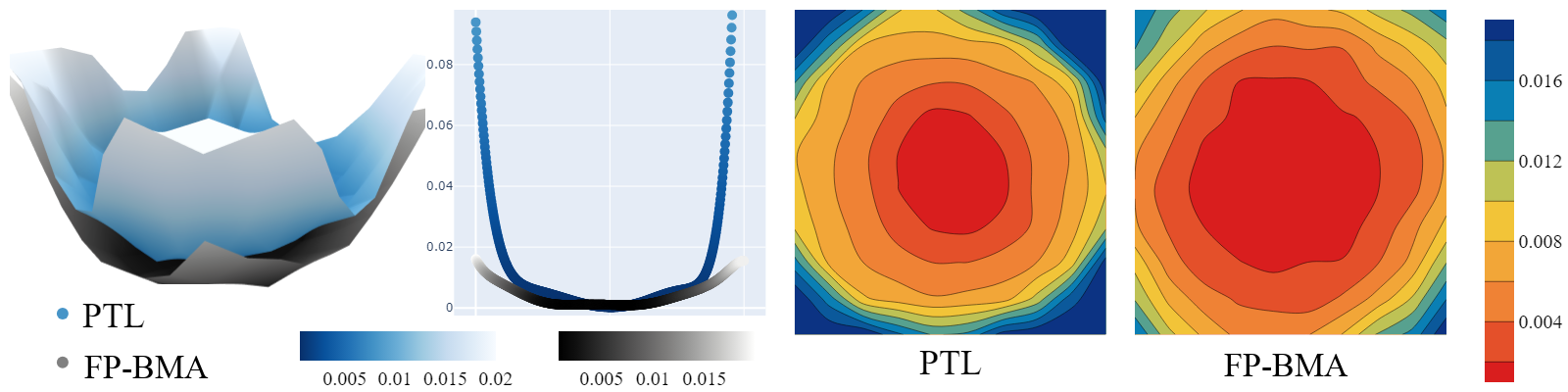} 
\caption{Comparison of the loss surfaces of FP-BMA (grey) and PTL (light blue) models. The comparison of loss surface shows that FP-BMA allows the posterior to be placed on a lower and flatter loss surface compared to that of PTL.}
\label{fig:loss_surface}
\end{figure*}

\begin{table*}[t]
\centering
\caption{Hessian analysis on ResNet18 trained with CIFAR10 10-shot. FP-BMA shows the lowest score on both $\lambda_1$ and $\lambda_1/\lambda_5$, proving it leads the model to flatter minima.}
\adjustbox{width=1.0\linewidth}{
\begin{tabular}{ccccccccccc}
\toprule
& SGD & SAM & FSAM & bSAM & MOPED & PTL & E-MCMC & SWAG & F-SWAG & \textbf{FP-BMA} \\ \midrule
$\lambda_1 \downarrow$ & 559.62 & 381.74 & 561.15 & 532.74 & 686.90 & 559.16 & 540.83 & 602.34 & 362.33 & \textbf{275.21} \\
$\lambda_1 / \lambda_5 \downarrow$ & 2.59 & 2.23 & 2.24 & 2.09 & 2.41 & 2.23 & 1.98 & 2.13 & 2.44 & \textbf{1.69} \\
\bottomrule
\end{tabular}
}
\label{tab:hessian}
\end{table*}

We analyze whether FP-BMA encourages the posterior of BNNs to lie in a flatter loss basin. Using ResNet18 trained on CIFAR10 with 10-shot, we compare weight samples from the approximate posterior obtained by FP-BMA and PTL and compare the Hessian's eigenvalue of model.

Figure~\ref{fig:loss_surface} presents different views of loss surface using sampled weights of FP-BMA and PTL. This result confirms that the posterior of FP-BMA is placed on a flatter loss basin with lower loss. Additional results and the protocol to visualize the loss basin are provided in Appendix~\ref{subsec:loss_surface_replic}.

Table~\ref{tab:hessian} compares the Hessian's eigenvalue of model $\lambda_i$ (Eq.~\ref{eq:bma_hessian}) where
$\lambda_1$ and $\lambda_5$ represent the largest eigenvalue and the fifth largest eigenvalue, respectively. This result indicates that FP-BMA achieves the lowest values compared to all baselines, implying that the posterior of BNNs is formed on the flattest local surface. This further supports our empirical observations that FP-BMA enhances generalization by encouraging a flatter posterior distribution.

\section{Related Works}\label{sec:related_works}
\subsection{Flatness and BNN}\label{subsec:flatness_and_bnn}
Recent works have suggested flat-seeking optimizers combined with BNN. First, SWAG~\citep{maddox2019simple} implicitly approximated posterior toward flatter optima based on SWA~\citep{izmailov2018averaging}. However, SWAG can fail to find flat minima, leading to limited improvement in generalization, as shown in Section~\ref{subsec:insufficient_flatness_of_bma}. bSAM~\citep{mollenhoff2022sam} showed that SAM can be interpreted as a relaxation of the Bayes and quantified uncertainty with SAM. Yet, bSAM only focused on uncertainty quantification by simply modifying Adam-based SAM~\citep{khan2018fast}, not newly considering the parametric geometry for perturbation. Moreover, scaling the variance with the number of data points hampers the direct implementation of bSAM in few-shot settings. SA-BNN~\citep{nguyen2023flat} proposed a sharpness-aware posterior derived directly from the variational objective and proved the effectiveness experimentally and theoretically. However, they simply employ the L2 norm to calculate the perturbation of SAM without considering the difference between the nature of DNN and BNN. Moreover, in contrast to FP-BMA, SA-BNN did not take into account the prior, which is a fundamental component of BNNs, in its pursuit of flatness. On the other hand, E-MCMC~\citep{li2023entropy} proposed an efficient MCMC algorithm capable of effectively sampling the posterior within a flat basin by removing the nested chain of Entropy-SGD~\citep{dziugaite2018entropy} and Entropy-SGLD~\citep{chaudhari2019entropy}. However, E-MCMC necessitates a guidance model, which doubles the parameters and heavily hinders its employment over large-scale models. FP-BMA is the first approach to explicitly promote flat posteriors within a rigorous Bayesian framework, providing a principled way to enhance robustness and generalization.

\subsection{Bayesian Transfer Learning}\label{subsec:bayesian_transfer_learning_related}
Applying Bayesian methods to transfer learning is a natural and theoretically well-founded approach, as the Bayesian framework systematically incorporates prior knowledge and quantifies uncertainty when adapting models to new tasks. Theoretical foundations for this perspective can be found in the literature on probabilistic machine learning and hierarchical Bayesian models~\citep{bishop2006pattern, murphy2012machine, gelman1995bayesian}, as well as early works on Bayesian transfer and domain adaptation~\citep{lawrence2004learning, raina2006constructing}. Building on these principles, a variety of Bayesian transfer learning methods have been developed, including approaches leveraging pre-trained models as priors, empirical Bayes techniques, and flexible posterior approximations~\citep{krishnan2020specifying, shwartz2022pre, lee2024enhancing}.
PTL~\citep{shwartz2022pre} constructs BNN by learning closed-form posterior approximation of the pre-trained model on the source task and uses it as a prior for the downstream task after scaling. The work requires additional training on the source task, making it restrictive when accessing the source task dataset is impossible. MOPED~\citep{krishnan2020specifying} employs pre-trained BNN as a prior for VI based on the empirical Bayes method. Using pre-trained DNN, MOPED enhances accessibility to BNN; however, it is only applicable to Mean-field VI. Non-parametric transfer learning~\citep{lee2024enhancing} suggested adopting non-parametric learning to make posterior flexible in terms of distribution shift. The proposed Flat Posterior-aware Bayesian Transfer Learning utilizes a pre-trained model as a prior, improving robustness to model misspecification by promoting a flat posterior.

\section{Conclusion}\label{sec:conclusion}


This study demonstrates the limitations of BNNs in capturing flatness—a property crucial for generalization—and reveals that BMA may fail to yield optimal results without considering flatness. To address this, we introduce FP-BMA, which seeks a flat posterior by effectively capturing flatness in the parameter space. FP-BMA generalizes existing sharpness-aware optimizers and aligns with the intrinsic nature of BNNs. We further propose a Flat Posterior-aware Bayesian Transfer Learning scheme, which enhances resilience against model misspecification. Our extensive experiments demonstrate that FP-BMA significantly improves the generalization ability of BNNs, underscoring the importance of flatness in posterior approximations. However, there are several limitations to our study. Specifically, our theoretical insights rely on strong assumptions, and the empirical evaluation does not cover the full spectrum of MCMC algorithms. Future work could extend FP-BMA to a wider variety of Bayesian inference methods and investigate its
effectiveness on more complex datasets. Additionally, exploring automated ways to quantify and enforce flatness during model training could further enhance the robustness and applicability of the proposed approach.

\clearpage
\section*{Acknowledgement}
This work was supported by the National Research Foundation of Korea(NRF) grant funded by the Korea government(MSIT)(RS-2024-00457216).





\bibliography{main}

\newpage
\onecolumn

\makeatletter
\renewcommand{\@thanks}{}
\makeatother

\title{Flat Posterior Does Matter For Bayesian Model Averaging\\(Appendix)}
\maketitle

\appendix
\section{Flatness Does Matter For Bayesian Model Averaging}\label{sec:flatness_does_matter_for_bma_app}

\subsection{Need For Flatness In BMA}\label{subsec:need_for_flatness_in_bma_app}
\paragraph{Experimental Details}
To measure the flatness of BNNs, $M$ of Eq.~\ref{eq:bma_hessian} is set to 30 for experiments in Section~\ref{subsec:need_for_flatness_in_bma}. We primarily use RN18 as the backbone. Our evaluation includes Error ($100 - \text{Accuracy}$), Expected Calibration Error (ECE)~\citep{guo2017calibration}, and Negative Log-Likelihood (NLL) to assess generalization on CIFAR10 and CIFAR100. To minimize confounding effects on flatness measurements, we do not adjust BN and data augmentation. For BNN frameworks, we consider VI, MCMC, and SWAG. We also consider three different learning rate scheduler: Constant, Cosine Decay (Cos Deacy), and SWAG learning rate (SWAG lr).

\subsubsection{Correlation Between Flatness And Generalization}\label{subsubsec:correlation_between_flatness_and_generalization}

\begin{figure}[th]
  \centering
  \begin{subfigure}{0.33\textwidth}
    \centering
    \includegraphics[width=\linewidth]{figure/corr/cifar10_swag-sgd-constant_corr.png}
    \caption{Constant}
  \end{subfigure}%
  \begin{subfigure}{0.33\textwidth}
    \centering
    \includegraphics[width=\linewidth]{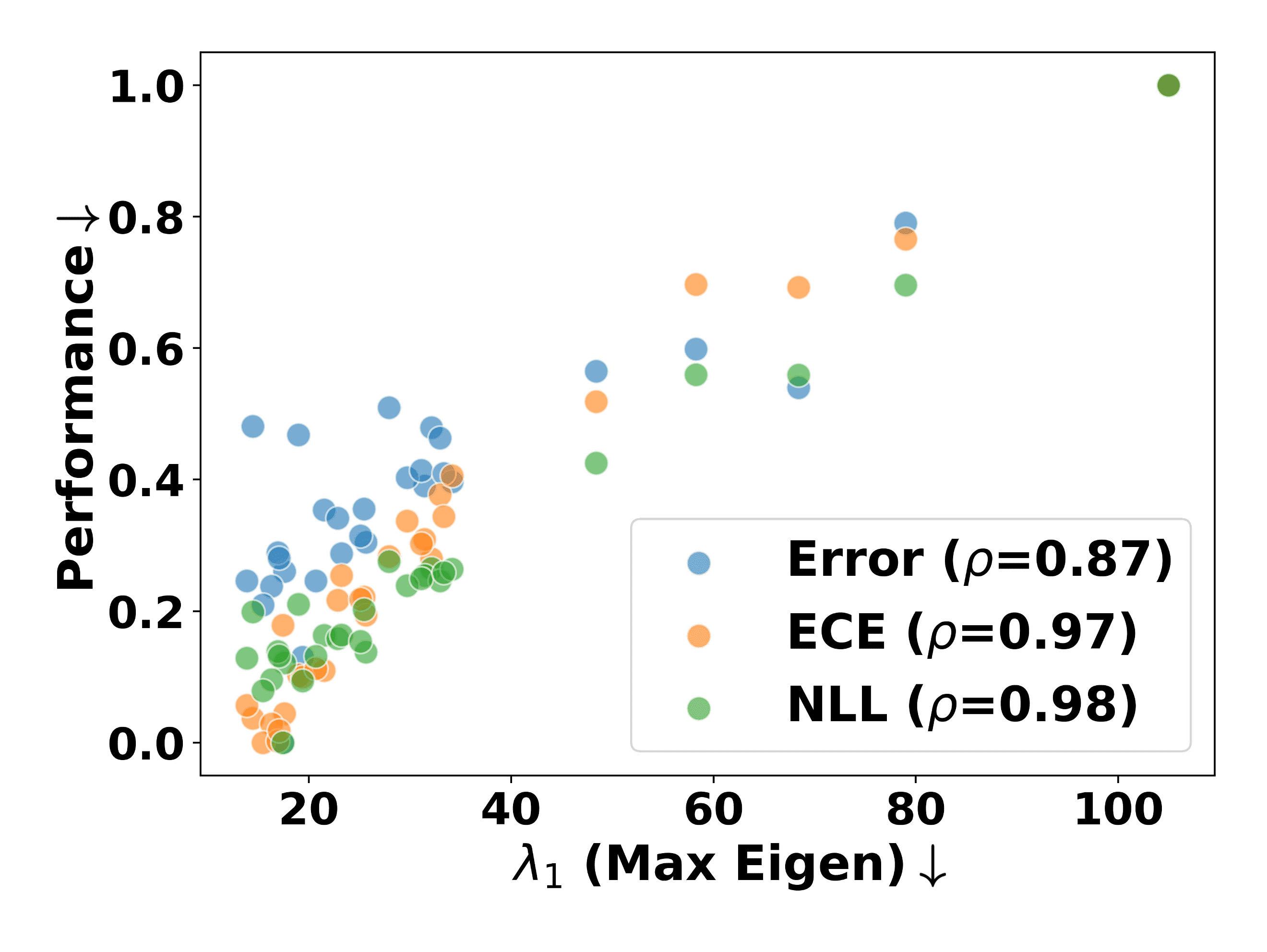}
    \caption{Cosine Decay}
  \end{subfigure}%
  \begin{subfigure}{0.33\textwidth}
    \centering
    \includegraphics[width=\linewidth]{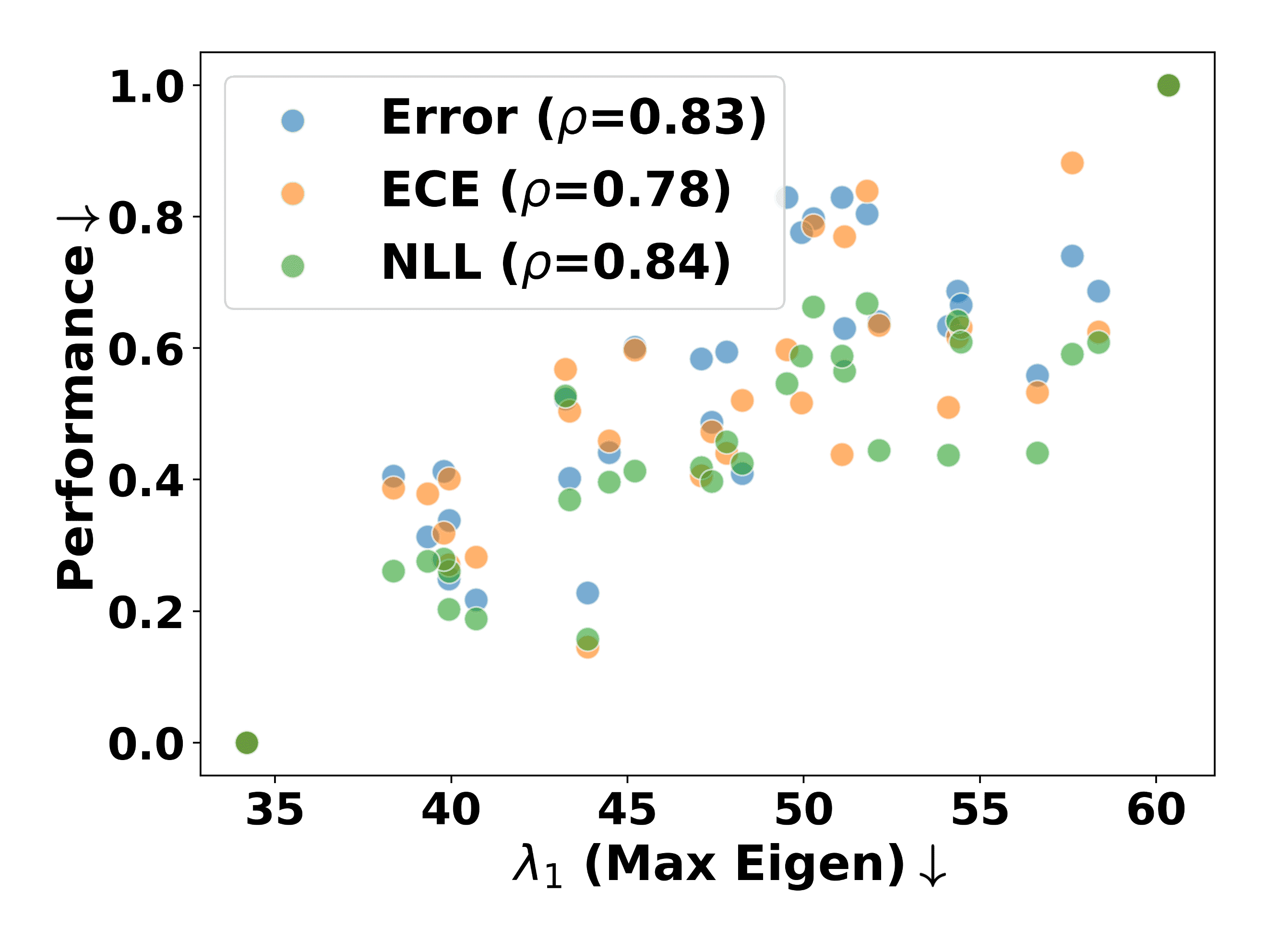}
    \caption{SWAG lr}
  \end{subfigure}\\[1ex]
  \begin{subfigure}{0.33\textwidth}
    \centering
    \includegraphics[width=\linewidth]{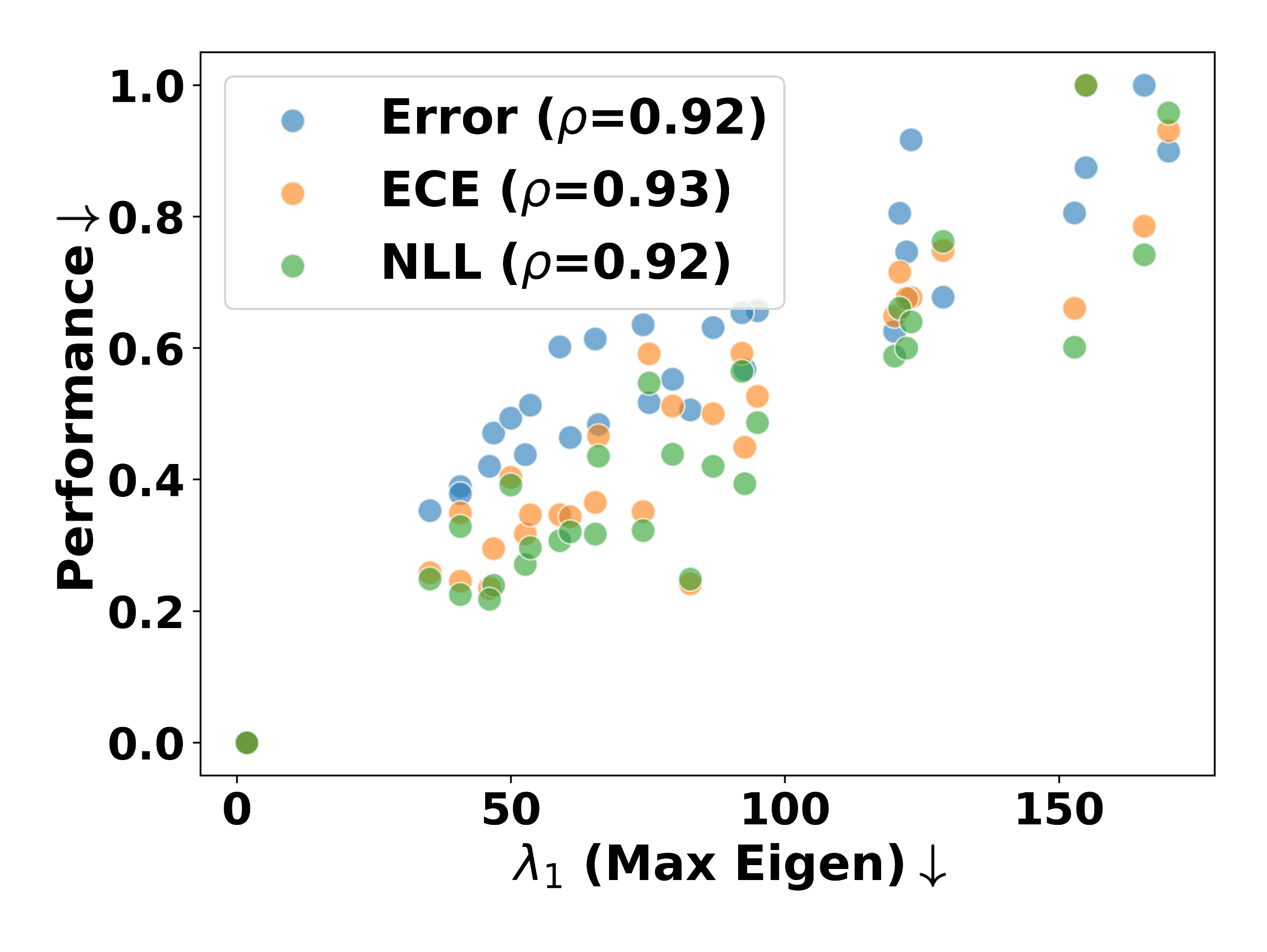}
    \caption{Constant}
  \end{subfigure}%
  \begin{subfigure}{0.33\textwidth}
    \centering
    \includegraphics[width=\linewidth]{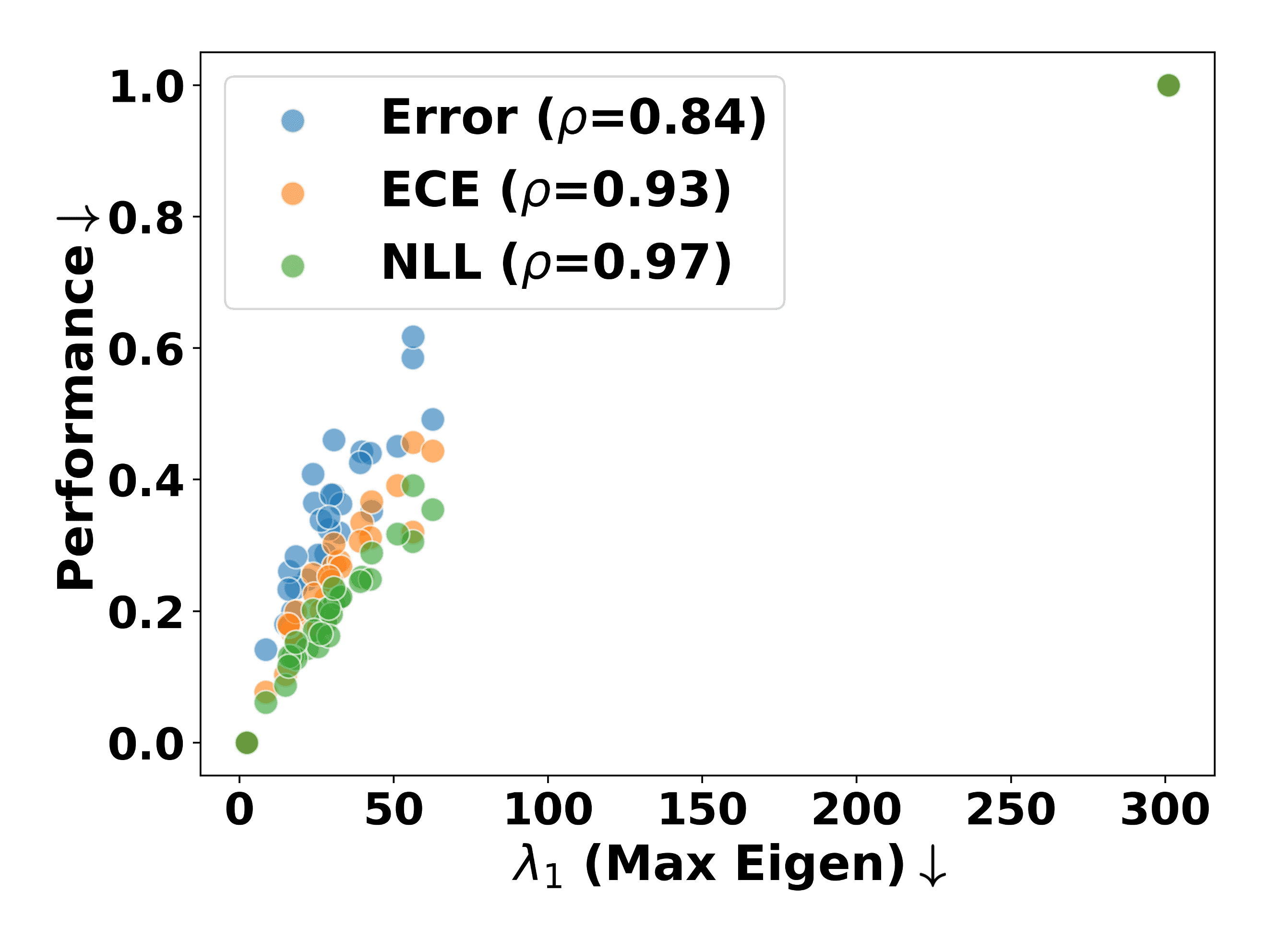}
    \caption{Cosine Decay}
  \end{subfigure}%
  \begin{subfigure}{0.33\textwidth}
    \centering
    \includegraphics[width=\linewidth]{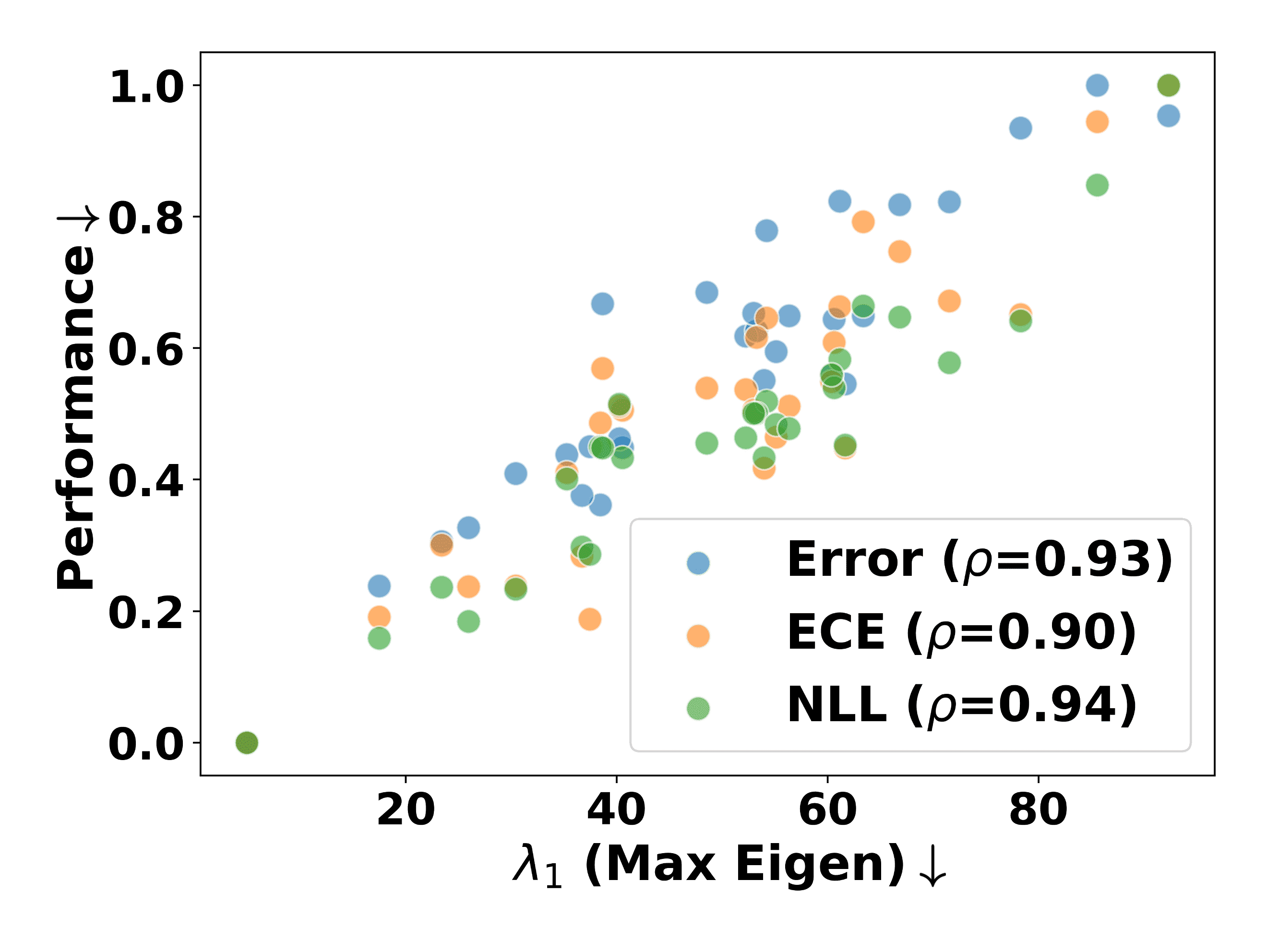}
    \caption{SWAG lr}
  \end{subfigure}
  \caption{Correlation between maximal eigenvalue and performances of 30 sampled models from SWAG throughout all considered schedulers. It shows classification error, ECE, and NLL are distinctly correlated with flatness. We conjecture that the flatness is crucial for the generalization performance of BNN}
  \label{fig:additional_corr_plot}
\end{figure}

We check the correlation between flatness and generalization performance of sampled models throughout all considered learning rate schedulers. We present the scatter plot of the model, sampled from ResNet18 trained on CIFAR10 and CIFAR100 in the first and second rows of Figure~\ref{fig:additional_corr_plot}. Each column of Figure~\ref{fig:additional_corr_plot} denotes Constant scheduler, Cosine Decay scheduler, and SWAG lr scheduler, respectively. All the models are trained with SWAG and SGD momentum, and we set maximal eigenvalue $\lambda_1$ as a flatness measure. Correlation with flatness and each generalization performance metric is suggested in the legend, as well. Regardless of the scheduler and dataset, all generalization performances, error, ECE, and NLL strongly correlate with flatness.

\subsection{Insufficient Flatness of BMA}\label{subsec:insufficient_flatness_of_bma_app}
Figure~\ref{fig:sgd_to_fpbma_cifar10} shows results consistent with Figure~\ref{fig:sgd_to_fpbma} across various learning rate schedulers and metrics. Specifically, (1) BNNs struggle to ensure flatness compared to DNNs when using SGD, and (2) the proposed FP-BMA enables BNN frameworks to achieve flat minima, thereby enhancing performance.

\begin{figure}[ht]
\captionsetup{skip=0pt}
\centering
\includegraphics[width=0.75\textwidth]{figure/sgd_to_sabma/sgd_to_fpbma_cifar10.png} 
\caption{Comparison of Error, NLL, and ECE with various schedulers on CIFAR10 in relation to the maximum eigenvalue $\lambda_1$.}
\label{fig:sgd_to_fpbma_cifar10}
\end{figure}


\clearpage
\subsection{Performance Changes Based On The Number of Models In BMA}\label{subsec:performance_changes_based_on_the_number_of_models_in_bma}
\begin{figure}[h]
  \centering
  \begin{subfigure}{0.33\textwidth}
    \centering
    \includegraphics[width=\linewidth]{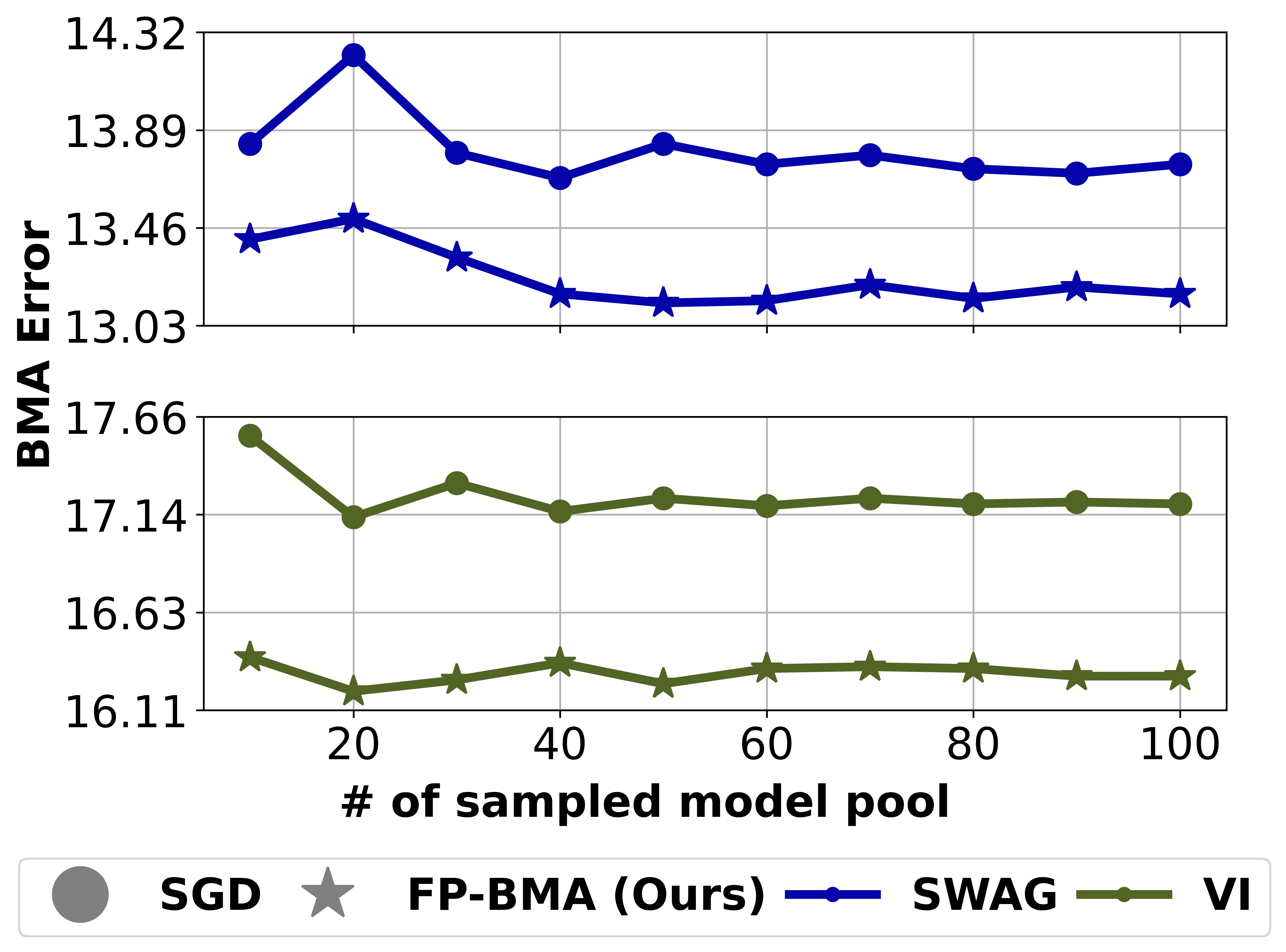}
    \caption{Constant - Error}
  \end{subfigure}%
  \begin{subfigure}{0.33\textwidth}
    \centering
    \includegraphics[width=\linewidth]{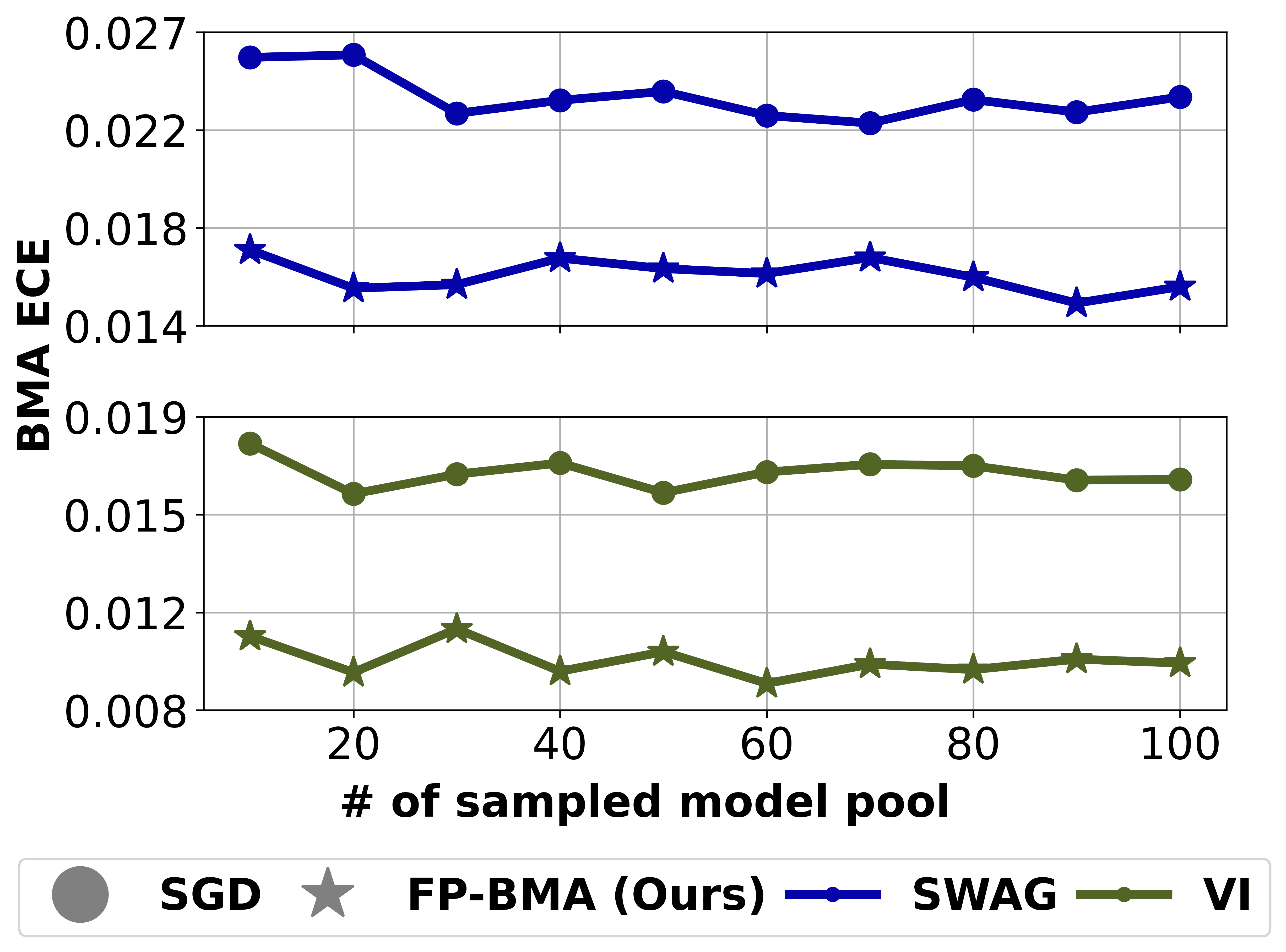}
    \caption{Constant - ECE}
  \end{subfigure}%
  \begin{subfigure}{0.33\textwidth}
    \centering
    \includegraphics[width=\linewidth]{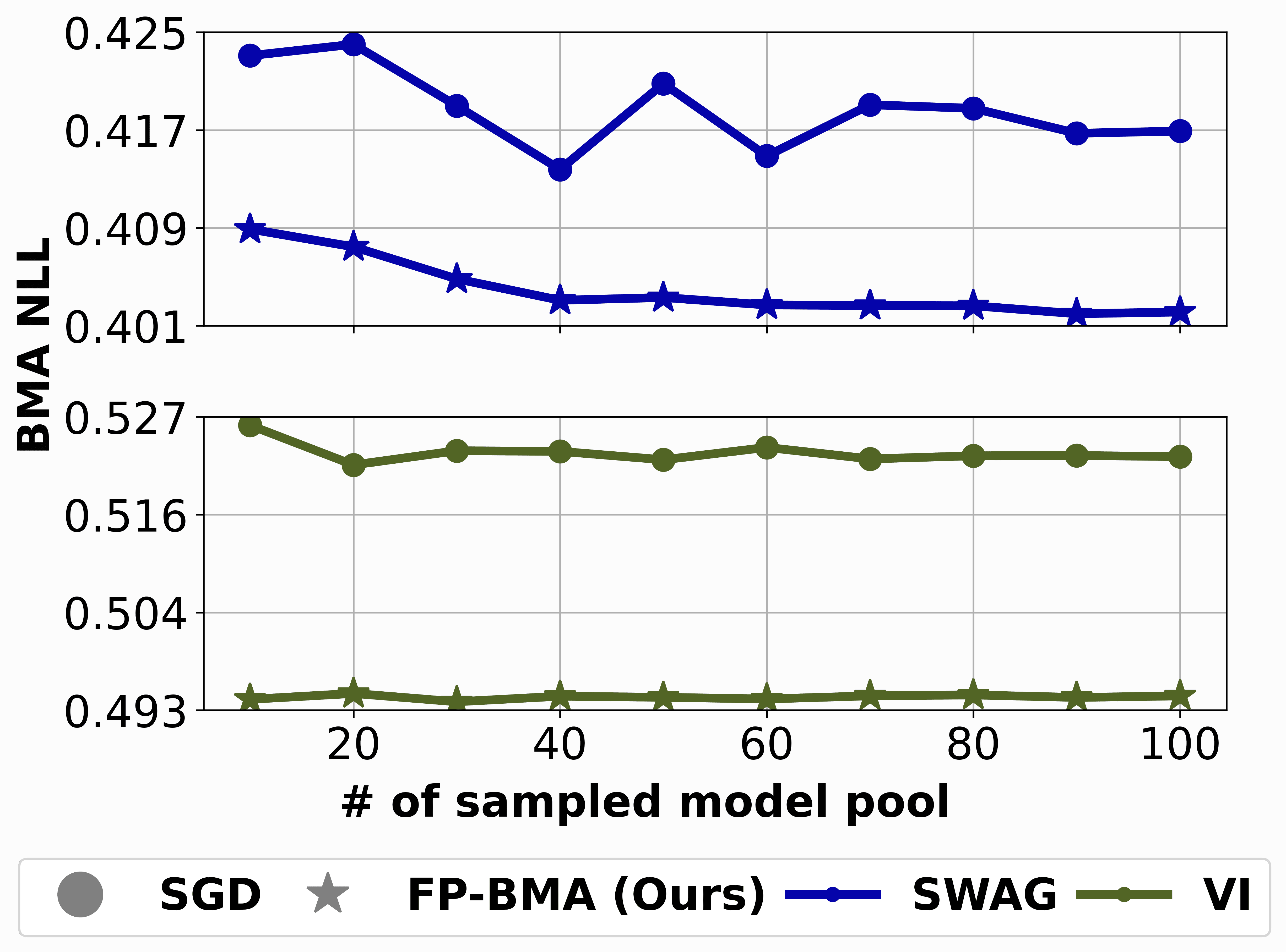}
    \caption{Constant - NLL}
  \end{subfigure}\\[1ex]
    \begin{subfigure}{0.33\textwidth}
    \centering
    \includegraphics[width=\linewidth]{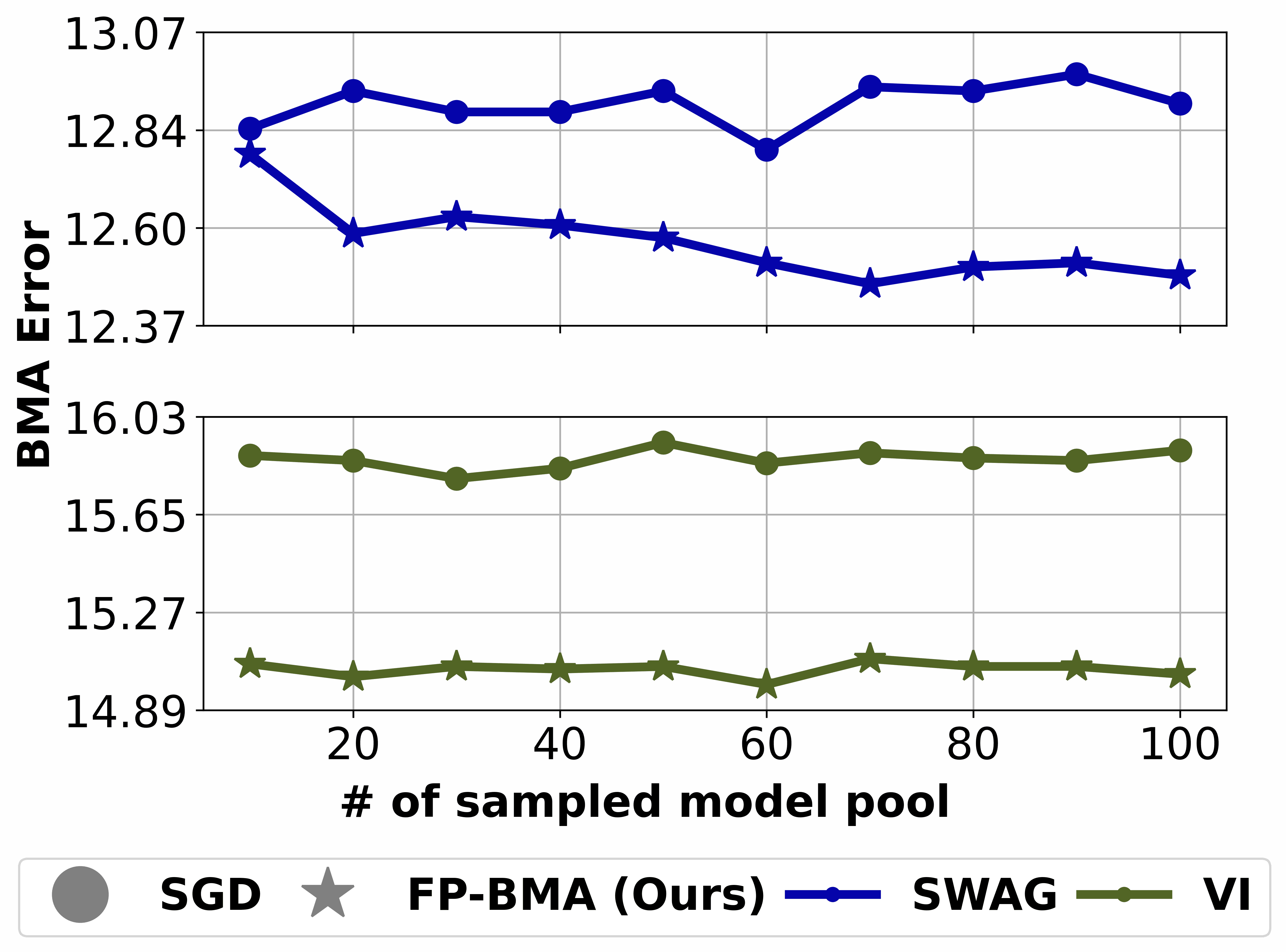}
    \caption{Cos Decay - Error}
  \end{subfigure}%
  \begin{subfigure}{0.33\textwidth}
    \centering
    \includegraphics[width=\linewidth]{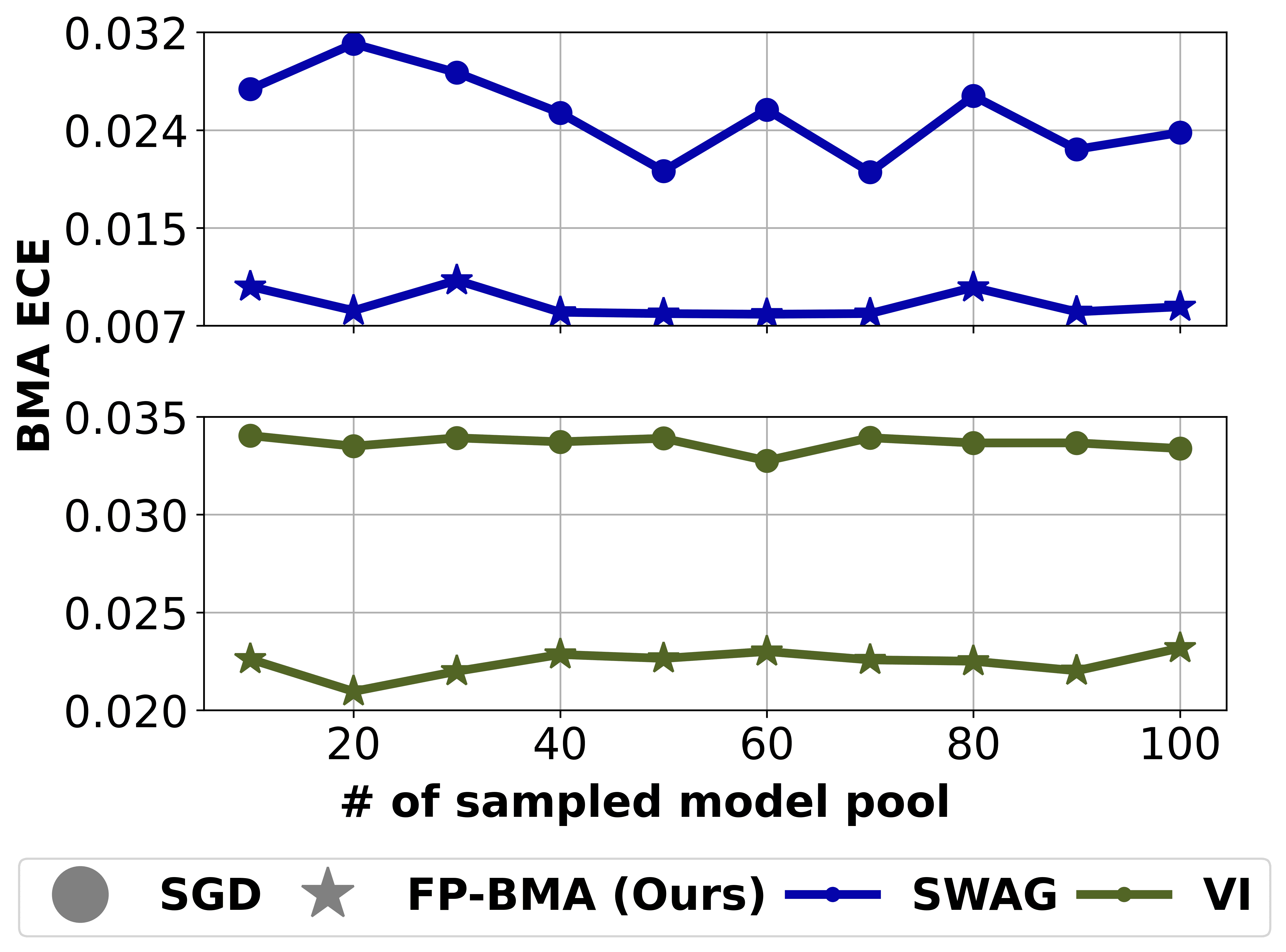}
    \caption{Cos Decay - ECE}
  \end{subfigure}%
  \begin{subfigure}{0.33\textwidth}
    \centering
    \includegraphics[width=\linewidth]{figure/bma_num_plot/c10/c10_cosdecay_nll.png}
    \caption{Cos Decay - NLL}
  \end{subfigure}\\[1ex]
    \begin{subfigure}{0.33\textwidth}
    \centering
    \includegraphics[width=\linewidth]{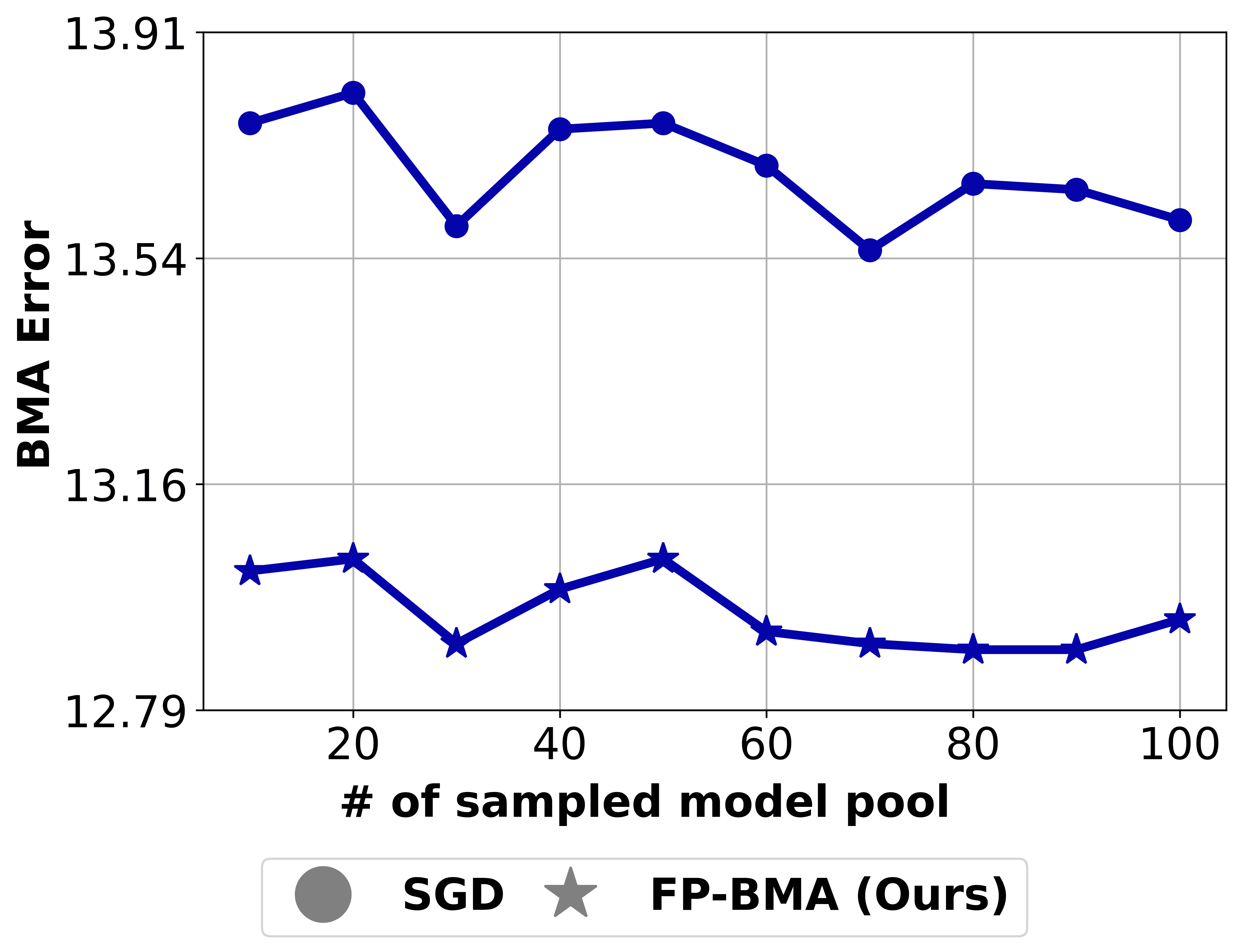}
    \caption{SWAG lr - Error}
  \end{subfigure}%
  \begin{subfigure}{0.33\textwidth}
    \centering
    \includegraphics[width=\linewidth]{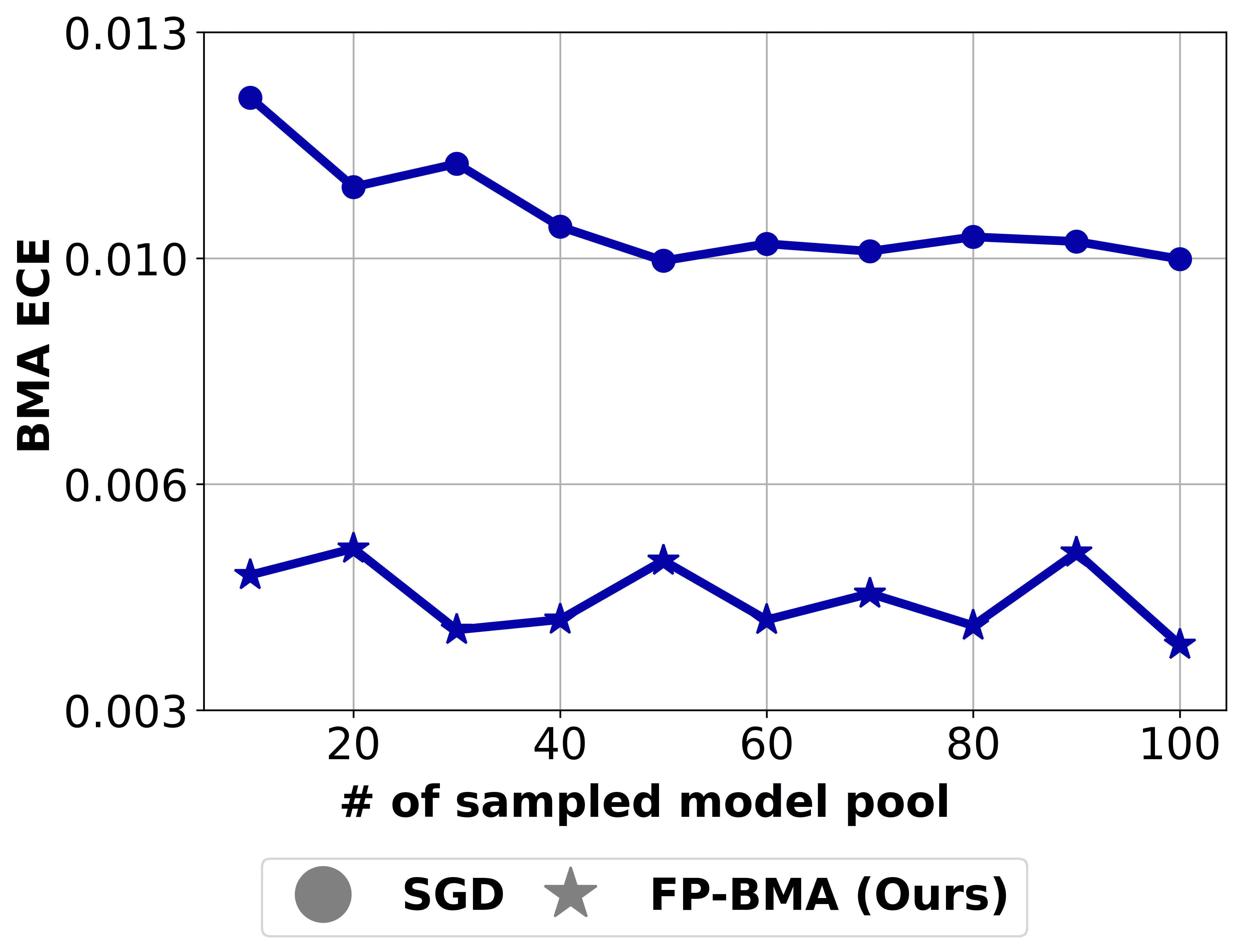}
    \caption{SWAG lr - ECE}
  \end{subfigure}%
  \begin{subfigure}{0.33\textwidth}
    \centering
    \includegraphics[width=\linewidth]{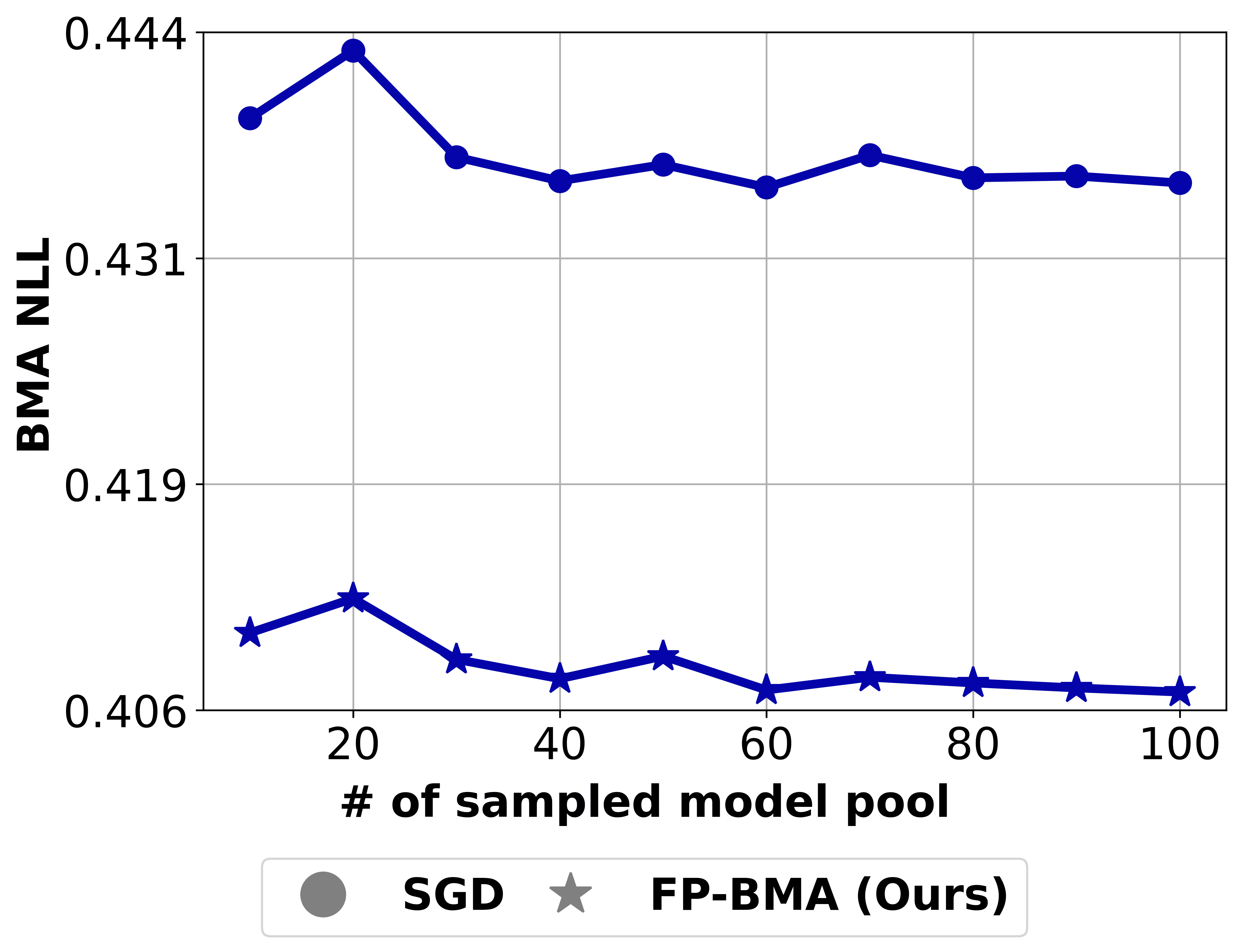}
    \caption{SWAG lr - NLL}
  \end{subfigure}\\[1ex]
  \caption{Performance variation based on sampling considering flatness among BMA on CIFAR10. Each row means the Constant, Cos Decay, and SWAG lr scheduler. Each column denotes classification error, ECE, and NLL. It reveals that the flatness should be taken into account for efficient BMA.
}
  \label{fig:c10_flat_bma_plot}
\end{figure}

We also inspect the influence of flatness on BMA performance throughout all considered schedulers. We train ResNet18 on CIFAR10. Figure~\ref{fig:c10_flat_bma_plot} shows the results. Each row means Constant, Cosine Decay, and SWAG lr scheduler, and each column denotes the classification error, ECE, and NLL.

Two main findings were observed consistent with Figure~\ref{fig:bma_num_plot_nll}: (1) BNNs trained using the proposed FP-BMA showed superior performance compared to those trained with SGD, suggesting that flatness influences posterior quality and contributes to enhanced BMA performance. (2) FP-BMA training allowed predictive distributions to converge with fewer BMA samples, meaning effective approximation can be achieved with a smaller number of samples.

\clearpage
\section{Proof and Derivation}\label{sec:proof_and_derivation}
\subsection{Proof of Theorem 1}\label{subsec:proof_of_theorem_1}
\paragraph{Flatness Bound of BMA}
We first derive flatness bound of BMA. We assume $M$ model $w_m, m=1, .., M$, whose Hessian matrices $H_{f_m}$ (defined in Eq.~\ref{eq:bma_hessian}) are Hermitian. $w_{\text{WA}} = 1/M \sum_{m=1}^M w_m$ is simple weight averaging and the Hessian of $w_{\text{WA}}$, $H_{f_{\text{WA}}}$, also be a Hermitian matrix.
Weyl’s inequality is known to bound the eigenvalues of Hermitian matrices.

\begin{proposition}
\label{proposition:wely_inequality}
\textbf{(Weyl's Inequality)} For Hermitian matrices $C_m \in \mathbb{C}^{p \times p}$, $k, l = 1, ..., p$, \begin{align}
\label{eq:wely_ineq}
\lambda_{k+l-1}(C_i + C_j) \le \lambda_k(C_i) + \lambda_l (C_j) \le  \lambda_{k+l-N}(C_i + C_j). \nonumber\\
\end{align}
\end{proposition}

Let $k=1$ and $l=1$, then Eq.~\ref{eq:wely_ineq} can be written as:
\begin{align}
\lambda_1(C_i + C_j) \le \lambda_1(C_i) + \lambda_1(C_j). \nonumber
\end{align}

As we have $M$ Hermitian matrices, it can be expanded as:
\begin{align}
\label{eq:upper_bound}
    \lambda_1(\frac{1}{M}\sum_{m=1}^M H_{f_m}) \le \frac{1}{M}\sum_{m=1}^M \lambda_1 (H_{f_m}).
\end{align}

One the other hand, we can let $(k,l) = \{(1, p), (p, 1)\}$ and rewrite the Eq.~\ref{eq:wely_ineq} as:
\begin{align}
    \max \{\lambda_1(C_i) + \lambda_p(C_j), \lambda_p(C_i) + \lambda_1(C_j)\} \le \lambda_1 (C_i + C_j). \nonumber
\end{align}

Again, set $M$ Hermitian matrices we have, it can be expanded as:
\begin{equation}
\label{eq:lower_bound}
    \max \left( \Bigg{\{} \frac{1}{M} \bigg{(} \lambda_{1}(H_{f_m}) + \sum_{\substack{n=1 \\ n \neq m}}^{M} \lambda_{p}(H_{f_n}) \bigg{)} \Bigg{\}}_{m=1}^M \right) \le \lambda_{1}(\frac{1}{M}\sum_{m=1}^M H_{f_m}).
\end{equation}

By combining Eq.~\ref{eq:upper_bound} with Eq.~\ref{eq:lower_bound} and substituting $\lambda_1$ to $\lambda_{\text{max}}$ and $\lambda_p$ to $\lambda_{\text{min}}$, the flatness of averaged weight parameter is bounded as:
\begin{equation}
\label{eq:wa_flatness_bound}
    \max \left( \Bigg{\{} \frac{1}{M} \bigg{(} \lambda_{\text{max}}(H_{f_m}) + \sum_{\substack{n=1 \\ n \neq m}}^{M} \lambda_{\text{min}}(H_{f_n}) \bigg{)} \Bigg{\}}_{m=1}^M \right) \le \lambda_{\text{max}}(\frac{1}{M}\sum_{m=1}^M H_{f_m}) \le \frac{\sum_{m=1}^M \lambda_{\text{max}}(H_{f_m})}{M}.
\end{equation}

However, Eq~\ref{eq:wa_flatness_bound}, as a bound for weight averaging (WA), cannot be directly applied to BMA, which marginalizes diverse predictions. To bridge this gap, we leverage Proposition~\ref{proposition:wa_and_ens}. which characterized the close relation between WA and BMA~\citep{izmailov2018averaging, wortsman2022model, rame2022diverse}. 

\begin{proposition}
\label{proposition:wa_and_ens}
(\citet{rame2022diverse}) Given predictions of model $f_m(\cdot)$ parameterized by $w_m$, those of weight averaged model $f_{\text{WA}}$ parameterized by $w_{\text{WA}}= \frac{1}{M} \sum_{m=1}^M w_m$, those of BMA $f_{\text{BMA}}$, and arbitrary twice differentiable loss function $\ell(\cdot)$, let $\Delta =\|f_{\text{BMA}}(x) - f_{\text{WA}}(x)\|_2$. Then, $\forall (x,y)$
\begin{align}
    \ell(f_{\text{WA}}(x), y) &= \ell(f_{\text{BMA}}(x), y) + O(\Delta). \nonumber
\end{align}
\end{proposition}

Lemma~\ref{proposition:wa_and_ens} shows that the predictions of BMA can be approximated with those of WA linearly. The error term is discarded in the process of obtaining the Hessian:
\begin{equation}
\label{eq:wa_bma_hessian}
H_{f_{\text{WA}}} \approx H_{f_{\text{BMA}}}
\end{equation}

By putting Eq.~\ref{eq:wa_bma_hessian} into Eq.~\ref{eq:wa_flatness_bound}, it leads to Lemma~\ref{lemma:bma_eign_bound}.

\setcounter{lemma}{0}
\begin{lemma}
Let twice differentiable loss $\ell(\cdot)$, predictions of model $f_m(\cdot)$ parameterized by $w_m$, and predictions of BMA $f_{\text{BMA}}(\cdot).$
With $M$ model sample $\{w_m\}_{m=1}^M$, the maximal eigenvalue of averaged Hessian of loss $\lambda_{\text{max}}(H_{f_{\text{BMA}}})$ is bounded as follow:
\begin{align}
\max \left( \Bigg{\{} \frac{1}{M} \bigg{(} \lambda_{\max}(H_{f_m}) + \sum_{\substack{n=1 \\ n \neq m}}^{M} \lambda_{\min}(H_{f_n}) \bigg{)} \Bigg{\}}_{m=1}^M \right) \le \lambda_{\max}(H_{f_{\text{BMA}}}) \le \frac{\sum_{m=1}^M \lambda_{\max}(H_{f_m})}{M}. \nonumber
\end{align}
\end{lemma}

\paragraph{Generalization Error Bound for BMA}
To further elucidate the theoretical connection between posterior flatness and generalization, we present the following proposition which is adapted from the generalization analysis of Eigen-SAM~\citep{luo2024explicit}:

\begin{proposition}[Adapted from Eigen-SAM]\label{proposition:sharpness_generalization}
Let $\ell : \mathbb{R}^p \to \mathbb{R}$ be a loss function upper bounded by $L$, and assume its third-order derivatives are uniformly bounded by a constant $C$. Suppose the following inequality holds for any parameter $\theta$:
\[
    \ell_{\mathcal{D}}(\theta) \leq \mathbb{E}_{\epsilon}[\ell_{\mathcal{D}}(\theta + \epsilon)] \qquad \text{where } \epsilon \sim \mathcal{N}(0, \sigma^2 I_p).
\]
Then, for any $\delta \in (0, 1)$ and $\sigma > 0$, with probability at least $1 - \delta$ over the training set $\mathcal{S} \sim \mathcal{D}^n$, we have:
\[
\begin{aligned}
    \ell_{\mathcal{D}}(\theta) \leq \ell_{\mathcal{S}}(\theta) + \frac{p \sigma^2}{2} \lambda_{\max}(\nabla^2 \ell_{\mathcal{S}}(\theta)) + \frac{C p^3 \sigma^3}{6}  + \frac{L}{2\sqrt{n}} \sqrt{p \log\left(1 + \frac{\|\theta\|^2}{p \sigma^2}\right) + O(1) + 2 \log\frac{1}{\delta} + 4 \log(n + p)}
\end{aligned}
\]
\end{proposition}

Building upon Lemma~\ref{proposition:wa_and_ens} and Proposition~\ref{proposition:sharpness_generalization}, we prove a new generalization bound that formally connects posterior flatness to the generalization performance of BMA:

\setcounter{theorem}{0}
\begin{theorem}[Generalization Bound for BMA]\label{theorem:bma_generalization_formal}
Let $w_m$ be samples from a variational posterior $q(w)$, and let $f_{\text{BMA}}$ denote the BMA predictor obtained by averaging predictions over these samples. Assume the conditions in Proposition~\ref{proposition:wa_and_ens} and Proposition~\ref{proposition:sharpness_generalization} hold for each sampled model $f_m$ with parameter $w_m$. Then, for any $\delta \in (0, 1)$ and $\sigma > 0$, with probability at least $1 - \delta$ over the training set $\mathcal{S} \sim \mathcal{D}^n$, the generalization error of the BMA predictor is upper bounded as:
\[
\ell_{\mathcal{D}}(f_{\text{BMA}}) \leq \ell_{\mathcal{S}}(f_{\text{BMA}}) + \frac{p \sigma^2}{2} \lambda_{\max}(H_{f_{\text{BMA}}}) + \frac{C p^3 \sigma^3}{6} + \frac{L}{2\sqrt{n}} \sqrt{p \log\left(1 + \frac{\|f_{\text{BMA}}\|^2}{p \sigma^2}\right) + O(1) + 2 \log \frac{1}{\delta} + 4 \log(n + p)}
\]
where $H_{f_{\text{BMA}}}$ denotes the Hessian of the loss evaluated at $f_{\text{BMA}}$.
\end{theorem}

\subsection{Derivation of Bayesian Flat-Seeking Optimizer}\label{subsec:derivation_odf_bayesian_flat_seeking_optimizer}
\subsubsection{Setting}\label{subsubsec:setting}
Let model parameter $w \subseteq \mathbb{R}^p$ and $w \sim \mathcal{N}(\mu, \Sigma)$.
While fully-factorized or mean-field covariance is de facto in Bayesian Deep Learning, it cannot capitalize on strong points of Bayesian approach.
Inspired from SWAG, we approximate covariance combining diagonal covariance $\sigma \subseteq \mathbb{R}^p$ and low-rank matrix $L \subseteq \mathbb{R}^{p \times K}$ with low-rank component $K$.
Then, we can simply sample $w = \mu + \frac{1}{\sqrt{2}}(\sigma z_1 + L z_2)$, where $z_1 \sim \mathcal{N}(0, I_p)$ and $z_2 \sim \mathcal{N}(0, I_K)$ where $p$, $K$ denotes the number of parameter, low-rank component, respectively. We treat flattened $\mu$, $\sigma$, and $L$, and concatenate as $\theta = \text{Concat}(\mu; \sigma; L)$.

\subsubsection{Objective function}\label{subsubsec:objective_function}
We compose our objective function with probabilistic weight, using KL Divergence as a metric to compare between two weights.

\begin{equation}
\label{eq:main_loss_app}
    \ell^\gamma_{\text{FP-BMA}}(\theta) = \max_{d|\theta+\Delta\theta, \theta| \leq \gamma^2} \ell(\theta + \Delta\theta) + \beta \textrm{D}_{\textrm{KL}}(p_\theta (w|\mathcal{D}) || p (w))
\end{equation}
\begin{equation}
\label{eq:divergence_app}
    \textrm{s.t.} \ \  d|\theta+\Delta\theta, \theta| = \textrm{D}_{\textrm{KL}} \big[ p_{\theta+\Delta\theta}(w|\mathcal{D})||p_\theta(w|\mathcal{D}) \big].
\end{equation}

\subsubsection{Optimization}\label{subsubsec:optimization}

\paragraph{From KL Divergence to Fisher Information Matrix}
We can consider three options of perturbation on mean and covariance parameters of $w$: 1) Perturbation on mean, 2) perturbation on mean and diagonal variance, 3) Perturbation on mean and whole covariance. All of them can be approximated to Fisher Information Matrix.
Here, we show the relation between KLD and FIM considering the probation option 3.

Following FSAM, we deal with parameterized and conditioned as same notation:
\begin{equation}
    p_{\theta+\Delta\theta}(w|\mathcal{D}) = p(w|\mathcal{D}, \theta+\Delta\theta). \nonumber
\end{equation}

By definition of KL divergence, we rewrite Eq.~\ref{eq:divergence_app} as:
\begin{equation}
\textrm{D}_{\textrm{KL}}[p(w|\mathcal{D}, \theta + \Delta \theta) || p(w|\mathcal{D}, \theta)] = \int_{w} p(w | \mathcal{D}, \theta + \Delta \theta) \ \log \frac{p(w|\mathcal{D}, \theta + \Delta \theta)}{p(w|\mathcal{D}, \theta)} dw.
\label{eq:def_KLD_app}
\end{equation}
\\
In Eq.~\ref{eq:def_KLD_app}, we apply first-order Taylor Expansion:
\begin{equation}
\begin{split}
    &p(w|\mathcal{D}, \theta + \Delta \theta) \approx p(w|\mathcal{D}, \theta) + \nabla_{\theta} p(w | \mathcal{D}, \theta)^{T} \Delta \theta. \\
    \label{eq:taylor_expansion_app}
    &\log p(w|\mathcal{D}, \theta + \Delta \theta) \approx \log p(w|\mathcal{D}, \theta) + \nabla_{\theta} \log p(w | \mathcal{D}, \theta)^{T} \Delta \theta. \\
\end{split}
\end{equation}
\\
Substitute right terms of Eq.~\ref{eq:def_KLD_app} with Eq.~\ref{eq:taylor_expansion_app}:
\begin{align}
    &\int_{w} p(w | \mathcal{D}, \theta + \Delta \theta) \ \log \frac{p(w|\mathcal{D}, \theta + \Delta \theta)}{p(w|\mathcal{D}, \theta)} dw \nonumber \\ 
    &=\int_{w} \big( p(w|\mathcal{D}, \theta) + \Delta \theta^{T} \nabla_{\theta} p(w | \mathcal{D}, \theta) \big) \nabla_{\theta} \log p(w|\mathcal{D}, \theta)^{T}\Delta\theta  \ dw \nonumber \\ 
    &=\int_{w} p(w|\mathcal{D}, \theta) \nabla_{\theta}\log p(w|\mathcal{D}, \theta)^{T}\Delta\theta dw \nonumber \\ 
    & \ \ + \int_{w}\Delta\theta^{T}p(w|\mathcal{D}, \theta)\nabla_{\theta}\log p(w|\mathcal{D}, \theta)\nabla_{\theta}\log p(w|\mathcal{D}, \theta)^{T}\Delta\theta \ dw.
    \label{eq:sub_taylor_app}
\end{align}
\\
First term of Eq.~\ref{eq:sub_taylor_app} is equal to 0:
\begin{equation}
\begin{split}
    &\int_{w}p(w|\mathcal{D}, \theta)\nabla_{\theta}\log p(w|\mathcal{D}, \theta) \ dw\\
    &=\int_{w}p(w|\mathcal{D}, \theta)\frac{\nabla_{\theta}p(w|\mathcal{D}, \theta)}{p(w|\mathcal{D}, \theta)} \ dw\\
    &=\int_{w} \nabla_{\theta} p(w|\mathcal{D}, \theta) \ dw \ = \nabla_{\theta} \int_{w} p(w|\mathcal{D}, \theta) = 0. \\
    \label{eq:pf_zero_term_app}
\end{split}
\end{equation}
\\
Using Eq.~\ref{eq:sub_taylor_app} and Eq.~\ref{eq:pf_zero_term_app}, Eq.~\ref{eq:def_KLD_app} can be rewritten as Fisher information matrix by the definition of expectation:
\begin{equation}
\begin{split}
    &D_{KL}[p(w|\mathcal{D}, \theta+\Delta\theta) ||p(w|\mathcal{D}, \theta)]  \\
    &=\int_{w} \Delta\theta^{T}p(w|\mathcal{D}, \theta)\nabla_{\theta}\log p(w|\mathcal{D}, \theta) \nabla_{\theta} \log p(w|\mathcal{D}, \theta)^{T}\Delta\theta  \\
    &=\Delta\theta^{T}\mathbb{E}_{w} [\nabla_{\theta}\log p(w|\mathcal{D}, \theta) \nabla_{\theta} \log p(w|\mathcal{D}, \theta)^{T}]\Delta\theta \\
    &=\Delta \theta^T F_\theta(\theta)\Delta\theta, \\
\end{split}
\label{eq:KLD_to_FIM_app}
\end{equation}
\\
where $F_\theta(\theta)=\mathbb{E}_{w, \mathcal{D}}[\nabla_{\theta}\log p(w|\mathcal{D}, \theta) \nabla_{\theta} \log p(w|\mathcal{D}, \theta)^{T}]$.

It's too expensive to calculate Fisher information matrix $F(\theta)$ in practice.
We introduce a pseudo inverse for Fisher information matrix $F_\theta(\theta)^{-1}$ with Samelson inverse of a vector~\citep{gentle2007matrix, sidi2017vector, wynn1962acceleration} :
\begin{equation}
F_\theta(\theta)^{-1} = \frac{\nabla_{\theta} \log p(w|\mathcal{D}, \theta) \nabla_{\theta} \log p(w|\mathcal{D}, \theta)^{T}}{\| \nabla_{\theta} \log p(w|\mathcal{D}, \theta)  \|^{4}}.
\end{equation}

\paragraph{Lagrangian Dual Problem}

From the result of Eq.~\ref{eq:KLD_to_FIM_app}, we can rewrite the Eq.~\ref{eq:main_loss_app}:
\begin{equation}
\label{eq:main_loss_2_app}
    \ell^\gamma_{\text{FP-BMA}}(\theta) = \max_{\Delta\theta^T F_\theta(\theta) \Delta\theta \le \gamma^2} \ell(\theta + \Delta \theta).
\end{equation}

We can reach the optimal perturbation of FP-BMA $\Delta\theta^*$ by using Taylor Expansion on $\ell(\theta+\Delta\theta)$ of Eq.~\ref{eq:main_loss_app}:

\begin{equation}
\label{eq:approximation_l(w^star)_app}
    \ell(\theta + \Delta\theta) = \ell(\theta) + \nabla_\theta \ell(\theta)^T \Delta \theta.
\end{equation}
\\

Using Eq.~\ref{eq:approximation_l(w^star)_app}, we can rewrite Eq.~\ref{eq:main_loss_app} as Lagrangian dual problem:
\begin{equation}
\label{eq:lagrangian_loss_app}
     L(\Delta\theta, \lambda) =  \ell(\theta)  + \nabla \ell_\theta (\theta)^{T} \Delta\theta - \lambda(\Delta\theta^{T}F_\theta(\theta)\Delta\theta - \gamma^2).
\end{equation}
\\

Differentiating Eq.~\ref{eq:lagrangian_loss_app}, we get $\Delta\theta^*$:
\begin{align}
\label{eq:Delta_theta^*_app}
    &\frac{\alpha L(\Delta\theta, \lambda)}{\alpha \Delta\theta} = \nabla_\theta \ell(\theta)^T - 2\lambda \Delta\theta^T F_\theta(\theta) = 0 \notag \\
    & \therefore \ \Delta\theta^* = \frac{1}{2\lambda}F_\theta(\theta)^{-1} \nabla_\theta \ell(\theta).
\end{align}
\\

Putting $\Delta\theta^*$ of Eq.~\ref{eq:Delta_theta^*_app} into $\Delta\theta$ of Eq.~\ref{eq:lagrangian_loss_app}, we can rewrite Eq.~\ref{eq:lagrangian_loss_app}:
\begin{equation}
\label{eq:lagrangian_loss_with_Delta_theta_*_app}
\begin{split}
    L(\Delta\theta^*, \lambda) &= \ell(\theta) + \frac{1}{2\lambda}\nabla_\theta \ell(\theta)^T F_\theta(\theta)^{-1} \nabla_\theta \ell(\theta) \\
     &- \frac{1}{4\lambda}\nabla_\theta \ell(\theta)^T F_\theta(\theta)^{-1} \nabla_\theta \ell(\theta) + \lambda\gamma^2.
\end{split}
\end{equation}
\\

By taking derivative of Eq.~\ref{eq:lagrangian_loss_with_Delta_theta_*_app} w.r.t. $\lambda$, we can also get $\lambda^*$:
\begin{align}
\label{eq:lambda^*_app}
    &\frac{\alpha L(\Delta\theta^*, \lambda)}{\alpha \lambda} = -\frac{1}{2\lambda^2}\nabla_\theta \ell(\theta)^T F_\theta(\theta)^{-1} \nabla_\theta \ell(\theta)
    + \frac{1}{4\lambda^2}\nabla_\theta \ell(\theta)^T F_\theta(\theta)^{-1} \nabla_\theta \ell(\theta) +\gamma^2 = 0 \notag \\
    & 4\lambda^2\gamma^2 = \nabla_\theta \ell(\theta)^T F_\theta(\theta)^{-1}  \nabla_\theta \ell(\theta) \notag \\
    & \therefore \ \lambda^* = \frac{\sqrt{\nabla_\theta \ell(\theta)^T F_\theta(\theta)^{-1} \nabla_\theta \ell(\theta)}}{2\gamma}.
\end{align}
\\

Finally, we get our $\Delta \theta_{\text{FP-BMA}}$ by substituting Eq.~\ref{eq:lambda^*_app} into
Eq.~\ref{eq:Delta_theta^*_app}:
\begin{equation}
\label{eq:Delta_theta_BSAM_app}
    \Delta\theta_{\text{FP-BMA}} = \gamma\frac{F_\theta(\theta)^{-1} \nabla_\theta \ell(\theta)}{\sqrt{\nabla_\theta \ell(\theta)^T F_\theta(\theta)^{-1}\nabla_\theta \ell(\theta)}}.
\end{equation}

\subsection{Proof of Theorem 2}\label{subsec:proof_of_theorem2}
\subsubsection{FP-BMA to FSAM}
Theorem \ref{theorem:generalized_sabma} shows that FP-BMA is degenerated to FSAM under DNN and diagonal FIM setting. Deterministic parameters draw out the constant prior $p(w|x)=c$ and mean-only variational parameters $w = \theta$.

First, we can rewrite the log posterior $\log p_\theta (w | x, y)$ with Bayes rule:
\begin{equation}
\label{eq:posterior_bayes}
    \log p_\theta (w|x, y) = \log p_\theta (y|x,w) + \log p_\theta (w|x) - Z,
\end{equation}
where $Z$ is constant independent of $w$. Is is noted that the log posterior is divided into the log predictive distribution and log prior. Also, note that the prior is conditioned on the data to align with a generalized notation. The prior can depend on the input; however, this dependence is often ignored in practice~\citep{marek2024can}.

By taking derivative with respect to $\theta$ on Eq.~\ref{eq:posterior_bayes}, the constant $Z$ goes to $0$:
\begin{equation}
    \nabla_\theta \log p_\theta (w|x,y) = \nabla_\theta p_\theta (y|x,w) + \nabla_\theta \log p_\theta (w|x). \nonumber
\end{equation}

We have constant prior $p(w|x)=c$ in deterministic setting and it makes the gradient of log posterior and log predictive distribution:
\begin{equation}
\label{eq:same_output_weight_gradient}
    \nabla_\theta \log p_\theta (w|x,y) = \nabla_\theta p_\theta (y|x,w).
\end{equation}
Underlying Eq.~\ref{eq:same_output_weight_gradient}, it is possible to substitute the gradient of log posterior into the gradient of log predictive distribution and FIM over posterior goes to FIM over predictive distribution:
\begin{align}
\label{eq:same_output_weight_fisher}
    F_\theta (\theta) = &\mathbb{E}_{w, \mathcal{D}} [ \nabla_\theta \log p_\theta (w|x,y) \nabla_\theta \log p_\theta (w|x, y)^T] \nonumber\\
    &= \mathbb{E}_{w, \mathcal{D}} [ \nabla_\theta \log p_\theta (y|x,w) \nabla_\theta \log p_\theta (y|x, w)^T].
\end{align}
By taking diagonal computation over Eq.~\ref{eq:same_output_weight_fisher}, it goes to $F_y(\theta)$. After that, using the fact that mean-only variational parameters, FP-BMA degnerates to FSAM with $F_y(\theta)$ finally.
\begin{equation}
\label{eq:sabma_to_fsam}
    \Delta \theta_{\text{FP-BMA}} = \gamma\frac{F_y(\theta)^{-1}\nabla_\theta \ell(\theta)}{\sqrt{ F_y(\theta)^{-1} \nabla_\theta \ell(\theta) F_y(\theta)^{-1}}}.
\end{equation}

\subsubsection{FP-BMA to SAM}
It is simple to show that FP-BMA is extended version of SAM by defining FIM over output distribution $F_y(w)$ as identity matrix $I$ in Eq.~\ref{eq:sabma_to_fsam}, FP-BMA goes to SAM.
\begin{equation}
    \Delta \theta_{\text{FP-BMA}} = \gamma\frac{\nabla_w \ell(w)}{\| \nabla_w \ell(w)\|_2}.
\end{equation}

\subsubsection{FP-BMA to NG}
Theorem~\ref{theorem:generalized_sabma} also states the NG can be approximated with FP-BMA under specific conditions.
The update rule of natural gradient and FP-BMA can be written as Eq.~\ref{eq:ng_update_rule} and Eq.~\ref{eq:sabma_update_rule}, respectively.
\begin{align}
    &\theta \leftarrow \theta + \eta_{\text{NG}} F_y (\theta)^{-1} \nabla_\theta \ell(\theta). \label{eq:ng_update_rule} \\
    &\theta \leftarrow \theta + \eta_{\text{FP-BMA}} \nabla_\theta \ell(\theta+\Delta\theta). \label{eq:sabma_update_rule}
\end{align}
where $\eta_{\text{NG}}$ and $\eta_{\text{FP-BMA}}$ denote the learning rate of NG and FP-BMA. Note that we assume the log likelihood as loss fuction.

The $\nabla_\theta \ell(\theta + \Delta\theta)$ in Eq.~\ref{eq:sabma_update_rule} can be approximated with Taylor Expansion, the connection between Hessian and FIM, and Eq.~\ref{eq:same_output_weight_fisher} in DNN setup:
\begin{align}
\label{eq:approx_perturbation}
    \nabla_\theta \ell(\theta + \Delta\theta) &\approx \nabla_\theta \ell(\theta) + \nabla^2_\theta \Delta\theta \nonumber\\
    & = \nabla_\theta \ell(\theta) + \nabla^2_\theta \ell(\theta) \cdot \gamma\frac{F_\theta (\theta)^{-1} \nabla_\theta \ell(\theta)}{\sqrt{\nabla_\theta \ell(\theta)^T F_\theta (\theta)^{-1} \nabla_\theta \ell(\theta)}} \nonumber\\
    &= \nabla_\theta \ell(\theta) + \gamma^\prime \nabla_\theta^2 \ell(\theta) F_\theta(\theta)^{-1} \nabla_\theta \ell(\theta) \ \bigg(\because \text{Let} \ \gamma^\prime = \frac{\gamma}{\sqrt{\nabla_\theta \ell(\theta)^T F_\theta(\theta)^{-1} \nabla_\theta \ell(\theta)}} \bigg) \nonumber\\
    &= [I + \gamma^\prime \nabla^2_\theta \ell(\theta) F_\theta(\theta)^{-1}]\nabla_\theta \ell(\theta) \nonumber\\
    &\approx (1+\gamma^\prime)\nabla_\theta \ell(\theta) \ (\because \nabla_\theta^2 \ell(\theta) \approx F_y (\theta), F_\theta(\theta) = F_y(\theta)).
\end{align}

By using the denoted learning rate $\eta_{\text{FP-BMA}}=\frac{\eta_{\text{NG}}}{I + \gamma^\prime}F_\theta (\theta)^{-1}$, Eq.~\ref{eq:same_output_weight_fisher}, and Eq.~\ref{eq:approx_perturbation}, update rule of FP-BMA approximates to NG.

\clearpage
\section{Experiments}\label{sec:experiments_app}
\subsection{Synthetic Example}\label{subsec:synthetic_example_app}
\begin{figure}[h]
  \centering
  \begin{subfigure}{0.33\textwidth}
    \centering
    \includegraphics[width=\linewidth]{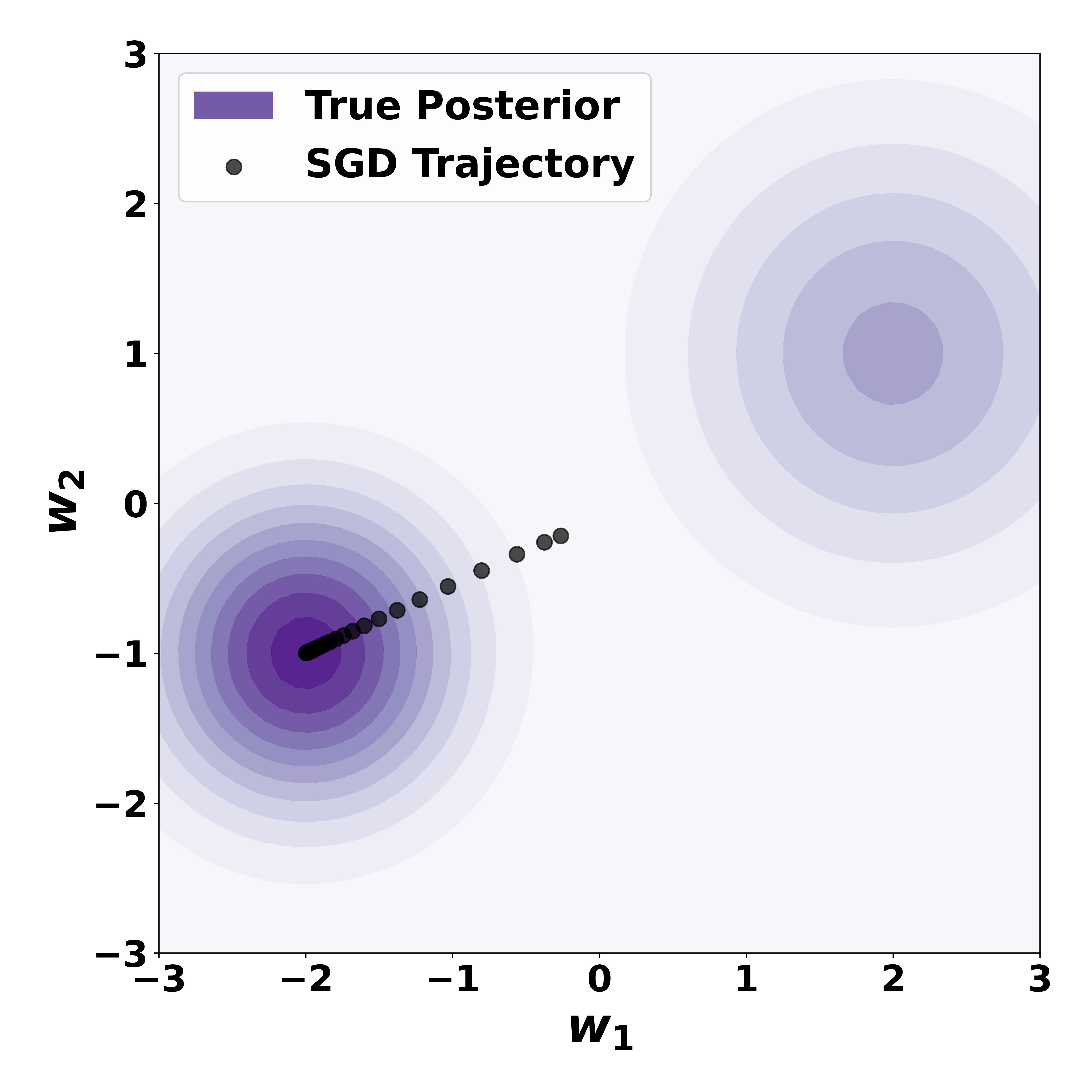}
    \caption{SGD}
  \end{subfigure}%
  \begin{subfigure}{0.33\textwidth}
    \centering
    \includegraphics[width=\linewidth]{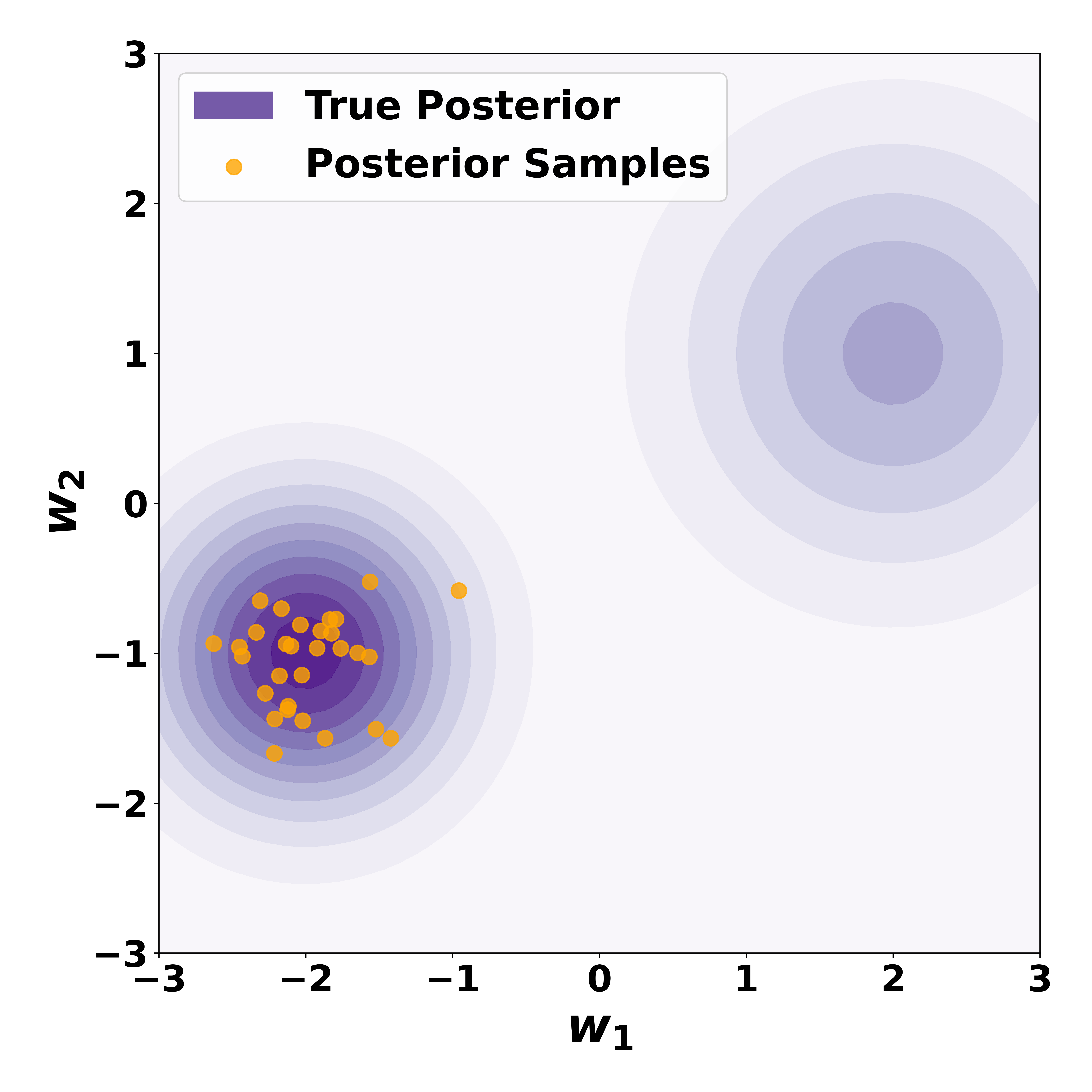}
    \caption{MCMC}
  \end{subfigure}%
  \begin{subfigure}{0.33\textwidth}
    \centering
    \includegraphics[width=\linewidth]{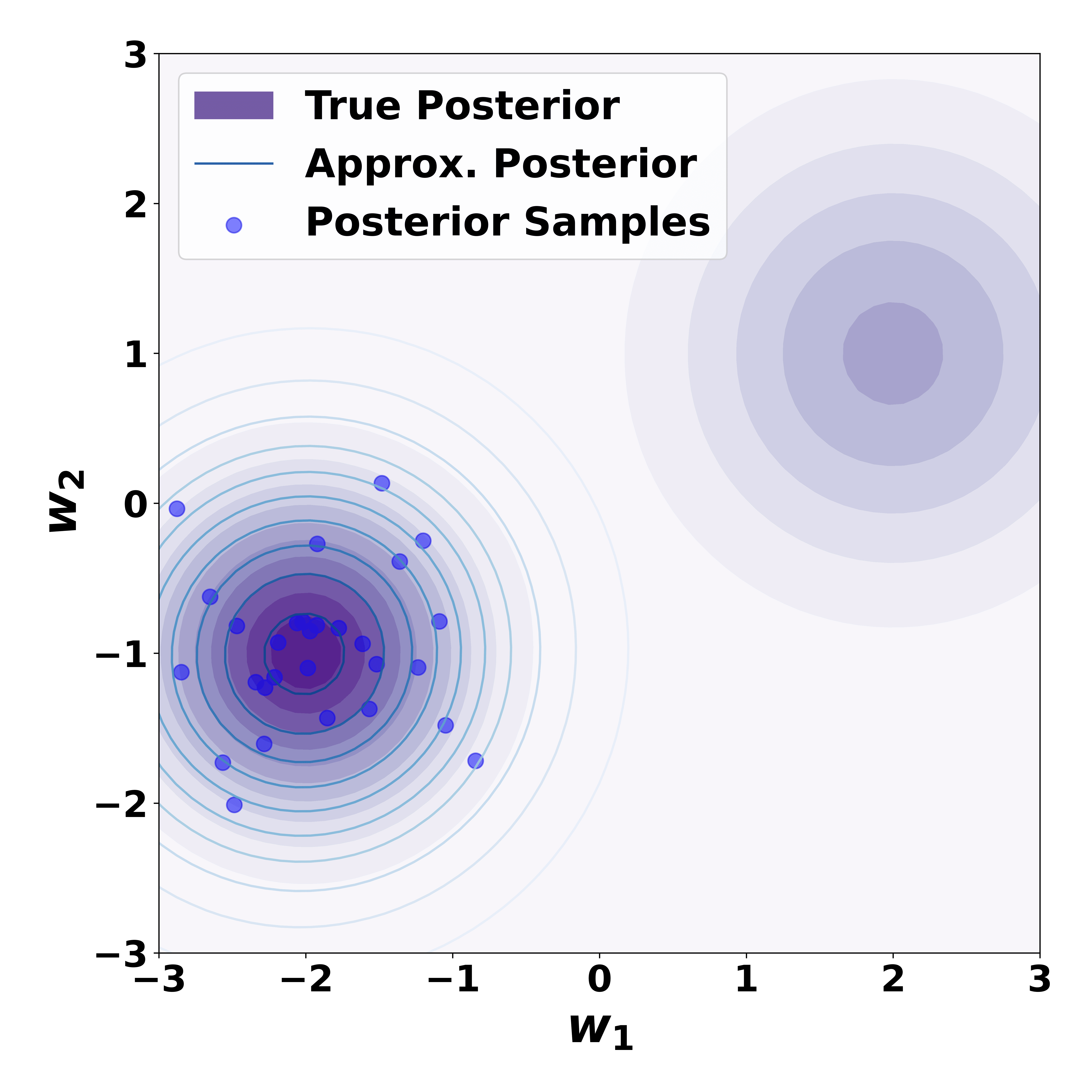}
    \caption{SWAG}
  \end{subfigure}\\[1ex]
    \begin{subfigure}{0.33\textwidth}
    \centering
    \includegraphics[width=\linewidth]{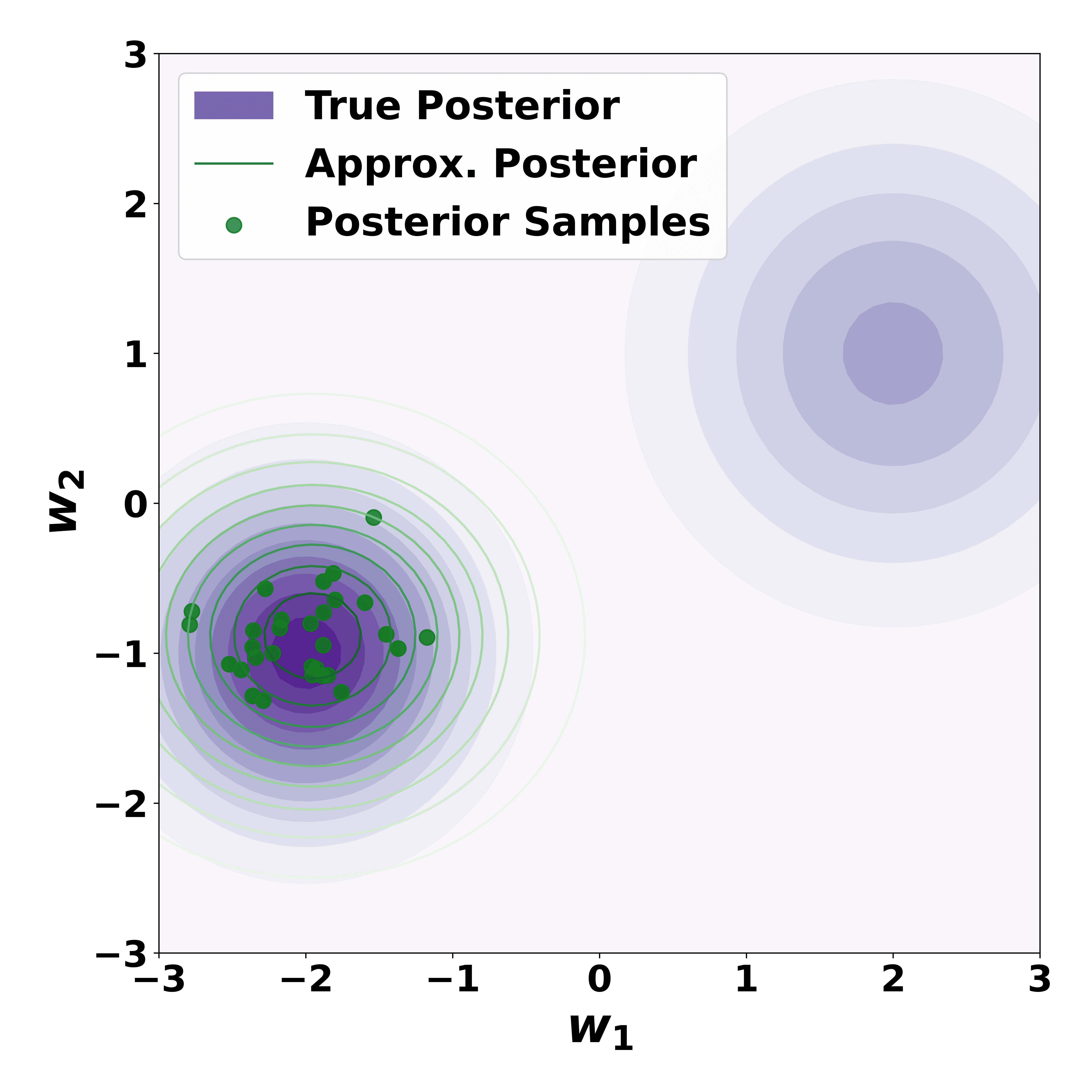}
    \caption{VI}
  \end{subfigure}%
  \begin{subfigure}{0.33\textwidth}
    \centering
    \includegraphics[width=\linewidth]{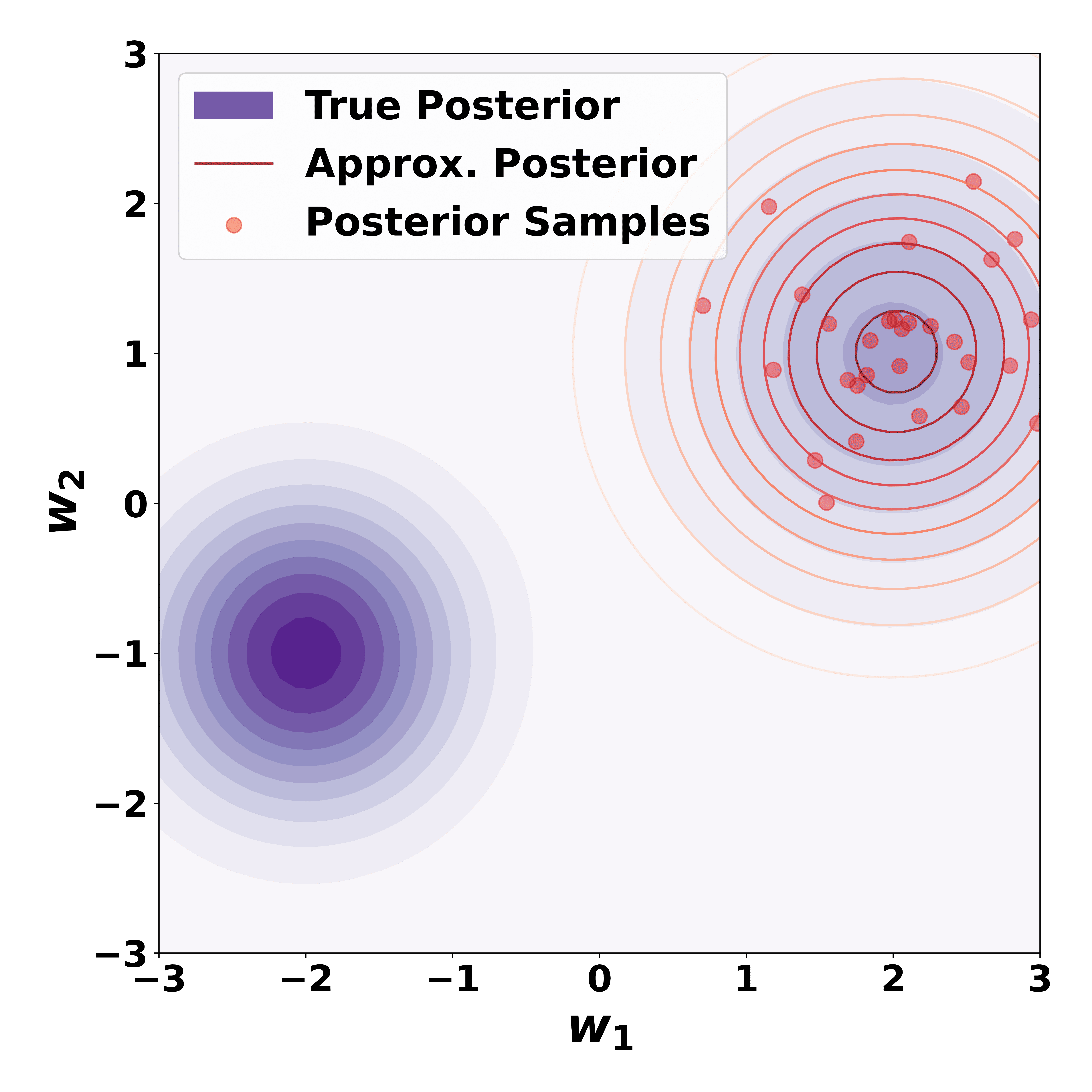}
    \caption{FP-BMA (SWAG)}
  \end{subfigure}%
  \begin{subfigure}{0.33\textwidth}
    \centering
    \includegraphics[width=\linewidth]{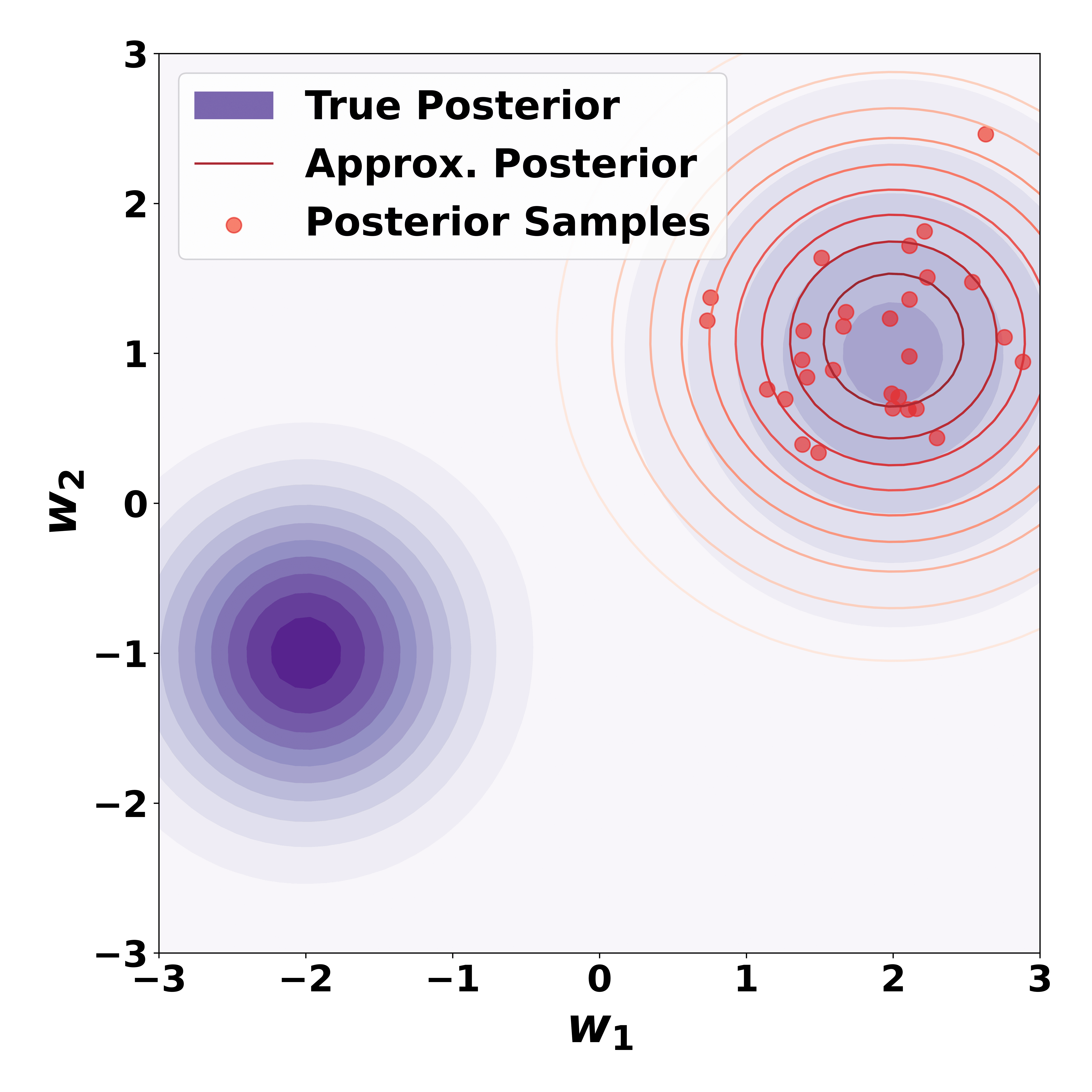}
    \caption{FP-BMA (VI)}
  \end{subfigure}\\[1ex]
  \caption{Posterior approximation with synthetic example. When both flat and sharp modes coexist, we compared how optimizers approximate the posterior. Unlike other methods, the proposed FP-BMA converged to the flat mode, demonstrating its effectiveness in finding more stable solutions.}
  \label{fig:posterior_approx_all}
\end{figure}

Following \citet{li2023entropy}, we construct a loss surface following the distribution $\frac{1}{2}(\mathcal{N}([-2, -1]^T, 0.5I)) + \frac{1}{2}(\mathcal{N}([2, 1]^T, I))$ and set the initial point at $(-0.4, -0.4)$. Unlike other SGD-based methods, FP-BMA efficiently identifies flat modes regardless of the underlying BNN frameworks.

\clearpage
\subsection{Learning From Scratch}\label{subsec:learning_from_scratch_app}
\subsubsection{FP-BMA with diverse BNN frameworks}\label{subsubsec:sabma_with_diverse_bnn_frameworks_1}
In Eq.~\ref{eq:main_loss}, FP-BMA can be applied with various BNN frameworks by using an empirical loss function $\ell(\cdot)$ and adjusting the parameter $\beta$. We commonly set $\ell(\cdot)$ as cross-entropy loss in context of image classification task. Note that FP-BMA was applied only to the normalization layers and the last layer, while all other layers were trained using SGD.

\paragraph{FP-BMA (VI)}
For VI, we follow the loss function of Eq.~\ref{eq:main_loss}.

\paragraph{FP-BMA (MCMC)}
We mainly adopt SGLD for MCMC in this work. For SGLD, we incorporated noise into Eq.~\ref{eq:main_loss} without KLD term ($\beta = 0$) based on the learning rate and the hyperparameter, temperature. In this approach, during the first step, the adversarial posterior is computed without any noise (Eq.~\ref{eq:Delta_theta_star}). In the second step, both the noise and the adversarial posterior are used together in the learning process.

\paragraph{FP-BMA (SWAG)}
SWAG updates the first and second moments along the trajectory of SWA and uses these moments to approximate the posterior with a Gaussian distribution. In Eq.~\ref{eq:main_loss}, $\beta$ is fixed to 0, and as the trajectory of SWA is optimized through FP-BMA, posterior approximation can be performed accordingly.

\subsubsection{Hyperparameters for Experiments}\label{subsubsec:hyperparameters_for_experiments_1}
In this section, we provide the details of the experimental setup for Section~\ref{subsec:learning_from_scratch}. In the other experiments, the range of hyperparameters, excluding the number of epochs, is shared across different backbones and methods. For all experiments, the hyperparameters are selected using grid-search. Configuration of best hyperparameters for each baseline is summarized in Table~\ref{tab:hyperparameter_c10_scratch} and Table~\ref{tab:hyperparameter_c100_scratch}.

\begin{table*}[ht!]
\caption{Hyperparameter Configuration for CIFAR10}
\label{tab:hyperparameter_c10_scratch}
\adjustbox{width=1.0\linewidth}{
\begin{tabular}{cccccccc}
\toprule
Backbone                     & Baseline      & learning rate       & \begin{tabular}[c]{@{}c@{}}$\beta_1$\\ (momentum)\end{tabular} & $\beta_2$ & $\gamma$      & weight decay       \\ \midrule\midrule
\multirow{9}{*}{RN18}  & SGD           & 5e-2 & 9e-1 &  $\times$     & $\times$  &  5e-4   \\
                                     & SAM & 1e-1  & 9e-1 &  $\times$ & 1e-1 & 5e-4   \\
                                     & FSAM & 5e-2 & 9e-1 & $\times$ & 1e-2  & 5e-4    \\
                                     & bSAM & 8e-1 & 9e-1 & 0.999 & 1e-1 & 5e-4    \\
                                     & VI   & 5e-3 & 9e-1 & $\times$ & $\times$ & 5e-4 \\
                                     & \textbf{FP-BMA (VI)} & 5e-2 & 9e-1 & $\times$ & 1e-1 & 5e-4 \\
                                     & MCMC & 1e-1 & $\times$ & $\times$ &  $\times$        & 5e-4 \\
                                     & E-MCMC & 1e-1 & $\times$ & $\times$ & $\times$ & 5e-4 \\
                                     & \textbf{FP-BMA (MCMC)} & 5e-2 & 9e-1 & $\times$ & 5e-2 & 5e-4 \\
                                     & SWAG & 1e-1 & 9e-1 & $\times$ & $\times$ & 5e-4 \\
                                     & F-SWAG & 1e-1 & 9e-1 & $\times$ & 1e-1 & 5e-4 \\
                                     & \textbf{FP-BMA (SWAG)} & 1e-1 & 9e-1 & $\times$ & 1e-1 & 5e-4 \\ \midrule\midrule
\multirow{9}{*}{ViT-B/16$^{\dagger}$}       & SGD & 1e-1 & 9e-1 & $\times$ & $\times$ & 5e-4 \\
                                     & SAM & 1e-1 & 9e-1 & $\times$ & 5e-2 & 5e-4 \\
                                     & FSAM & 1e-1 & 9e-1 & $\times$ & 1e-1 & 5e-4 \\
                                     & bSAM & 5e-1 & 9e-1 & 0.999 & 1e-1 & 5e-4 \\
                                     & VI & 5e-3 & 9e-1 & $\times$ & $\times$ & 5e-4 \\
                                     & \textbf{FP-BMA (VI)} & 5e-3 & 9e-1 & $\times$ & 5e-3 &  5e-4 \\
                                     & MCMC & 2e-2 & $\times$ & $\times$ & $\times$ & 5e-4 \\
                                     & EMCMC & 2e-2 & $\times$ & $\times$ & $\times$ & 5e-4 \\
                                     & \textbf{FP-BMA (MCMC)} & 3e-2 & 9e-1 & $\times$ & 1e-2 & 5e-4 \\
                                     & SWAG & 5e-2 & 9e-1 & $\times$ & $\times$ & 5e-4 \\
                                     & F-SWAG & 5e-2 & 9e-1 & $\times$ & & 5e-4 \\
                                     & \textbf{FP-BMA (SWAG)} & 5e-2 & 9e-1 & $\times$ & 1e-2 & 5e-4 \\ \bottomrule
\end{tabular}
}
\end{table*}
\begin{table*}[ht!]
\caption{Hyperparameter Configuration for CIFAR100}
\label{tab:hyperparameter_c100_scratch}
\adjustbox{width=1.0\linewidth}{
\begin{tabular}{cccccccc}
\toprule
Backbone                     & Baseline      & learning rate       & \begin{tabular}[c]{@{}c@{}}$\beta_1$\\ (momentum)\end{tabular} & $\beta_2$ & $\gamma$      & weight decay       \\ \midrule\midrule
\multirow{9}{*}{RN18}  & SGD           & 1e-1 & 9e-1 &  $\times$     & $\times$  &  5e-4   \\
                                     & SAM &  5e-2 & 9e-1 &  $\times$ & 1e-1 & 5e-4   \\
                                     & FSAM & 1e-1  & 9e-1 & $\times$ & 1e-2 & 5e-4    \\
                                     & bSAM & 1 & 9e-1 & 0.999 & 1e-1 & 5e-4    \\
                                     & VI   & 5e-3 & 9e-1 & $\times$ & $\times$ & 5e-4 \\
                                     & \textbf{FP-BMA (VI)} & 8e-3 & 9e-1 & $\times$ & 2e-1 & 5e-4 \\
                                     & MCMC   & 5e-1 & $\times$ & $\times$ &  $\times$ & 5e-4 \\
                                     & E-MCMC & 5e-1 & $\times$ & $\times$ & $\times$ & 5e-4 \\
                                     & \textbf{FP-BMA (MCMC)} & 1e-1 & 9e-1 & $\times$ & 3e-2 & 5e-4 \\
                                     & SWAG & 1e-1 & 9e-1 & $\times$ & $\times$ & 5e-4 \\
                                     & F-SWAG & 1e-1 & 9e-1 & $\times$ & 1e-1 & 5e-4 \\
                                     & \textbf{FP-BMA (SWAG)} & 3e-1 & 9e-1 & $\times$ & 2e-1 & 5e-4 \\ \midrule\midrule
\multirow{9}{*}{ViT-B/16$^{\dagger}$}       & SGD & 1e-1 & 9e-1 & $\times$ & $\times$ & 5e-4 \\
                                     & SAM & 1e-1 & 9e-1 & $\times$ & 1e-1 & 5e-4 \\
                                     & FSAM & 1e-1 & 9e-1 & $\times$ & 1e-2 & 5e-4\\
                                     & bSAM & 5e-1 & 9e-1 & 0.999 & 1e-1 & 5e-4 \\
                                     & VI & 3e-2 & 9e-1 & $\times$ & $\times$ & 5e-4 \\
                                     & \textbf{FP-BMA (VI)} & 8e-3 & 9e-1 & $\times$ & 1e-1 & 5e-4 \\
                                     & MCMC & 2e-1 & $\times$ & $\times$ & $\times$ & 5e-4 \\
                                     & EMCMC & 1e-1 & $\times$ & $\times$ & $\times$ &  5e-4  \\
                                     & \textbf{FP-BMA (MCMC)} & 5e-2 & 9e-1 & $\times$ & 5e-2 & 5e-4 \\
                                     & SWAG & 1e-1 & 9e-1 & $\times$ & $\times$ & 5e-4 \\
                                     & F-SWAG & 1e-1 & 9e-1 & $\times$ & 1e-1 & 5e-4 \\
                                     & \textbf{FP-BMA (SWAG)} & 1e-1 & 9e-1 & $\times$ & 1e-1 & 5e-4 \\ \bottomrule
\end{tabular}
}
\end{table*}

\paragraph{Stochastic Gradient Descent with Momentum (SGD)}
In this study, we adopt Stochastic Gradient Descent with Momentum as an optimizer for DNN. Learning rate schedule is fixed to cosine decay. We run 300 epochs. The hyperparameter tuning range included learning rate in [1e-4, 1e-3, 1e-2].

\paragraph{Sharpness Aware Minimization (SAM)}
We set SGD with momentum as the base optimizer of SAM. It also ran upon a cosine decay learning rate scheduler. All the range of hyperparameters is shared with SGD with Momenmtum. Additional hyperparameter $\gamma$, the ball size of perturbation, is in [1e-2, 5e-2, 0.1].

\paragraph{Fisher SAM (FSAM)}
We set SGD with momentum as the base optimizer of FSAM. It also ran upon a cosine decay learning rate scheduler. All the range of hyperparameters is shared with SGD with Momenmtum. Additional hyperparameter $\eta$, regularize Fisher impact, is in [1e-2, 1e-1, 1].

\paragraph{SAM as an optimal relaxation of Bayes (bSAM)} We use a cosine learning rate decay scheme. We run 300 epochs with fixed $\beta_1$ and $\beta_2$. The hyperparameter tuning rage included: learning rate in [1e-1, 3e-1, 5e-1, 8e-1, 1], weight decay in [1e-4, 5e-4, 1e-3, 1e-2], damping in [1e-1, 1e-2, 1e-3], and $\gamma$ in [1e-3, 1e-2, 5e-2, 1e-1, 5e-1]. Damping parameter stabilizes the method by adding constant when updating variance estimate.

\paragraph{Variational Inference (VI)} We use MOPED to change DNN into BNN, first. We set prior mean and variance as 0 and 1, respectively. Besides, we set the posterior mean as 0 and variance as 1e-3. We adopt Reparameterization as type of VI. The essential hyperparmeter for MOPED is $\delta$, which adjusts how much to incorporate pre-trained weights. The $\delta$ was searched in [1e-3, 5e-3, 1e-2]. Moreover, we add a hyperparameter $\beta$ for MOPED that can balance the loss term in VI. The $\beta$ is in range [1e-2, 1e-1 ,1]

\paragraph{MCMC} We consistently use SGLD~\citep{welling2011bayesian} for MCMC in this work. It ran upon a cyclic cosine decay learning rate scheduler. The number of cycles was ranged in [2, 4]. The number of sampled models is in [10, 20, 28]. We search temperature in [1e-5, 5e-4, 1e-4, 5e-3, 1e-3, 1e-2].

\paragraph{Entropy-MCMC (E-MCMC)} We use a cosine learning rate decay scheme, annealing the learning rate to zero. We run 300 epochs. We search $\eta$ in [1e-4, 5e-3, 1e-3, 5e-2, 1e-2, 1e-1] and a system temperature $T$ in [1e-4, 5e-4, 1e-3, 5e-3, 1e-2]. Note that the $\eta$ handles flatness, and the system temperature adjusts the weight update's step size.

\paragraph{SWAG} We use a cosine learning rate decay scheme for SWAG. All the range of hyperparameters is shared with SGD with Momenmtum. Additionally, we search for three additional hyperparameters for SWAG, capturing DNN snapshots and calculating statistics. First, the epoch to start SWA is in [161, 201], and epoch is 300. Second, the frequency of capturing the model snapshot is in [1, 2, 3]. Third, the low rank for covariance is in [2, 3, 5, 7, 10].

\paragraph{F-SWAG} F-SWAG shares hyperparameter with SWAG, except $\gamma$. We search $\gamma$ in [1e-2, 5e-2, 1e-1].

\paragraph{FP-BMA} In case of FP-BMA (VI), we set $\mathcal{N}(0, 1e-3)$ as prior and $\delta$ as 1e-3 to make DNN to BNN using MOPED. After getting prior distribution, we search three hyperparameters: learning rate and $\gamma$. The hyperparameter tuning range included: learning rate in [1e-3, 5e-3, 1e-2, 5e-2], $\gamma$ in [1e-2, 5e-2, 1e-1, 5e-1]. We set weight decay as $5e-4$ for all backbones and train the model over 300 epochs with early stopping. We fix $\beta$ as 1e-8 for all experiments. In case of FP-BMA (MCMC), we search learning rate, temperature for learning rate scheduling, and $\gamma$. The hyperparameter ranges are [1e-3, 5e-3, 1e-2, 5e-2] for learning rate, [1e-4, 5e-3, 1e-3, 5e-2, 1e-2, 1e-1] for temperature, and [5e-3, 1e-2, 5e-2, 1e-1, 5e-1] for $\gamma$. In case of FP-BMA (SWAG), we follow the hyperparameter for SWAG, except $\gamma$ in [1e-2, 5e-2, 1e-1].

\clearpage
\subsection{Bayesian Transfer Learning}\label{subsec:few_shot_image_classification_with_bayesian_transfer_learning_app}
\subsubsection{FP-BMA with diverse BNN frameworks}\label{subsubsec:sabma_with_diverse_bnn_frameworks_2}
Diverse BNN frameworks can be adopted for Bayesian Transfer Learning. Specifically, there are several options for making pre-trained DNN into BNN. In this work, we mainly adopt MOPED and SWAG for the converting.

In addition, FP-BMA can be applied with various BNN frameworks by using an empirical loss function
$\ell(\cdot)$ and adjusting the parameter $\beta$ in Eq.~\ref{eq:bayesian_transfer_learning}. We commonly set $\ell(\cdot)$ as cross-entropy loss in context of image classification task.

\paragraph{FP-BMA (VI)}
First, we convert pre-trained DNN into BNN with MOPED. We set the converted BNN as prior, $q_\theta^{\text{pr}}(w|\mathcal{D}^{\text{pr}})$ in Eq.~\ref{eq:bayesian_transfer_learning}, and initial point of model. We only train parameters of normalization and last layer and freeze others. We train them with the loss function of Eq.~\ref{eq:bayesian_transfer_learning}.

\paragraph{FP-BMA (MCMC)}
For SGLD, it is unnecessary to convert pre-trained DNN into BNN. Instead, we directly set the pre-trained DNN as initialization. We incorporated noise into Eq.~\ref{eq:bayesian_transfer_learning} without the KLD term ($\beta = 0$) based on the learning rate and the hyperparameter, temperature. During the first step, the adversarial posterior is computed without any
noise (Eq.~\ref{eq:Delta_theta_star}). In the second step, both the noise and the adversarial posterior are used together in the learning process.

\paragraph{FP-BMA (SWAG)}
SWAG is also one of the options to convert pre-trained DNN into BNN. Specifically, we run a few epochs with source or downstream datasets to make BNN from pre-trained DNN. After this step, we set the BNN as the prior, $q_\theta^{\text{pr}}(w|\mathcal{D}^{\text{pr}})$ in Eq.~\ref{eq:bayesian_transfer_learning}. We also let the converted BNN as initialization and train with downstream dataset. We optimize model with the loss function in Eq.~\ref{eq:bayesian_transfer_learning}.

\subsubsection{Hyperparameters for Experiments}\label{subsubsec:hyperparameters_for_experiments}
In this section, we provide the details of the experimental setup for Section~\ref{subsec:few-shot_image_classification}. In the other experiments, the range of hyperparameters, excluding the number of epochs, is shared across different backbones and methods.

First, we provide remarks for each baseline method, followed by the tables of hyperparameter configuration with respect to downstream datasets and the baselines. For all experiments, the hyperparameters are selected using grid-search. Configuration of best hyperparameters for each baseline is summarized in Table~\ref{tab:hyperparameter_c10} and Table~\ref{tab:hyperparameter_c100}. We ran all experiments using GeForce RTX 3090 and NVIDIA RTX A6000 with GPU memory of 24,576MB and 49,140 MB.

\begin{table*}[ht!]
\caption{Hyperparameter Configuration for CIFAR10}
\label{tab:hyperparameter_c10}
\adjustbox{width=1.0\linewidth}{
\begin{tabular}{cccccccc}
\toprule
Backbone                     & Baseline      & learning rate       & \begin{tabular}[c]{@{}c@{}}$\beta_1$\\ (momentum)\end{tabular} & $\beta_2$ & $\gamma$      & weight decay       \\ \midrule\midrule
\multirow{9}{*}{RN18}  & SGD           & 5e-3 & 9e-1 &  $\times$     & $\times$         & 1e-3    \\
                                     & SAM & 1e-2     & 9e-1 &  $\times$     & 1e-1      & 1e-4   \\
                                     & FSAM & 1e-2 & 9e-1 & $\times$ & 1e-1    & 1e-4    \\
                                     & bSAM & 1e-1 & 9e-1 & 0.999 & 5e-2    & 1e-1    \\
                                     & MOPED   & 1e-2 & 9e-1 & $\times$ & $\times$ & 1e-4 \\
                                     & \textbf{FP-BMA (VI)} & 1e-2 & 9e-1 & $\times$ & 7e-1 & 1e-3 \\
                                     & MCMC   & 5e-2 & 9e-1 & $\times$ &  $\times$        & 5e-4 \\
                                     & PTL & 1e-1 &  $\times$ & $\times$ & $\times$ & 1e-3    \\
                                     & E-MCMC & 5e-2 & $\times$ & $\times$ & $\times$ & 1e-3 \\
                                     & \textbf{FP-BMA (MCMC)} & 5e-3 & 9e-1 & $\times$ & 8e-3 & 5e-4 \\
                                     & SWAG & 5e-3 & 9e-1 & $\times$ & $\times$ & 1e-5 \\
                                     & F-SWAG & 5e-3 & 9e-1 & $\times$ & 5e-2 & 5e-4 \\
                                     & \textbf{FP-BMA (SWAG)} & 5e-2 & 9e-1 & $\times$ & 1e-1 & 5e-4 \\ \midrule\midrule
\multirow{9}{*}{ViT-B/16}       & SGD & 1e-3 & 9e-1 & $\times$ & $\times$ & 1e-4 \\
                                     & SAM & 1e-3 & 9e-1 & $\times$ & 1e-2 & 1e-3 \\
                                     & FSAM & 5e-3 & 9e-1 & $\times$ & 1e-2    & 1e-3     \\
                                     & bSAM & 1e-1 & 9e-1 & 0.999 & 1e-2 & 1e-1 \\
                                     & MOPED & 1e-3 & 9e-1 & $\times$ & $\times$ & 1e-4 \\
                                     & \textbf{FP-BMA (VI)} & 1e-2 & 9e-1 & $\times$ & 1e-1 & 5e-4 \\
                                     & MCMC & 3e-2 & 9e-1 & $\times$ & $\times$ &  5e-4 \\
                                     & PTL & 6e-2 & $\times$ & $\times$ & $\times$ & 1e-3    \\
                                     & EMCMC & 5e-3 & $\times$ & $\times$ & $\times$ & 1e-2     \\
                                     & \textbf{FP-BMA (MCMC)} & 5e-3 & 9e-1 & $\times$ & 8e-3 & 5e-4 \\
                                     & SWAG & 1e-3 & 9e-1 & $\times$ & $\times$ & 1e-3 \\
                                     & F-SWAG & 1e-3 & 9e-1 & $\times$ & 1e-2     & 1e-3 \\
                                     & \textbf{FP-BMA (SWAG)} & 5e-3 & 9e-1 & $\times$ & 5e-1 & 5e-4 \\ \bottomrule
\end{tabular}
}
\end{table*}
\begin{table*}[h]
\caption{Hyperparameter Configuration for CIFAR100}
\label{tab:hyperparameter_c100}
\adjustbox{width=1.0\linewidth}{
\begin{tabular}{cccccccc}
\toprule
Backbone                     & Baseline      & learning rate       & \begin{tabular}[c]{@{}c@{}}$\beta_1$\\ (momentum)\end{tabular} & $\beta_2$ & $\gamma$      & weight decay       \\ \midrule\midrule
\multirow{9}{*}{RN18}  & SGD         & 1e-2 & 9e-1 & $\times$ & $\times$ & 5e-3 \\
                        & SAM & 1e-2 & 9e-1 & $\times$ & 5e-2 & 1e-2 \\
                     & FSAM & 1e-2 & 9e-1 & $\times$ & 1e-1    & 1e-4    \\
                                     & bSAM & 1 & 9e-1 & 0.999 & 1e-2 & 1e-2 \\
                                     & MOPED   &  1e-2 & 9e-1 & $\times$ & $\times$ & 1e-3 \\
                                     & \textbf{FP-BMA (VI)} & 5e-2 & 9e-1 & $\times$ & 1e-2 & 5e-4 \\
                                     & MCMC   & 3e-2 & 9e-1 & $\times$ &  $\times$ & 5e-4 \\
                                     & PTL & 5e-1 & $\times$ & $\times$ & $\times$ & 1e-3 \\
                                     & E-MCMC & 5e-2 & $\times$ & $\times$ & $\times$ & 1e-3 \\
                                     & \textbf{FP-BMA (MCMC)} & 1e-2 & 9e-1 & $\times$ & 1e-1 & 5e-4 \\
                                     & SWAG & 1e-2 & 9e-1 & $\times$ & $\times$ & 1e-4 \\
                                     & F-SWAG & 1e-2 & 9e-1 & $\times$ & 5e-2 & 1e-2 \\
                                     & \textbf{FP-BMA (SWAG)} & 5e-2 & 9e-1 & $\times$ & 5e-1 & 5e-4 \\ \midrule\midrule
\multirow{9}{*}{ViT-B/16}  & SGD    & 1e-3 & 9e-1 & $\times$ & $\times$ & 1e-2 \\
                        & SAM & 1e-3 & 9e-1 & $\times$ & 1e-2 & 1e-2 \\
                     & FSAM & 5e-3 & 9e-1 & $\times$ & 1e-2 & 1e-4 \\
                                     & bSAM & 2.5e-1 & 9e-1 & 0.999 & 1e-2 & 1e-3 \\
                                     & MOPED   &  1e-3 & 9e-1 & $\times$ & $\times$ & 1e-3 \\
                                     & \textbf{FP-BMA (VI)} & 1e-2 & 9e-1 & $\times$ & 5e-2 & 5e-4 \\
                                     & MCMC   & 5e-2 & 9e-1 & $\times$ &  $\times$        & 5e-4 \\
                                     & PTL & 1e-1 & $\times$ & $\times$ & $\times$ & 1e-3 \\
                                     & E-MCMC & 5e-2 & $\times$ & $\times$ & $\times$ & 1e-3 \\
                                     & \textbf{FP-BMA (MCMC)} & 8e-3 & 9e-1 & $\times$ & 8e-3 & 5e-4 \\
                                     & SWAG & 1e-3 & 9e-1 & $\times$ & $\times$ & 1e-2 \\
                                     & F-SWAG & 1e-3 & 9e-1 & $\times$ & 1e-2 & 1e-2 \\
                                     & \textbf{FP-BMA (SWAG)} & 1e-2 & 9e-1 & $\times$ & 5e-1 & 5e-4 \\ \bottomrule
\end{tabular}
}
\end{table*}

\paragraph{Stochastic Gradient Descent with Momentum (SGD)}
In this study, we adopt Stochastic Gradient Descent with Momentum as an optimizer for DNN. Learning rate schedule is fixed to cosine decay with warmup length of 10. We tested [100, 150] epoch and set 100 epoch as the best option. In overall experiments, we set momentum as 0.9. The hyperparameter tuning range included learning rate in [1e-4, 1e-3, 1e-2], and weight decay in [1e-4, 5e-4, 1e-3, 1e-2].

\paragraph{Sharpness Aware Minimization (SAM)}
We set SGD with momentum as the base optimizer of SAM. It also ran upon a cosine decay learning rate scheduler. All the range of hyperparameters is shared with SGD with Momenmtum. Additional hyperparameter $\gamma$, the ball size of perturbation, is in [1e-2, 5e-2, 1e-1].

\paragraph{Fisher SAM (FSAM)}
We set SGD with momentum as the base optimizer of FSAM. It also ran upon a cosine decay learning rate scheduler. All the range of hyperparameters is shared with SGD with Momenmtum. Additional hyperparameter $\eta$, regularize Fisher impact, is in [1e-2, 1e-1, 1].

\paragraph{SAM as an optimal relaxation of Bayes (bSAM)} We use a cosine learning rate decay scheme, annealing the learning rate to zero. We fine-tuned pre-trained models for 150 epochs with fixed $\beta_1$ and $\beta_2$. The hyperparameter tuning range included: learning rate in [1e-3, 1e-2, 5e-2, 1e-1, 0.25, 0.5, 1], weight decay in [1e-3, 1e-2, 1e-1], damping in [1e-3, 1e-2, 1e-1], noise scaling parameter in [1e-4, 1e-3, 1e-2, 1e-1], and $\gamma$ in [1e-3, 1e-2, 5e-2, 1e-1]. Damping parameter stabilizes the method by adding constant when updating variance estimate. Since SAM as Bayes optimizer depends on the number of samples to scale the prior, we introduced additional noise scaling parameters to mitigate the gap between the experimental settings, where SAM as Bayes assumed training from scratch and our method assumed few-shot fine-tuning on the pre-trained model. We multiplied noise scaling parameter to the variance of the Gaussian noise to give strong prior, assuming pre-trained model.

\paragraph{Model Priors with Empirical Bayes using DNN (MOPED)} MOPED was a baseline to compare for Bayesian Transfer Learning. It employs pre-trained DNN and transforms it into Mean-Field Variational Inference (MFVI). We set prior mean and variance as 0 and 1, respectively. Besides, we set the posterior mean as 0 and variance as 1e-3. We adopt Reparameterization as type of VI. The essential hyperparameter for MOPED is $\delta$, which adjusts how much to incorporate pre-trained weights. The $\delta$ was searched in [5e-2, 1e-1, 2e-1]. Moreover, we add a hyperparameter $\beta$ for MOPED that can balance the loss term in VI. The $\beta$ is in range [1e-2, 1e-1, 1].

\paragraph{MCMC} We consistently use SGLD~\citep{welling2011bayesian} for MCMC in this work. It ran upon a cyclic cosine decay learning rate scheduler. The number of cycles was ranged in [2, 4]. The number of sampled models is in [10, 20, 28]. We search temperature in [1e-5, 1e-4, 1e-3, 1e-2, 1e-1, 1].

\paragraph{Pre-train Your Loss (PTL)} The backbones both ResNet18 and Vit-B/16 were refined through fine-tuning with a classification head for the target task, leveraging a prior distribution learned from SWAG on the ImageNet 1k dataset using SGD. First, the hyperparameter tuning range of the pre-training epoch is [2, 3, 5, 15, 30] to generate the prior distribution on the source task, ImageNet 1k. The learning rate was 0.1. We approximated the covariance low rank as 5. Second, in the downstream task, the fine-tuning optimizer is SGLD with a cosine learning rate schedule, sampling 30 in 5 cycles. The hyperparameter tuning range included: learning rate in [1e-4, 1e-3, 1e-2, 5e-2, 6e-2, 1e-1, 5e-1], weight decay in [1e-4, 1e-3 ,1e-2 ,1e-1], and prior scale in [1e+4, 1e+5, 1e+6]. Prior scaling in the downstream task is to reflect the mismatch between the pre-training and downstream tasks and to add coverage to parameter settings that might be consistent with the downstream. Training was conducted over 150 epochs; tuning range of fine-tuning epoch is [100, 150, 200, 300, 1000].

\paragraph{Entropy-MCMC (E-MCMC)} We use a cosine learning rate decay scheme, annealing the learning rate to zero. We set the range of the hyperparameter sweep to the surroundings of the best hyperparameter in E-MCMC for ResNet18: learning rate in [5e-3, 5e-2, 5e-1], weight decay in [1e-4, 1e-3, 1e-2], $\eta$ in [1e-6, 5e-6, 1e-5, 5e-5, 1e-4, 4e-4, 5e-3, 8e-3, 1e-2] and a system temperature $T$ in [1e-5, 1e-4, 1e-3]. In this study, we performed an extensive exploration of the hyperparameter space of ViT-B/16, as it has a mechanism different from the CNN family and may not be found near the best hyperparameter range of ResNet18: learning rate in [1e-3, 5e-3, 1e-2, 5e-2, 5e-1], weight decay in [1e-5, 1e-4, 5e-4, 1e-3, 1e-2, 5e-2], $\eta$ in [5e-7, 1e-6, 5e-6, 5e-5, 1e-4, 4e-4, 5e-4, 1e-3, 8e-3, 1e-2, 1e-1] and a system temperature $T$ in [1e-6, 5e-6, 1e-5, 5e-5, 1e-4, 1e-3, 1e-2, 1e-1]. We fine-tuned pre-trained models for 150 epochs. Note that the $\eta$ handles flatness, and the system temperature adjusts the weight update's step size.

\paragraph{SWAG} We use a cosine learning rate decay scheme for SWAG. All the range of hyperparameters is shared with SGD with Momenmtum. Additionally, we search three additional hyperparameters for SWAG, capturing DNN snapshots and calculating statistics. First, the epoch to start SWA is in [51, 76, 101] and epoch is in [100, 150]. Second, the frequency to capture the model snapshot is in [1, 2, 3]. Third, the low rank for covariance is in [2, 3, 5, 7, 10].

\paragraph{F-SWAG} F-SWAG shares hyperparameter with SWAG, except $\gamma$. We search $\gamma$ in [1e-2, 5e-2, 1e-1].

\paragraph{FP-BMA} In case of FP-BMA (SWAG), we train SWAG on source task IN 1K to make prior distribution and follow the pre-training protocol of PTL. In case of employing MOPED to make prior distribution, we do not go through any training step. In case of FP-BMA (VI), we just set $\delta$ as 0.05 for MOPED and make DNN into BNN. In case of FP-BMA (MCMC), we just set pre-trained weight as initialization and run experiments. After getting prior distribution, we search three hyperparameters: learning rate, $\gamma$, and $\alpha$. The hyperparamter tuning range included: learning rate in [1e-3, 5e-3, 1e-2, 5e-2], $\gamma$ in [5e-3, 8e-3, 1e-2, 5e-2, 1e-1, 5e-1, 7e-1], and $\alpha$ in [1e-6, 1e-5, 1e-4, 1e-3]. We set weight decay as $5e-4$ for all backbones and train the model over 150 epochs with early stopping. We fix $\beta$ as 1e-8 for all experiments.

\subsection{Algorithm of FP-BMA}\label{subsec:algorith_of_fpbma}
\begin{algorithm}
\caption{FP-BMA with Bayesian Transfer Learning}\label{alg:FP-BMA}
\begin{algorithmic}
\Require Variational parameter $\theta$, Neighborhood size $\gamma$, Epochs $E$, and Learning rate $\eta_{\text{FP-BMA}}$ 
\State 1) Load pre-trained DNN
\State 2) Make pre-trained DNN model into BNN $q_\theta^{\text{pr}}(w|\mathcal{D}^{\text{pr}})$ and set as prior
\For{$t \ = \ 1, 2, ... , E$}
    \State 3-1) $w \sim q_\theta(w|\mathcal{D}^{\text{ft}})$\Comment{Sample weight from posterior}
    \State 3-2) Forward and calculate the loss $\ell(\theta)$ with the sampled $w$ 
    \State 3-3) Backward pass and compute $\nabla_\theta \log q_\theta (w|\mathcal{D})$
    \State 3-4) Compute $F_\theta^{-1} (\theta) = \frac{\nabla_\theta \log q_\theta (w|\mathcal{D}) \nabla_\theta \log q_\theta (w|\mathcal{D})^T}{\| \nabla_\theta \log q_\theta(w|\mathcal{D})\|^4}$
    \State 3-5) Compute the perturbation $\Delta\theta_{\text{FP-BMA}} = \gamma \frac{F_\theta(\theta)^{-1} \nabla_\theta \ell(\theta)}{\sqrt{\nabla_\theta \ell(\theta)^T F_\theta(\theta)^{-1} \nabla_\theta \ell(\theta)}}$
    \State 3-6) Compute gradient approximation for the FP-BMA $\nabla_\theta  \ell^\gamma_{\text{FP-BMA}} (\theta) = \frac{\partial \ell(\theta)}{\partial \theta} |_{\theta + \Delta\theta_{\text{FP-BMA}}}$ \\ 
    \State 3-7) Update $\theta \rightarrow \theta - \eta\nabla_\theta \ell_{\text{FP-BMA}}(\theta)$
\EndFor
\end{algorithmic}
\end{algorithm}

Training algorithm of FP-BMA with Bayesian Transfer Learning can be depicted as Algorithm~\ref{alg:FP-BMA}. In the first step, load a model pre-trained on the source task. Note that the pre-trained models do not have to be BNN. Namely, it is capable of using DNN, which can be easier to find than pre-trained BNN. Second, change the loaded DNN into BNN on the source or downstream task. Every BNN framework can be adopted to make DNN into BNN. We can skip this second step if you load a pre-trained BNN model before. Third, train the subnetwork of the converted BNN model with the proposed flat-seeking seeking optimizer. It allows model to converge into flat minina efficiently.

\subsection{Efficiency of FP-BMA}\label{subsec:efficieny_of_fpbma}
\begin{wraptable}{r}{0.4\textwidth}
\centering
\vspace{-4.5em}
\begin{tabular}{lccc}
\toprule
\textbf{Method} & \textbf{Time} & \textbf{Wall} & \textbf{Mem.} \\
                & \textbf{Comp.} & \textbf{Clock} & \textbf{Comp.} \\
\midrule
SGD & $O(p)$ & 2.78s & $O(p)$ \\
SAM & $O(2p)$ & 4.58s & $O(p)$ \\
FSAM & $O(2p)$ & 4.65s & $O(2p)$ \\
bSAM & $O(2p)$ & 4.62s & $O(3p)$ \\ \midrule
MF VI & $O(2p)$ & 4.09s & $O(2p)$ \\
FF VI & $O(p^2)$ & -- & $O(p^2)$ \\ \midrule
MCMC & $O(p)$ & 2.95s & $O(Mp)$ \\
E-MCMC & $O(2p)$ & 5.13s & $O(Mp)$ \\ \midrule
SWAG & $O(p)$ & 7.89s & $O(Kp)$ \\
F-SWAG & $O(2p)$ & 11.48s & $O(Kp)$ \\
\textbf{FP-BMA} & $O(2p)$ & 6.21s & $O(Kp_1)$ \\
\bottomrule
\end{tabular}
\caption{Time and memory complexity for all methods.}
\label{tab:efficiency}
\vspace{-9em}
\end{wraptable}

The following Table~\ref{tab:efficiency} summarizes the per-epoch wall-clock time, theoretical time complexity, and memory usage across methods under a unified experimental setting. Evaluation conducted on ResNet-18 with CIFAR-10 10-shot classification. AMP (automatic mixed precision) was enabled for fair efficiency comparison.

\textbf{Notation:}
\begin{itemize}
    \item $p$: total number of model parameters
    \item $p_1$: number of trainable parameters used in FP-BMA subnetwork ($p_1 \ll p$)
    \item $M$: number of MCMC samples
    \item $K$: rank for low-rank approximations (e.g., in SWAG or FP-BMA)
\end{itemize}

To ensure practical efficiency, FP-BMA is implemented with a subnetwork strategy and inverse vector product approximation (as shown in Algorithm~\ref{alg:FP-BMA}). These design choices allow us to limit both runtime and memory overhead, which we found to be comparable to standard baselines.

\subsection{Fine-Grained Image Classification}\label{subsec:fine-grained_image_classification_app}
In addition to classification accuracy, FP-BMA shows superior performance compared to the baseline in NLL metric, indicating that FP-BMA effectively quantifies uncertainty.
\begin{table}[h]
\caption{Downstream task NLL with RN50 and ViT-B/16 pre-trained on IN 1K. FP-BMA (SWAG) denotes using SWAG to convert pre-trained model into BNN. \textbf{Bold} and \underline{underline} denote best and second best performance each. FP-BMA demonstrates superior performance across all 16-shot datasets, including EuroSAT~, Oxford Flowers, Oxford Pets, and UCF101.}
\label{tab:fine-grained_nll}
\adjustbox{width=1.0\linewidth}{
\begin{tabular}{cccccacccca}
\toprule
Backbone &
  \multicolumn{5}{c}{RN50} &
  \multicolumn{5}{c}{ViT-B/16} \\ \cmidrule(lr){2-6} \cmidrule(lr){7-11}
Method  &
  EuroSAT &
  Oxford Flowers &
  Oxford Pets &
  UCF101 &
  Avg &
  EuroSAT &
  Oxford Flowers &
  Oxford Pets &
  UCF101 &
  Avg \\ \midrule
SGD &
  $0.416_{\pm 0.043}$ &
  $0.265_{\pm 0.010}$ &
  $0.367_{\pm 0.008}$ &
  $1.331_{\pm 0.024}$ &
  $0.595_{\pm 0.010}$ &
  $0.573_{\pm 0.044}$ &
  $0.361_{\pm 0.027}$ &
  $0.385_{\pm 0.044}$ &
  $1.246_{\pm 0.044}$ &
  $0.641_{\pm 0.020}$ \\
SAM &
  $0.376_{\pm 0.003}$ &
  $\underline{0.190}_{\pm 0.001}$ &
  $\underline{0.344}_{\pm 0.014}$ &
  $\underline{1.157}_{\pm 0.035}$ &
  $0.517_{\pm 0.005}$ &
  $0.522_{\pm 0.023}$ &
  $\underline{0.276}_{\pm 0.029}$ &
  $\underline{0.287}_{\pm 0.022}$ &
  $\underline{1.140}_{\pm 0.034}$ &
  $\underline{0.556}_{\pm 0.020}$ \\
SWAG &
  $0.343_{\pm 0.046}$ &
  $0.264_{\pm 0.011}$ &
  $0.367_{\pm 0.007}$ &
  $1.347_{\pm 0.022}$ &
  $0.580_{\pm 0.009}$ &
  $0.547_{\pm 0.021}$ &
  $0.361_{\pm 0.027}$ &
  $0.366_{\pm 0.010}$ &
  $1.286_{\pm 0.045}$ &
  $0.640_{\pm 0.006}$ \\
F-SWAG &
  $\underline{0.301}_{\pm 0.039}$ &
  $0.190_{\pm 0.002}$ &
  $0.351_{\pm 0.010}$ &
  $1.186_{\pm 0.034}$ &
  $\underline{0.507}_{\pm 0.008}$ &
  $0.514_{\pm 0.018}$ &
  $0.276_{\pm 0.033}$ &
  $0.297_{\pm 0.030}$ &
  $1.234_{\pm 0.031}$ &
  $0.580_{\pm 0.017}$ \\
MOPED &
  $0.481_{\pm 0.100}$ &
  $0.347_{\pm 0.019}$ &
  $0.388_{\pm 0.007}$ &
  $1.367_{\pm 0.029}$ &
  $0.646_{\pm 0.028}$ &
  $\underline{0.484}_{\pm 0.018}$ &
  $0.354_{\pm 0.025}$ &
  $0.309_{\pm 0.015}$ &
  $1.180_{\pm 0.028}$ &
  $0.582_{\pm 0.017}$ \\
PTL &
  $0.319_{\pm 0.006}$ &
  $0.307_{\pm 0.010}$ &
  $0.360_{\pm 0.015}$ &
  $1.391_{\pm 0.036}$ &
  $0.594_{\pm 0.010}$ &
  $0.493_{\pm 0.012}$ &
  $0.616_{\pm 0.066}$ &
  $0.381_{\pm 0.008}$ &
  $1.670_{\pm 0.050}$ &
  $0.790_{\pm 0.013}$ \\
FP-BMA &
  $\boldsymbol{0.297}_{\pm 0.038}$ &
  $\boldsymbol{0.147}_{\pm 0.037}$ &
  $\boldsymbol{0.339}_{\pm 0.023}$ &
  $\boldsymbol{1.113}_{\pm 0.009}$ &
  $\boldsymbol{0.474}_{\pm 0.023}$ &
  $\boldsymbol{0.455}_{\pm 0.006}$ &
  $\boldsymbol{0.219}_{\pm 0.037}$ &
  $\boldsymbol{0.272}_{\pm 0.006}$ &
  $\boldsymbol{1.071}_{\pm 0.036}$ &
  $\boldsymbol{0.504}_{\pm 0.012}$ \\ \bottomrule
\end{tabular}
}
\end{table}

\subsection{Performance under Distribution shift}\label{subsec:performance_under_distribution_shift_app}
\begin{figure}[h]
    \centering
    \includegraphics[width=1\textwidth]{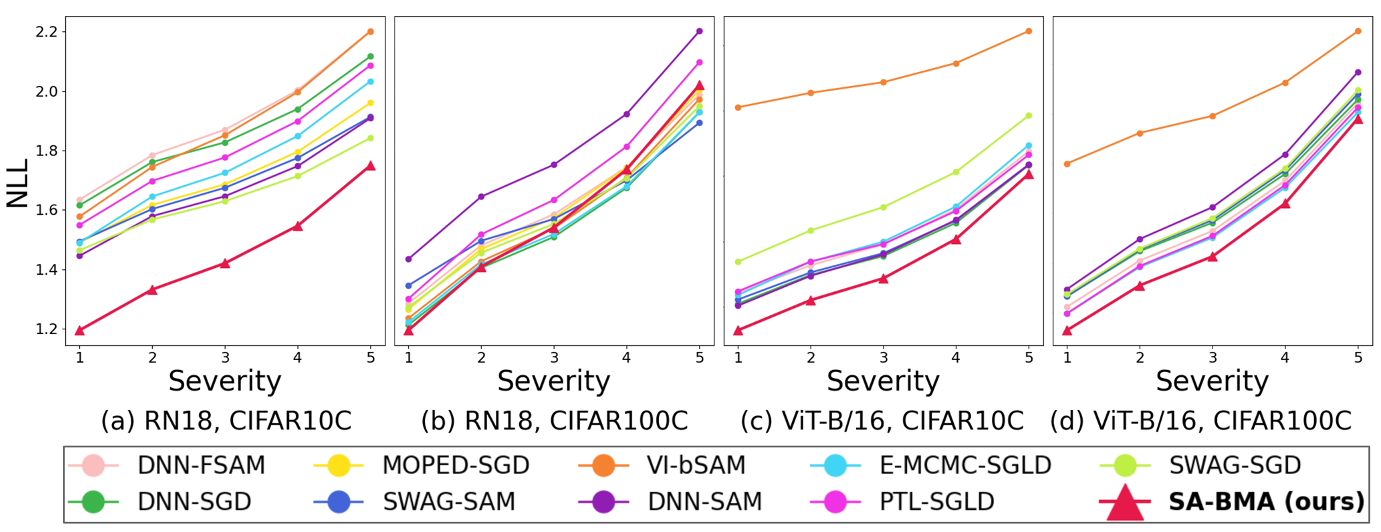}
    \caption{NLL performance of ResNet 18 and ViT-B/16 on corrupted CIFAR10 and CIFAR100, respectively \citep{hendrycks2019benchmarking}.}
    \label{fig:severity_nll}
\end{figure}

We adopt the corrupted dataset CIFAR10/100C to test the robustness over distribution shift. The corrupted dataset transform the CIFAR10/100-test dataset, which has been modified to shift the distribution of the test data further away from the training data. It contains 19 kinds of corrupt options, such as varying brightness or contrast to adding Gaussian noise. The severity level indicates the strength of the transformation and is typically expressed as a number from 1 to 5, where the higher the number, the stronger the transformation. In Figure~\ref{fig:severity_nll}, our method ensures relatively robust performance in the data distribution shift, even as the severity increases.

We also provide the detailed results of three repeated experiments with corrupted sets.

\begin{figure}[h]
    \centering
    \begin{subtable}[t]{1.0\textwidth}
        \centering
        \caption{RN18 CIFAR10C}
        \label{tab:seve-rn18-c10c}
        \adjustbox{width=1.0\linewidth}{
        \begin{tabular}{ccccccccccc}
        \toprule
        \multirow{3}{*}{Method} & \multicolumn{10}{c}{Severity} \\ \cmidrule(lr){2-11} 
                                & \multicolumn{2}{c}{1}  & \multicolumn{2}{c}{2}  & \multicolumn{2}{c}{3}  & \multicolumn{2}{c}{4}  & \multicolumn{2}{c}{5} \\ 
                                \cmidrule(lr){2-3} \cmidrule(lr){4-5} \cmidrule(lr){6-7} \cmidrule(lr){8-9} \cmidrule(lr){10-11} 
                                & ACC $\uparrow$   & NLL  $\downarrow$  & ACC $\uparrow$  & NLL $\downarrow$ & ACC $\uparrow$  & NLL $\downarrow$ & ACC $\uparrow$  & NLL $\downarrow$ & ACC $\uparrow$   & NLL $\downarrow$ \\ \midrule
        SGD                    & $49.57_{\pm{0.97}}$          & $1.49_{\pm{0.02}}$          & $45.78_{\pm{1.43}}$          & $1.62_{\pm{0.04}}$          & $43.78_{\pm{1.44}}$          & $1.69_{\pm{0.04}}$          & $40.83_{\pm{1.59}}$          & $1.80_{\pm{0.06}}$          & $36.30_{\pm{1.79}}$          & $1.96_{\pm{0.08}}$          \\
        SAM                    & $50.23_{\pm{2.11}}$          & $1.62_{\pm{0.07}}$          & $46.56_{\pm{2.00}}$          & $1.76_{\pm{0.03}}$          & $44.59_{\pm{2.26}}$          & $1.83_{\pm{0.03}}$          & $41.85_{\pm{2.42}}$          & $1.94_{\pm{0.04}}$          & $37.33_{\pm{2.52}}$          & $2.12_{\pm{0.07}}$          \\
        FSAM                   & $48.76_{\pm{4.00}}$          & $1.63_{\pm{0.03}}$          & $45.11_{\pm{3.91}}$          & $1.78_{\pm{0.01}}$          & $42.94_{\pm{3.88}}$          & $1.87_{\pm{0.03}}$          & $40.06_{\pm{3.85}}$          & $2.00_{\pm{0.08}}$          & 3$5.70_{\pm{3.50}}$          & $2.20_{\pm{0.12}}$          \\ \midrule
        SWAG & $50.05_{\pm{0.76}}$          & $1.55_{\pm{0.09}}$          & $46.31_{\pm{1.16}}$          & $1.70_{\pm{0.11}}$          & $44.17_{\pm{1.07}}$          & $1.78_{\pm{0.11}}$          & $41.20_{\pm{1.13}}$          & $1.90_{\pm{0.13}}$          & $36.64_{\pm{1.26}}$          & $2.09_{\pm{0.15}}$          \\
        F-SWAG & $51.37_{\pm{1.08}}$          & $1.49_{\pm{0.05}}$          & $47.35_{\pm{0.71}}$          & $1.64_{\pm{0.04}}$          & $45.16_{\pm{0.66}}$          & $1.72_{\pm{0.06}}$          & $42.01_{\pm{0.57}}$          & $1.85_{\pm{0.06}}$          & $37.27_{\pm{0.64}}$          & $2.03_{\pm{0.07}}$          \\
        bSAM                   & $49.20_{\pm{2.40}}$          & $1.46_{\pm{0.05}}$          & $45.35_{\pm{1.93}}$          & $1.57_{\pm{0.04}}$          & $43.07_{\pm{2.10}}$          & $1.63_{\pm{0.04}}$          & $40.12_{\pm{1.74}}$          & $1.71_{\pm{0.03}}$          & $35.50_{\pm{1.36}}$          & $1.84_{\pm{0.02}}$          \\
        VI & $50.72_{\pm{0.80}}$          & $1.58_{\pm{0.11}}$          & $46.87_{\pm{0.32}}$          & $1.74_{\pm{0.11}}$          & $44.52_{\pm{0.39}}$          & $1.85_{\pm{0.12}}$          & $41.38_{\pm{0.29}}$          & $2.00_{\pm{0.12}}$          & $36.73_{\pm{0.17}}$          & $2.20_{\pm{0.10}}$          \\
        E-MCMC & $49.86_{\pm{1.54}}$          & $1.49_{\pm{0.03}}$          & $46.17_{\pm{1.55}}$          & $1.60_{\pm{0.04}}$          & $44.07_{\pm{1.72}}$          & $1.67_{\pm{0.07}}$          & $41.05_{\pm{1.65}}$          & $1.77_{\pm{0.10}}$          & $36.53_{\pm{1.74}}$          & $1.91_{\pm{0.13}}$          \\
        PTL & $50.44_{\pm{1.65}}$          & $1.45_{\pm{0.06}}$          & $46.22_{\pm{1.96}}$          & $1.58_{\pm{0.09}}$          & $44.06_{\pm{1.67}}$          & $1.65_{\pm{0.09}}$          & $41.02_{\pm{1.66}}$          & $1.75_{\pm{0.11}}$          & $36.14_{\pm{1.51}}$          & $1.91_{\pm{0.13}}$          \\ \midrule
        FP-BMA                  & $\textbf{58.53}_{\pm{0.75}}$ & $\textbf{1.19}_{\pm{0.02}}$ & $\textbf{53.72}_{\pm{0.70}}$ & $\textbf{1.33}_{\pm{0.00}}$ & $\textbf{50.61}_{\pm{0.84}}$ & $\textbf{1.42}_{\pm{0.01}}$ & $\textbf{46.76}_{\pm{1.15}}$ & $\textbf{1.55}_{\pm{0.03}}$ & $\textbf{40.70}_{\pm{1.34}}$ & $\textbf{1.75}_{\pm{0.05}}$ \\ \bottomrule
        \end{tabular}%
        }
    \end{subtable}
    \begin{subtable}[t]{1.0\textwidth}
        \caption{RN18 CIFAR100C}
        \label{tab:seve-rn18-c100c}
        \adjustbox{width=1.0\linewidth}{
        \begin{tabular}{ccccccccccc}
        \toprule
        \multirow{3}{*}{Method} & \multicolumn{10}{c}{Severity} \\ \cmidrule(lr){2-11} 
                                & \multicolumn{2}{c}{1}  & \multicolumn{2}{c}{2}  & \multicolumn{2}{c}{3}  & \multicolumn{2}{c}{4}  & \multicolumn{2}{c}{5} \\ 
                                \cmidrule(lr){2-3} \cmidrule(lr){4-5} \cmidrule(lr){6-7} \cmidrule(lr){8-9} \cmidrule(lr){10-11} 
                                & ACC $\uparrow$   & NLL  $\downarrow$  & ACC $\uparrow$  & NLL $\downarrow$ & ACC $\uparrow$  & NLL $\downarrow$ & ACC $\uparrow$  & NLL $\downarrow$ & ACC $\uparrow$   & NLL $\downarrow$ \\ \midrule
        SGD                    & $36.01_{\pm{0.86}}$ & $2.55_{\pm{0.06}}$ & $31.81_{\pm{0.73}}$ & $2.79_{\pm{0.06}}$ & $29.75_{\pm{0.57}}$ & $2.91_{\pm{0.04}}$ & $26.73_{\pm{0.25}}$ & $3.11_{\pm{0.02}}$ & $22.20_{\pm{0.08}}$ & $3.40_{\pm{0.00}}$ \\
        SAM                    & $37.94_{\pm{0.52}}$ & $2.46_{\pm{0.02}}$ & $33.57_{\pm{0.50}}$ & $\textbf{2.69}_{\pm{0.03}}$ & $31.46_{\pm{0.67}}$ & $\textbf{2.82}_{\pm{0.03}}$ & $28.19_{\pm{0.75}}$ & $3.02_{\pm{0.05}}$ & $23.32_{\pm{0.69}}$ & $3.33_{\pm{0.06}}$ \\
        FSAM                   & $36.46_{\pm{0.44}}$ & $2.53_{\pm{0.05}}$ & $32.24_{\pm{0.36}}$ & $2.77_{\pm{0.04}}$ & $30.19_{\pm{0.42}}$ & $2.90_{\pm{0.03}}$ & $27.12_{\pm{0.37}}$ & $3.10_{\pm{0.02}}$ & $22.48_{\pm{0.39}}$ & $3.42_{\pm{0.01}}$ \\
        bSAM                   & $36.20_{\pm{0.59}}$ & $2.73_{\pm{0.03}}$ & $32.48_{\pm{0.34}}$ & $2.99_{\pm{0.03}}$ & $30.66_{\pm{0.33}}$ & $3.12_{\pm{0.02}}$ & $27.94_{\pm{0.14}}$ & $3.32_{\pm{0.05}}$ & $23.66_{\pm{0.29}}$ & $3.66_{\pm{0.06}}$ \\ \midrule
        SWAG & $35.84_{\pm{5.17}}$ & $2.62_{\pm{0.30}}$ & $32.43_{\pm{4.55}}$ & $2.81_{\pm{0.27}}$ & $30.71_{\pm{4.21}}$ & $2.89_{\pm{0.25}}$ & $28.13_{\pm{3.81}}$ & $3.05_{\pm{0.22}}$ & $24.24_{\pm{2.99}}$ & $\textbf{3.29}_{\pm{0.17}}$ \\
        F-SWAG & $37.10_{\pm{0.60}}$ & $2.49_{\pm{0.03}}$ & $32.84_{\pm{0.62}}$ & $2.72_{\pm{0.03}}$ & $30.59_{\pm{0.72}}$ & $2.86_{\pm{0.04}}$ & $27.43_{\pm{0.91}}$ & $3.06_{\pm{0.06}}$ & $22.74_{\pm{0.93}}$ & $3.38_{\pm{0.08}}$ \\
        VI & $38.20_{\pm{0.57}}$ & $2.47_{\pm{0.02}}$ & $33.77_{\pm{0.59}}$ & $2.71_{\pm{0.03}}$ & $31.70_{\pm{0.75}}$ & $2.83_{\pm{0.03}}$ & $28.56_{\pm{0.77}}$ & $\textbf{3.03}_{\pm{0.04}}$ & $23.72_{\pm{0.78}}$ & $3.33_{\pm{0.05}}$ \\
        E-MCMC & $36.49_{\pm{0.89}}$ & $2.57_{\pm{0.06}}$ & $32.25_{\pm{0.76}}$ & $2.83_{\pm{0.06}}$ & $30.22_{\pm{0.63}}$ & $2.97_{\pm{0.05}}$ & $27.17_{\pm{0.38}}$ & $3.19_{\pm{0.03}}$ & $22.54_{\pm{0.27}}$ & $3.54_{\pm{0.01}}$ \\
        PTL & $36.43_{\pm{0.35}}$ & $2.53_{\pm{0.03}}$ & $32.24_{\pm{0.40}}$ & $2.76_{\pm{0.03}}$ & $30.20_{\pm{0.42}}$ & $2.87_{\pm{0.03}}$ & $27.17_{\pm{0.55}}$ & $3.06_{\pm{0.04}}$ & $22.56_{\pm{0.54}}$ & $3.36_{\pm{0.05}}$ \\ \midrule
        FP-BMA                  & $\textbf{39.41}_{\pm{0.72}}$ & $\textbf{2.44}_{\pm{0.04}}$ & $\textbf{35.07}_{\pm{0.64}}$ & $2.70_{\pm{0.05}}$ & $\textbf{32.75}_{\pm{0.71}}$ & $2.86_{\pm{0.05}}$ & $\textbf{29.41}_{\pm{0.67}}$ & $3.10_{\pm{0.05}}$ & $\textbf{24.25}_{\pm{0.70}}$ & $3.44_{\pm{0.05}}$ \\ \bottomrule
        \end{tabular}%
        }
    \end{subtable}
    \begin{subtable}[t]{1.0\textwidth}
        \caption{VIT-B/16 CIFAR10C}
        \label{tab:seve-vit-c10c}
        \adjustbox{width=1.0\linewidth}{
        \begin{tabular}{ccccccccccc}
        \toprule
        \multirow{3}{*}{Method} & \multicolumn{10}{c}{Severity} \\ \cmidrule(lr){2-11} 
                                & \multicolumn{2}{c}{1}  & \multicolumn{2}{c}{2}  & \multicolumn{2}{c}{3}  & \multicolumn{2}{c}{4}  & \multicolumn{2}{c}{5} \\ 
                                \cmidrule(lr){2-3} \cmidrule(lr){4-5} \cmidrule(lr){6-7} \cmidrule(lr){8-9} \cmidrule(lr){10-11} 
                                & ACC $\uparrow$   & NLL  $\downarrow$  & ACC $\uparrow$  & NLL $\downarrow$ & ACC $\uparrow$  & NLL $\downarrow$ & ACC $\uparrow$  & NLL $\downarrow$ & ACC $\uparrow$   & NLL $\downarrow$ \\ \midrule
        SGD                    &  $79.62_{\pm 0.56}$        & $0.64_{\pm 0.06}$         & $76.47_{\pm 0.67}$        & $0.73_{\pm 0.06}$         & $74.10_{\pm 0.83}$        & $0.79_{\pm 0.05}$         & $70.42_{\pm 1.23}$        & $0.90_{\pm 0.05}$         & $64.41_{\pm{1.85}}$        & $1.08_{\pm{0.05}}$         \\
        SAM                    & $79.78_{\pm 0.49}$        & $0.61_{\pm 0.01}$         & $76.59_{\pm 0.64}$        & $0.70_{\pm 0.02}$         & $74.58_{\pm 0.94}$        & $0.75_{\pm 0.02}$         & $71.12_{\pm 1.06}$        & $0.86_{\pm 0.03}$         & $65.26_{\pm 1.46}$        & $1.03_{\pm 0.04}$         \\
        FSAM                   & $79.87_{\pm 0.83}$        & $0.62_{\pm 0.02}$         & $76.78_{\pm 0.78}$        & $0.70_{\pm 0.02}$         & $74.70_{\pm 0.60}$        & $0.76_{\pm 0.01}$         & $71.29_{\pm 0.49}$        & $0.86_{\pm 0.01}$         & $65.53_{\pm 0.56}$        & $1.03_{\pm 0.03}$         \\
        bSAM                   & $78.80_{\pm 1.18}$        & $0.64_{\pm 0.04}$         & $75.43_{\pm 1.14}$        & $0.74_{\pm 0.04}$         & $73.45_{\pm 1.43}$        & $0.80_{\pm 0.04}$         & $70.07_{\pm 1.50}$        & $0.91_{\pm 0.05}$         & $64.21_{\pm 1.57}$        & $1.09_{\pm 0.05}$         \\ \midrule
        SWAG & $76.58_{\pm 1.69}$        & $1.21_{\pm 0.04}$         & $73.45_{\pm 1.98}$        & $1.25_{\pm 0.04}$         & $71.20_{\pm 2.18}$        & $1.29_{\pm 0.04}$         & $67.54_{\pm 2.46}$        & $1.35_{\pm 0.04}$         & $61.65_{\pm 2.82}$        & $1.44_{\pm 0.04}$         \\
        F-SWAG & $81.03_{\pm 2.20}$        & $0.60_{\pm 0.05}$         & $77.73_{\pm 2.63}$        & $0.69_{\pm 0.06}$         & $75.45_{\pm 2.96}$        & $0.76_{\pm 0.07}$         & $71.82_{\pm 3.31}$        & $0.87_{\pm 0.08}$         & $66.05_{\pm 3.59}$        & $1.03_{\pm 0.10}$         \\
        E-MCMC & $78.91_{\pm 2.31}$        & $0.65_{\pm 0.08}$         & $75.78_{\pm 2.36}$        & $0.74_{\pm 0.08}$         & $73.94_{\pm 2.56}$        & $0.79_{\pm 0.09}$         & $70.66_{\pm 2.63}$        & $0.89_{\pm 0.10}$         & $65.07_{\pm 2.77}$        & $1.06_{\pm 0.11}$         \\
        PTL & $76.26_{\pm 2.46}$        & $0.74_{\pm 0.06}$         & $72.36_{\pm 2.41}$        & $0.83_{\pm 0.06}$         & $69.61_{\pm 2.46}$        & $0.90_{\pm 0.07}$         & $65.47_{\pm 2.52}$        & $1.01_{\pm 0.07}$         & $59.04_{\pm 2.26}$        & $1.18_{\pm 0.06}$         \\ \midrule
        FP-BMA & $\textbf{82.89}_{\pm 1.09}$        & $\textbf{0.53}_{\pm 0.04}$         & $\textbf{79.68}_{\pm 1.26}$        & $\textbf{0.62}_{\pm 0.04}$         & $\textbf{77.30}_{\pm 1.43}$        & $\textbf{0.69}_{\pm 0.05}$         & $\textbf{73.41}_{\pm 1.62}$        & $\textbf{0.81}_{\pm 0.06}$         & $\textbf{66.94}_{\pm 1.79}$        & $\textbf{1.01}_{\pm 0.07}$         \\ \bottomrule
        \end{tabular}%
        }
    \end{subtable}
    \begin{subtable}[t]{1.0\textwidth}
        \caption{VIT-B/16 CIFAR100C}
        \label{tab:seve-vit-c100c}
        \adjustbox{width=1.0\linewidth}{
        \begin{tabular}{ccccccccccc}
        \toprule
        \multirow{3}{*}{Method} & \multicolumn{10}{c}{Severity} \\ \cmidrule(lr){2-11} 
                                & \multicolumn{2}{c}{1}  & \multicolumn{2}{c}{2}  & \multicolumn{2}{c}{3}  & \multicolumn{2}{c}{4}  & \multicolumn{2}{c}{5} \\ 
                                \cmidrule(lr){2-3} \cmidrule(lr){4-5} \cmidrule(lr){6-7} \cmidrule(lr){8-9} \cmidrule(lr){10-11} 
                                & ACC $\uparrow$   & NLL  $\downarrow$  & ACC $\uparrow$  & NLL $\downarrow$ & ACC $\uparrow$  & NLL $\downarrow$ & ACC $\uparrow$  & NLL $\downarrow$ & ACC $\uparrow$   & NLL $\downarrow$ \\ \midrule
        SGD                    & $62.19_{\pm 0.52}$          & $1.42_{\pm 0.02}$           & $57.81_{\pm 0.37}$          & $1.61_{\pm 0.02} $          & $55.04_{\pm 0.14}$          & $1.73_{\pm 0.02}$           & $50.73_{\pm 0.24}$          & $1.93_{\pm 0.01}$           & $44.12_{\pm 0.39}$          & $2.24_{\pm 0.01}$           \\
        SAM                    & $61.90_{\pm 0.53}$          & $1.47_{\pm 0.02}$           & $57.49_{\pm 0.43}$          & $1.65_{\pm 0.02}$          & $54.80_{\pm 0.29}$          & $1.76_{\pm 0.01}$           & $50.52_{\pm 0.25}$         & $1.96_{\pm 0.01}$           & $44.04_{\pm 0.24}$          & $2.26_{\pm 0.01}$           \\
        FSAM                   & $61.70_{\pm 0.52}$          & $1.47_{\pm 0.02}$           & $57.16_{\pm 0.44}$          & $1.65_{\pm 0.02}$           & $54.46_{\pm 0.37}$          & $1.77_{\pm 0.02}$           & $50.11_{\pm 0.39}$          & $1.97_{\pm 0.01}$           & $43.53_{\pm 0.42}$          & $2.28_{\pm 0.01}$           \\
        bSAM                   & $62.36_{\pm 0.73}$          & $1.40_{\pm 0.03}$           & $57.97_{\pm 0.70}$          & $1.58_{\pm 0.03}$           & $55.32_{\pm 0.61}$         & $1.70_{\pm 0.03}$          & $51.09_{\pm 0.49}$          & $1.90_{\pm 0.03}$           & $44.77_{\pm 0.42}$          & $2.21_{\pm 0.03}$           \\ \midrule
        SWAG & $59.19_{\pm 0.90}$          & $2.00_{\pm 0.03}$           & $55.45_{\pm 0.88}$          & $2.12_{\pm 0.03}$           & $53.34_{\pm 0.94}$          & $2.19_{\pm 0.03}$           & $49.44_{\pm 0.81}$          & $2.33_{\pm 0.03}$          & $43.71_{\pm 0.93}$          & $2.53_{\pm 0.03} $          \\
        F-SWAG & $59.55_{\pm 2.94}$          & $1.49_{\pm 0.11}$           & $55.10_{\pm 2.82}$          & $1.70_{\pm 0.10}$           & $52.37_{\pm 2.80}$          & $1.82_{\pm 0.10}$           & $48.18_{\pm 2.63}$          & $2.04_{\pm 0.09} $          & $41.84_{\pm 2.43}$          & $2.37_{\pm 0.09}$           \\
        E-MCMC & $62.28_{\pm 0.47}$          & $1.40_{\pm 0.02}$           & $57.84_{\pm 0.46}$          & $1.59_{\pm 0.02}$           & $55.14_{\pm 0.29}$          & $1.71_{\pm 0.02}$           & $50.87_{\pm 0.21}$          & $1.91_{\pm 0.02}$           & $44.49_{\pm 0.13}$          & $2.22_{\pm 0.02}$           \\
        PTL & $61.84_{\pm 0.33}$          & $1.47_{\pm 0.02}$           & $57.36_{\pm 0.22}$          & $1.66_{\pm 0.02}$           & $54.47_{\pm 0.08}$          & $1.78_{\pm 0.01}$           & $50.03_{\pm 0.23}$          & $1.98_{\pm 0.01}$           & $43.34_{\pm 0.36}$          & $2.29_{\pm 0.01}$           \\ \midrule
        FP-BMA & $\textbf{63.91}_{\pm 0.02}$          & $\textbf{1.33}_{\pm 0.00}$           & $\textbf{59.70}_{\pm 0.00}$          & $\textbf{1.51}_{\pm 0.00}$           & $\textbf{57.00}_{\pm 0.01}$         & $\textbf{1.63}_{\pm 0.00}$           & $\textbf{52.51}_{\pm 0.03}$          & $\textbf{1.84}_{\pm 0.00}$           & $\textbf{45.39}_{\pm 0.04}$          & $\textbf{2.18}_{\pm 0.00}$           \\ \bottomrule
        \end{tabular}
        }
    \end{subtable}
\end{figure}

\clearpage
\subsection{Comparison with Diverse Baselines and Inference Methods}\label{subsec:diverse_baselines}
To further validate the broad applicability and effectiveness of FP-BMA, we compare it with a variety of inference algorithms and baselines, including MCMC-based, multi-modal, and advanced VI-based methods. All results are reported for CIFAR-10 (10-shot) with ResNet-18.

\begin{table}[h]
    \centering
    \begin{tabular}{lccc}
        \toprule
        Method & Acc (\%) $\uparrow$ & ECE $\downarrow$ & NLL $\downarrow$ \\
        \midrule
        SGHMC & 55.41$_{\pm 0.88}$ & 0.112$_{\pm 0.009}$ & 1.371$_{\pm 0.025}$ \\
        \textbf{SGHMC + FP-BMA (Ours)} & \textbf{56.41}$_{\pm 1.75}$ & \textbf{0.055}$_{\pm 0.008}$ & \textbf{1.276}$_{\pm 0.021}$ \\
        \midrule
        MoLA & 65.77 & 0.045 & 1.058 \\
        \textbf{MoLA + FP-BMA (Ours)} & \textbf{66.77} & 0.063 & \textbf{0.998} \\
        \midrule
        IVON & 56.23$_{\pm 1.01}$ & 0.023$_{\pm 0.004}$ & 1.262$_{\pm 0.037}$ \\
        \textbf{FP-BMA (VI)} & \textbf{64.98}$_{\pm 1.37}$ & \textbf{0.016}$_{\pm 0.007}$ & \textbf{0.997}$_{\pm 0.046}$ \\
        \bottomrule
    \end{tabular}
    \caption{Comparison of FP-BMA with various inference baselines. All results are based on CIFAR-10 (10-shot) and ResNet-18.}
\end{table}

The table above demonstrates that \textbf{FP-BMA consistently improves predictive performance and calibration across a range of inference backbones and posterior structures:}
\begin{itemize}
    \item When applied on top of \textbf{SGHMC}~\citep{chen2014stochastic} (a standard MCMC method), FP-BMA yields clear improvements in accuracy, ECE, and NLL. This shows that our approach is compatible with and beneficial to MCMC-based inference, extending its utility beyond VI-based methods.
    \item In a multi-modal posterior setting (\textbf{MoLA}~\citep{eschenhagen2021mixtures}), FP-BMA remains effective, improving accuracy and NLL. However, the gains are less pronounced than in unimodal cases, suggesting that further extension of FP-BMA for multi-modal posteriors could be fruitful.
    \item Compared to \textbf{IVON}~\citep{shen2024variational} (which leverages efficient second-order optimization but does not explicitly encourage flatness), FP-BMA achieves significantly better results on all metrics. This highlights the effectiveness of explicitly promoting posterior flatness in Bayesian model averaging.
\end{itemize}

Overall, these results support the broad applicability and complementary nature of FP-BMA, demonstrating its value as a general-purpose enhancement for Bayesian inference, regardless of the underlying approximation strategy.

\clearpage
\subsection{Loss Surface Of Sampled Model}\label{subsec:loss_surface_replic}
\begin{figure}[h]
  \centering
  \begin{subfigure}{0.8\textwidth}
    \centering
    \includegraphics[width=\linewidth]{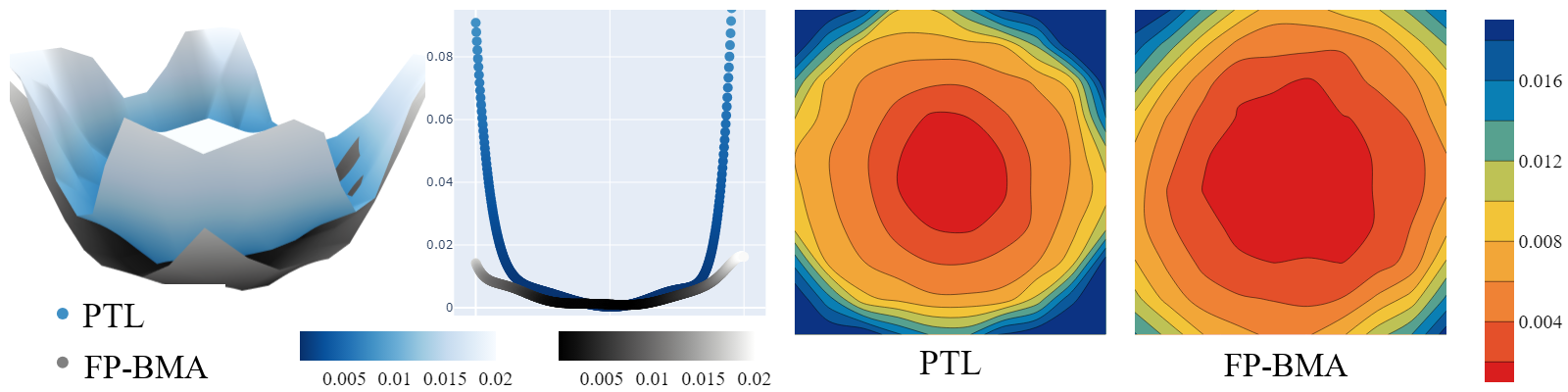}
    \caption{seed 1}
  \end{subfigure}\\%
    \begin{subfigure}{0.8\textwidth}
    \centering
    \includegraphics[width=\linewidth]{figure/loss_surface/loss_surface_Blues_seed2.png}
    \caption{seed 2}
  \end{subfigure}\\%
  \begin{subfigure}{0.8\textwidth}
    \centering
    \includegraphics[width=\linewidth]{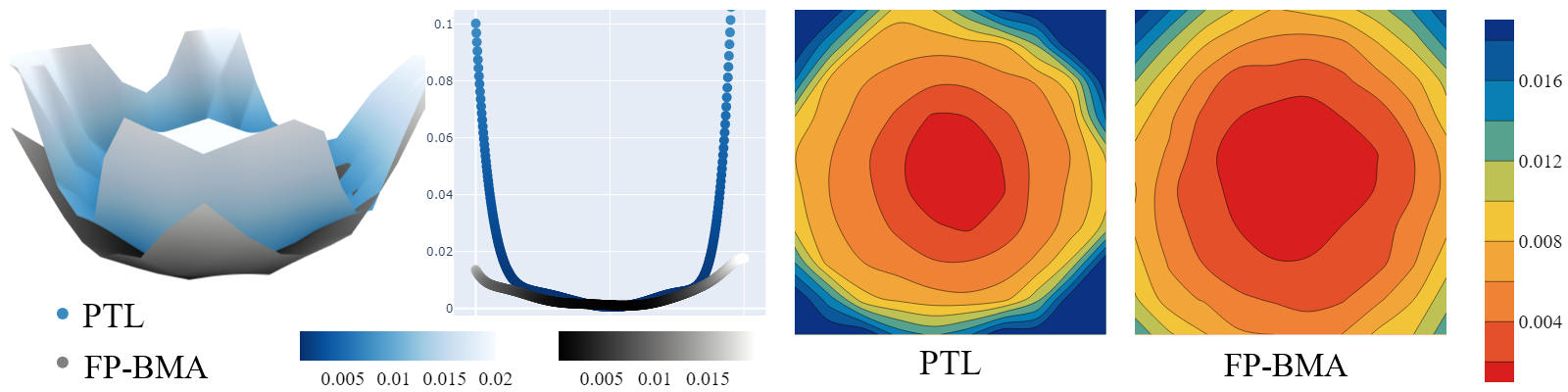}
    \caption{seed 3}
  \end{subfigure}\\%
  \begin{subfigure}{0.8\textwidth}
    \centering
    \includegraphics[width=\linewidth]{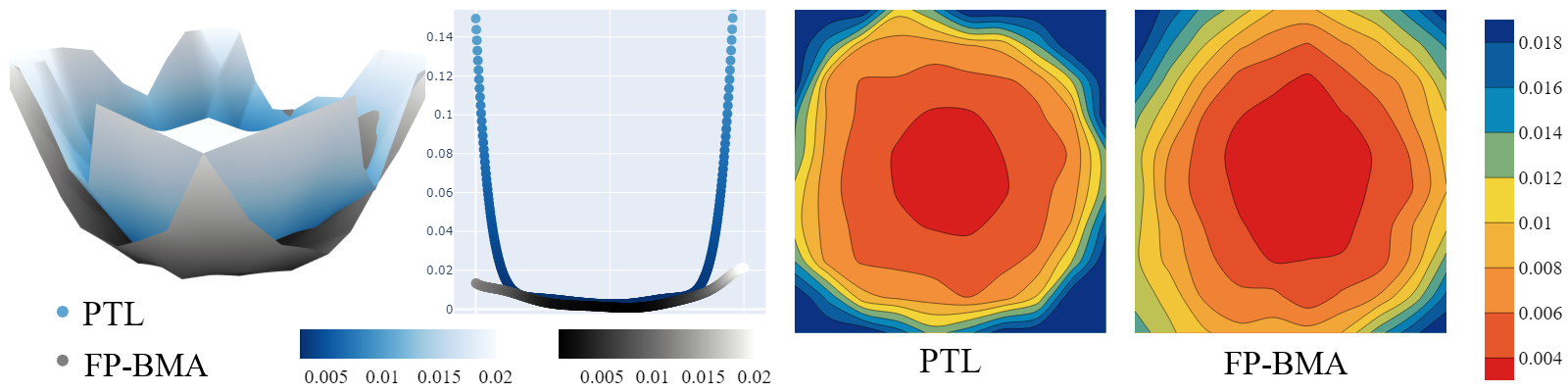}
    \caption{seed 4}
  \end{subfigure}\\[1ex]
  \caption{Four instances of sampled weights, including (b) as presented in Figure \ref{fig:loss_surface}. Across all plots, it is consistently observed that FP-BMA converges to a flatter loss surface compared to PTL.
}
\label{fig:loss_surfaces_seeds}
\end{figure}

As shown in Figure \ref{fig:loss_surface}, we sampled four model parameters from the posterior, which were trained on CIFAR10 with RN18. It shows the consistent and robust trend of flatness of FP-BMA in the loss surface. In Figure \ref{fig:loss_surfaces_seeds}, commencing with the leftmost panel, a 3D surface plot illustrates the loss surface, revealing the FP-BMA model's comparatively flatter topology against the PTL model. This initial plot intuitively demonstrates that the FP-BMA model exhibits a flatter loss surface compared to the PTL model. Following this, the second visualization compresses the information along a diagonal plane into a 1D scatter plot. This transformation reveals areas obscured in the 3D view, highlighting that FP-BMA maintains a considerably flatter and lower-loss landscape. The third and fourth images showcase the loss surface through 2D contour plots, from which one can easily discern that the area representing the lowest loss is significantly more expansive for FP-BMA than for PTL.

\end{document}